\documentclass{easychair}

\usepackage{doc}

\usepackage{pifont}
\usepackage{xcolor}
\definecolor{second}{HTML}{007000}

\usepackage{comment}

\usepackage{amssymb}
\usepackage{pifont}
\usepackage{booktabs}
\usepackage{longtable}
\usepackage{pgfplotstable}
\usepackage{titlesec}
\usepackage{multirow}
\usepackage{subcaption}
\usepackage{fancyvrb}
\usepackage{cleveref}
\usepackage{threeparttable}
\usepackage{wrapfig}
\usepackage{makecell}
\usepackage{hyperref}
\usepackage{tablefootnote}
\usepackage{float}

\pgfplotsset{compat=1.16}

\newcommand{\xmark}{\ding{55}}

\pgfkeysifdefined{/pgfplots/table/output empty row/.@cmd}{
    \pgfplotstableset{
        empty header/.style={
          every head row/.style={output empty row},
        }
    }
}{
    \pgfplotstableset{
        empty header/.style={
            typeset cell/.append code={%
                \ifnum\pgfplotstablerow=-1 %
                    \pgfkeyssetvalue{/pgfplots/table/@cell content}{}%
                \fi
            }
        }
    }
}

\titlespacing*{\paragraph}
{0pt}{2pt}{5pt}

\title{The 6th International Verification of Neural Networks Competition (VNN-COMP 2025): Summary and Results}%

\author{Konstantin Kaulen\inst{1}
        \and Tobias Ladner\inst{2}
        \and Stanley Bak\inst{3}
        \and Christopher Brix\inst{1}
        \and Hai Duong\inst{4}
        \and Thomas Flinkow\inst{5}
        \and Taylor T. Johnson\inst{6}
        \and Lukas Koller\inst{2}
        \and Edoardo Manino\inst{7}
        \and ThanhVu H Nguyen\inst{4}
        \and Haoze Wu\inst{8, 9}
}

\institute{
        RWTH Aachen University, Aachen, Germany\\
        \and
        Technical University of Munich, Munich, Germany\\
        \and 
        Stony Brook University, Stony Brook, New York, USA\\
        \and
        George Mason University, Fairfax, Virginia, USA\\
        \and
        Maynooth University, Maynooth, Ireland\\
        \and
        Vanderbilt University, Nashville, Tennessee, USA\\
        \and
        University of Manchester, Manchester, UK\\
        \and
        Amherst College, Amherst, Massachusetts, USA\\
        \and VMware Research by Broadcom, USA
}

\authorrunning{Kaulen et al.}
\titlerunning{VNN-COMP 2025 Report}

\date{\phantom{date}}

\begin{document}

\maketitle

\begin{abstract}
This report summarizes the 6th International Verification of Neural Networks Competition (VNN-COMP 2025), held as a part of the 8th International Symposium on AI Verification (SAIV), that was collocated with the 37th International Conference on Computer-Aided Verification (CAV). VNN-COMP is held annually to facilitate the fair and objective comparison of state-of-the-art neural network verification tools, encourage the standardization of tool interfaces, and bring together the neural network verification community. To this end, standardized formats for networks (ONNX) and specification (VNN-LIB) were defined, tools were evaluated on equal-cost hardware (using an automatic evaluation pipeline based on AWS instances), and tool parameters were chosen by the participants before the final test sets were made public. In the 2025 iteration, 8 teams participated on a diverse set of 16 regular and 9 extended benchmarks. This report summarizes the rules, benchmarks, participating tools, results, and lessons learned from this iteration of this competition.
\end{abstract}



\section{Introduction}
\label{sec:introduction}

Deep-learning-based systems 
are increasingly being deployed in a wide range of domains, including recommendation systems, computer vision, and autonomous driving. While the nominal performance of these methods has increased significantly over the last years, they largely lack formal guarantees on their behavior. 
However, in safety-critical applications, including autonomous systems, robotics, cybersecurity, and cyber-physical systems (CPS), such guarantees are essential for certification and reliability.

While the literature on the verification of traditionally designed systems is wide and successful, neural network verification remains an open problem, despite significant efforts over the last years. In 2020, the International Verification of Neural Networks Competition (VNN-COMP) was established to facilitate comparison between existing approaches, bring researchers working on this problem together, and help shape future directions of the field. VNN-COMP has been held annually since then~\cite{bak2021vnncomp,muller2022vnncomp,brix2023years,brix2023vnncomp,brix2024vnncomp}. In 2025, the 6th iteration of the annual VNN-COMP\footnote{\url{https://sites.google.com/view/vnn2025/home}} was held as a part of the 8th International Symposium on AI Verification (SAIV)~\cite{saiv2025} that was collocated with the 37th International Conference on Computer-Aided Verification (CAV)~\cite{cav2025}.

This 6th iteration of the VNN-COMP continues last year's trend of increasing standardization and automatization, aiming to enable a fair comparison between the participating tools and to simplify the evaluation of a large number of tools on a variety of (real-world) problems. 
As in the last iteration, VNN-COMP 2025 standardizes (1) neural network and specification formats, ONNX for neural networks and VNN-LIB \cite{vnnlib} for specifications, (2) evaluation hardware, providing participants the choice of a range of cost-equivalent AWS instances with different trade-offs between CPU and GPU performance, and (3) evaluation pipelines, enforcing a uniform interface for the installation and evaluation of tools. The 2025 edition also introduced Competition Contribution papers in the SAIV proceedings for benchmark and tool participants, of which, several of the tools participated with papers~\cite{duong2025saiv_cc,lopez2025saiv_cc,lemesle2025saiv_cc,das2025saiv_cc} that were reviewed by a program committee composed of members of the tool teams.

The competition was kicked off with the solicitation for participation in February 2025. Rule discussion started in March and rules were finalized in May 2025 (see an overview in \Cref{sec:rules}). From April to June 2025, benchmarks were proposed and discussed. Meanwhile, the organizing team decided to continue using AWS as the evaluation platform and started to implement an automated submission and testing system for both benchmarks and tools. By mid-July 2025, eight teams submitted their tools (one team withdrew during the evaluation phase) and the organizers evaluated all entrants to obtain the final results, discussed in \Cref{sec:results} and presented at SAIV on July 22, 2025. 
Discussions were structured into three issues on the official GitHub repository\footnote{\url{https://github.com/VNN-COMP/vnncomp2025/issues}}: rules discussion, benchmarks discussion, and tool submission. All submitted benchmarks\footnote{\url{https://github.com/VNN-COMP/vnncomp2025_benchmarks}} and final results\footnote{\url{https://github.com/VNN-COMP/vnncomp2025_results}} were aggregated in separate GitHub repositories. 

The remainder of this report is organized as follows: \Cref{sec:rules} discusses the competition rules, \Cref{sec:participants} lists all participating tools, \Cref{sec:benchmarks} lists all benchmarks,  \Cref{sec:results} summarizes the results, and \Cref{sec:conclusion} concludes the report, discussing potential future improvements.

\newpage

\section{Rules}
\label{sec:rules}

\paragraph*{Terminology}
An \emph{instance} is defined by a property specification (pre- and post-condition), a network, and a timeout. 
For example, one instance might consist of an MNIST classifier with one input image, a given local robustness threshold $\epsilon$, and a specific timeout.
A \emph{benchmark} is defined as a set of related instances.
For example, one benchmark might consist of a specific MNIST classifier with 100 input images, potentially different robustness thresholds $\epsilon$, and one timeout per input.

\paragraph*{Run-time caps}
Run-times are capped on a per-instance basis, i.e., any verification instance will timeout (and be terminated) after at most X seconds, determined by the benchmark proposer. These can be different for each instance. 
The total per-benchmark runtime (sum of all per-instance timeouts) may not exceed 6 hours per benchmark. 
For example, a benchmark proposal could have six instances with a one-hour timeout, or 100 instances with a 3.6-minute timeout, each.
To enable a fair comparison, we measure the startup overhead for each tool by running it on a range of tiny networks and subtract the minimal overhead from the total runtime.

\paragraph*{Hardware}
To allow for comparability of results, all tools were evaluated on equal-cost hardware using  Amazon Web Services (AWS).
Each team could decide between a range of AWS instance types (see \Cref{tab:instances}) providing a CPU, GPU, or mixed focus.


\begin{table}[h]
\centering
\caption{Available AWS instances.}\label{tab:instances}
\renewcommand{\arraystretch}{1.1}
\scalebox{0.985}{
\begin{tabular}{lccc} \toprule
         &vCPUs & RAM [GB] & GPU \\ 
         \midrule
         p3.2xlarge  & 8 & 61 & V100 GPU with 16 GB memory \\
         m5.16xlarge  & 64 & 256 & \xmark \\
         g5.8xlarge  & 32 & 128 & A10G GPU with 24 GB memory \\
         \bottomrule
\end{tabular}
}
\end{table}

\paragraph{Scoring} 
The final score is aggregate as the sum of all benchmark scores.
Each benchmark score is the number of points (sum of instance scores discussed below) achieved by a given tool, normalized by the maximum number of points achieved by any tool on that benchmark. 
Thus, the tool with the highest sum of instance scores for a benchmark will get a benchmark score of 100, ensuring that all benchmarks are weighted equally, regardless of the number of constituting instances.

\paragraph*{Instance score}
\label{sec:scoring}
Each instance is scored is as follows: 
\begin{itemize}\setlength{\itemsep}{0pt}
    \item Correct hold (property proven): 10 points;
    \item Correct violated (counterexample found): 10 points;
    \item Incorrect result: -150 points (penalty increased compared to 2022);
    \item Timeout / Runtime Error / Unknown: 0 points.

\end{itemize}
However, the ground truth for any given instance is generally not known a priori. In the case of disagreement between tools, we, therefore, place the burden of proof on the tool claiming that a specification is violated, i.e. that a counterexample can be found, and deem it correct exactly if it produces a valid counterexample.

%
The provided counterexamples were supposed to define both the input and the resulting output of the networks.
However, for some tools and instances, the output definition was either missing or differed from the network output as computed by the onnxruntime package used to evaluate counterexamples (by performing inference given the inputs).
The competition rules were ambiguous how this would be handled. We decided to discard all outputs in the counterexample files and base the evaluation solely on the given inputs and their respective outputs as computed by the onnxruntime.

\paragraph*{Time bonus}
As opposed to previous years, no time bonus was awarded.
Instead, all tools that compute the correct result within the time limit receive the same amount of points.

\paragraph{Overhead Correction}
The overhead of tools was measured, but only used to adapt the timeouts. It did not influence the scores, as no time bonus was awarded.
To measure the tool-specific overhead, we created trivial network instances and included those in the measurements. We then observed the minimum verification time over all instances and considered that to be the overhead time for the tool.

\paragraph{Format}
As since 2021, we standardized neural networks to be in \texttt{onnx} format, specifications in \texttt{vnnlib} format, and counterexamples in a format similar to the \texttt{vnnlib} format.
Further, tool authors were required to provide scripts fully automating the installation process of their tool, including the acquisition of any licenses that might be needed. Similar to the previous year, a preparation and execution script had to be provided for running their tool on a specific instance consisting of a network file, specification file, and timeout.
The specifications are interpreted as definitions of counterexamples, meaning that a property is proven ``correct'' if the specification is shown to be unsatisfiable, conversely, the property is shown to be violated if a counterexample fulfilling the specification is found. 
Specifications consisted of disjunctions over conjunctions in both pre- and post-conditions, allowing a wide range of properties from adversarial robustness over multiple hyper-boxes to safety constraints to be encoded.
For example, robustness with respect to inputs in a hyper-box had to be encoded as disjunctive property, where any of the other classes is predicted.

\paragraph{Tracks}
Similar to VNN-COMP'24, the competition was split into two tracks, both of them scored: A "regular" track, and an "extended" track.
The regular track consists of benchmarks selected by the tool participants based on a voting process, where a benchmark is included in the regular track if at least 50\% of tool participants voted for it to be scored. Benchmarks with at least 1 vote and not included in the regular track were scored in the extended track.
All benchmarks received at least one vote, so no benchmark was unscored.

\newpage

\section{Participants}
\label{sec:participants}
We list the tools and teams that participated in the VNN-COMP 2025
in \Cref{tab:tools}
and reproduce their own descriptions of their tools below.



\begin{table*}[h]
\begin{center}
\begin{minipage}{\linewidth}
\caption{Summary of the key features of participating tools. The hardware column describes the used AWS instance with \texttt{p3} and \texttt{g5} making GPUs available, see \Cref{tab:instances} for more details. Licenses refer to the external licenses required to use the corresponding tool, not the licensing of the tool itself.} 
\label{tab:tools}%
\renewcommand{\arraystretch}{1.25}
\resizebox{\textwidth}{!}{ 
\begin{tabular}{lcm{5.0cm}ccc} \toprule
         Tool & References & Organizations & Place  & Hardware & Licenses\\
         \midrule
         $\alpha$,$\beta$-CROWN & \cite{xu2020automatic,xu2021fast,wang2021betacrown,zhang2022general,shi2024genbab} & UIUC, UCLA, Drexel, Duke, RWTH Aachen & \textbf{1} & g5 & GUROBI\\
%
         CORA &  \cite{althoff_2015,koller_et_al_2025settraining,koller_et_al_2025shadows,kochdumper_et_al_2023,ladner_althoff_2023} & Technical University of Munich & 4 & g5 & MATLAB \\
         
%

%
         NeuralSAT & \cite{duong2023dpllt,duong2024harnessing,duong2025neuralsat} & George Mason University & 2 & g5 & GUROBI\\
         nnenum & \cite{bak2020cav,bak2021nnenum} & Stony Brook University & 6 & m5 & - \\
         NNV & \cite{tran2020cav_tool,manzanas2023cav} & Vanderbilt University & 5 & g5 & MATLAB\\
         PyRAT &  \cite{pyrat2024} & Université Paris-Saclay, CEA, List & 3 & m5 & closed source CEA license \\

         SobolBox &  \cite{sobolbox} & IIT Gandhinagar & 7 & m5 & - \\
         \bottomrule
\end{tabular}
}
\end{minipage}
\end{center}
\end{table*}

Place in Table~\ref{tab:tools} is based on the results in \Cref{subsec:zerotol}. We considered an alternative scoring scheme in \Cref{subsec:smalltol}, in which case NNV and nneum switch places. See Section~\ref{sec:results} for a discussion.


\subsection{$\alpha,\!\beta$-CROWN}

\paragraph*{Team} Co-leaders: Xiangru Zhong (student lead) and Huan Zhang (faculty advisor); 
Team members:
Keyi Shen,
Duo Zhou,
Hesun Chen,
Haoyu Li,
Hao Cheng,
Ruize Gao.
All members are from the University of Illinois Urbana-Champaign (UIUC).

The team \textbf{acknowledges} (ordered by last names) Jorge Chavez (UIUC), Hao Chen (UIUC), Keyu Lu (UIUC), and Zhouxing Shi (UCLA), Zhuoxuan Zhang (Brown University), who were involved in the development of the verifier in the year of 2025 but did not directly work on any benchmarks of the competition. A full list of contributors is available at \url{https://abcrown.org}.

\paragraph*{Description} 
$\alpha,\!\beta$-CROWN (\texttt{alpha-beta-CROWN}) is an efficient neural network verifier based on the linear bound propagation framework and built on a series of works on bound-propagation-based neural network verifiers:  CROWN~\cite{zhang2018efficient}, auto\_LiRPA~\cite{xu2020automatic}, $\alpha$-CROWN~\cite{xu2021fast}, $\beta$-CROWN~\cite{wang2021betacrown}, GCP-CROWN~\cite{zhang2022general}, GenBaB~\cite{shi2024genbab}, BICCOS~\cite{zhou2024scalable}.
The core techniques in $\alpha,\!\beta$-CROWN combine the efficient and GPU-accelerated linear bound propagation method with branch-and-bound methods specialized for neural network verification. 

The linear bound propagation algorithms in $\alpha,\!\beta$-CROWN are based on our \texttt{auto\_LiRPA} library~\cite{xu2020automatic}, which supports general neural network architectures (including convolutional layers, pooling layers, residual connections, recurrent neural networks, Transformers, and general computation graphs) and a wide range of nonlinear functions (e.g., ReLU, tanh, trigonometric functions, sigmoid, max pooling and average pooling), and is efficiently implemented on GPUs with Pytorch and CUDA. We jointly optimize intermediate layer bounds and final layer bounds using gradient ascent (referred to as $\alpha$-CROWN or optimized CROWN/LiRPA~\cite{xu2021fast}). Most importantly, we use branch and bound~\cite{bunelunified2018} (BaB) and incorporate split constraints in BaB into the bound propagation procedure efficiently via the $\beta$-CROWN algorithm~\cite{wang2021betacrown}, use cutting-plane method in GCP-CROWN~\cite{zhang2022general} and BICCOS~\cite{zhou2024scalable} to further tighten the bound, and support general nonlinearities in the branch-and-bound by GenBaB~\cite{shi2024genbab}.
For smaller networks, we also use a mixed integer programming (MIP) formulation~\cite{Tjeng2019EvaluatingRO} combined with tight intermediate layer bounds from $\alpha$-CROWN (referred to as $\alpha$-CROWN + MIP~\cite{zhang2022general}). The combination of efficient, optimizable and GPU-accelerated bound propagation with BaB produces a powerful and scalable neural network verifier.

New in this year: $\alpha,\!\beta$-CROWN increases its capability in branch-and-bound using our new Clip-and-Verify framework, which can eliminate partial unsat domains and tighten intermediate layer bounds for both input-space and activation-space branch-and-bound. This approach is quite helpful in handling hard verification problems and can greatly reduce the number of subproblems during branch-and-bound. For example, in the verification of neural network Lyapunov functions~\cite{li2025two}, our new framework can shrink the problem size and total running time by over 50\%. A research paper is currently in submission.

\paragraph*{Link} \url{https://github.com/Verified-Intelligence/alpha-beta-CROWN} (main version)
\paragraph*{Competition submission} \url{https://github.com/Verified-Intelligence/alpha-beta-CROWN_vnncomp2025} (only for reproducing competition results; please use the main version for other purposes)
\paragraph*{Hardware and licenses} CPU and GPU with 32-bit or 64-bit floating point; Gurobi license required for certain benchmarks.
\paragraph*{Participated benchmarks} All benchmarks.

\subsection{CORA}
\paragraph*{Team} Lukas Koller, Tobias Ladner, Matthias Althoff (Technical University of Munich)
\paragraph*{Description} 
CORA~\cite{althoff_2015} enables the formal verification of neural networks, both in open-loop as well as in closed-loop scenarios~\cite{kochdumper_et_al_2023,ladner_althoff_2023}. 
Open-loop verification refers to the task where properties of the output set of a neural network are verified, 
e.g., correctly classified images given noisy input, as also considered at VNN-COMP. 
In closed-loop scenarios, the neural network is used as a controller of a dynamic system, e.g., controlling a car while keeping a safe distance over some time horizon. Moreover, CORA can also train robust neural networks~\cite{koller_et_al_2025settraining,wendl2024training}, 
which uses an efficient batch-wise propagation of zonotopes through a neural network on a GPU. 

For the competition, reachability analysis with zonotopes is used to compute a convex enclosure of the output set of a neural network. In addition to simultaneous input and neuron splitting, a specification-driven input refinement procedure iteratively encloses the set of unsafe inputs~\cite{koller_et_al_2025shadows}. All operations use batch-wise computations to take full advantage of GPU-acceleration.

\paragraph*{Link} \url{https://github.com/kollerlukas/cora-vnncomp2025}

\paragraph*{Commit} 6f1923030baafadfadca3982b72fdea217a92479

\paragraph*{Hardware and licenses} GPU, MATLAB license.

\paragraph*{Participated Benchmarks} 
\texttt{acasxu}, \texttt{cifar100}, \texttt{collins-rul-cnn}, \texttt{cora}, \texttt{dist-shift}, \texttt{nn4sys}, \texttt{safenlp}, \texttt{tinyimagenet}, \texttt{tllverifybench}.

\subsection{NeuralSAT}
\paragraph*{Team} Hai Duong and Thanhvu Nguyen (George Mason).

\paragraph*{Description} 
NeuralSAT~\cite{duong2023dpllt,duong2024harnessing,duong2025neuralsat,duong2025neuralsat2,duong2025compositional,duong2025generating,duong2025saiv_cc} integrates conflict-driven clause learning (CDCL) in SAT/SMT-solving with an DNN abstraction-based theory solver for infeasibility checking.
NeuralSAT implements the DPLL(T) framework used in modern SMT solvers such as  Z3. 
The design of NeuralSAT is inspired by the core algorithms used in SMT solvers such as CDCL components and theory solving. 
The tool is written in Python and uses Gurobi for LP solving. 
Unlike many other modern VNN verification tools,  NeuralSAT \emph{does not} require parameter tuning and works out of the box, e.g., the tool runs on the wide-range of benchmarks in VNN-COMPs without any tuning.


\paragraph*{Link}
\url{https://github.com/dynaroars/neuralsat}

\paragraph*{Commit}
8c6a493d2b9314b06c6f19cd452f1c7ab5bd2657

\paragraph*{Hardware and licences}
GPU, Gurobi License

\paragraph*{Participated benchmarks} All benchmarks in the regular track.

\subsection{nnenum}
\paragraph*{Team} Ali Arjomandbigdeli (Student), Stanley Bak (Supervisor) (Stony Brook University)

\paragraph*{Description} 
The nnenum tool~\cite{bak2021nnenum} uses multiple levels of abstraction to achieve high-performance verification of ReLU networks without sacrificing completeness~\cite{bak2020vnn}. 
The core verification method is based on reachability analysis using star sets~\cite{tran2019fm}, combined with the ImageStar method~\cite{tran2020cav} to propagate sets through all linear layers supported by the ONNX runtime, such as convolutional layers with arbitrary parameters.  
The tool is written in Python 3 and uses GLPK for LP solving.
Over the past few years, we have added preliminary support for single lower and single upper bounds propagation in addition to zonotopes, similar to the DeepPoly method or the CROWN approach. 
We have also added an option to use Gurobi instead of GLPK for LP solving.

\paragraph*{Link}
\url{https://github.com/aliabigdeli/nnenum}

\paragraph*{Commit}
8346c6855e50f5c34c6b9981b18271793c8005cf

\paragraph*{Hardware and licences}
CPU, Gurobi license (optional)

\paragraph*{Participated benchmarks}
\texttt{acasxu}, \texttt{cgan}, \texttt{collins-rul-cnn}, \texttt{cora}, \texttt{linearizenn}, \texttt{metaroom}, \texttt{nn4sys}, \texttt{relusplitter}, \texttt{safeNLP}, \texttt{sat-relu},  \texttt{tllverifybench}, \texttt{vggnet16}.

\subsection{NNV}
\paragraph*{Team} Diego Manzanas Lopez, Navid Hashemi, Samuel Sasaki, Taylor T. Johnson (Vanderbilt University)
\paragraph*{Description} The Neural Network Verification (NNV) Tool~\cite{tran2020cav_tool,manzanas2023cav,lopez2025saiv_cc} is a formal verification software tool for deep learning models and cyber-physical systems with neural network components written in MATLAB and available at \url{https://github.com/verivital/nnv}. NNV uses a star-set state-space representation and reachability algorithm that allows for a layer-by-layer computation of exact or overapproximate reachable sets for feed-forward~\cite{tran2019fm}, convolutional~\cite{tran2020cav}, semantic segmentation (SSNN)~\cite{tran2021cav}, and recurrent (RNN)\cite{tran2023hscc} neural networks, as well as neural network control systems (NNCS)~\cite{tran2019emsoft,tran2020cav_tool} and neural ordinary differential equations (Neural ODEs)~\cite{manzanas2022formats}. 
The star-set based algorithm is naturally parallelizable, which allows NNV to be designed to perform efficiently on multi-core platforms. Additionally, if a particular safety property is violated, NNV can be used to construct and visualize the complete set of counterexample inputs for a neural network (exact-analysis). 
%
For this competition, updated from last year's, we tailor the solver approach depending on the benchmark at hand, although all follow a similar flow. First, we perform a simulation-guided search for counterexamples for a fixed number of samples. If no counterexamples are found (i.e., demonstrate that the property is SAT), then we utilize an iterative refinement approach using reachability analysis to verify the property (UNSAT). This consists of performing reachability analysis using a relax-approximation method~\cite{tran2021cav}, if not verified, then a less conservative approximation based on zonotope pre-filtering approach~\cite{tran2021fac}, and finally using the exact analysis when possible~\cite{tran2020cav} until the specification is verified or there is a timeout. Based on the benchmark to evaluate, the initial reachability analysis may be any of the overapproximation methods or the exact method, based on the complexity of the benchmarks (size of network, input, etc). 
In addition to the star set-based approaches, we implement another approach based on conformal prediction that can scale better to larger networks and input sizes, although only probabilistic guarantees can be provided~\cite{hashemi2025neurips}. This is the only method that uses GPU, all others run solely on CPU.

\paragraph*{Link} \url{https://github.com/verivital/nnv}

\paragraph*{Commit} 696e20d3dbe566ee45cd2e2a3f6c352e44bcd448

\paragraph*{Hardware and licenses} CPU and GPU, MATLAB license.

\paragraph*{Participated Benchmarks} 
Regular track (all) and Extended track (vggnet16, ml4acopf, reluspliter).

\subsection{PyRAT}

\paragraph*{Team} Augustin Lemesle, Julien Lehmann, Tristan Le Gall (CEA-List)
\paragraph*{Description} 
PyRAT (Python Reachability Assessment Tool) \cite{pyrat2024,lemesle2025saiv_cc} is an abstract interpretation based tool to verify the safety and robustness of neural networks. PyRAT relies on multiple zonotopic domains to efficiently compute the reachable states for various architectures of neural networks (fully-connected, convolutionnal, recurrent, transformers, ...). It supports multiple non-linear activation functions (ReLU, Sigmoid, Softmax, Floor, ...) with precise abstractions while maintaining floating-point soundness. PyRAT is correct in that the output bounds reached will always be a sound over-approximation of the results and complete for ReLU based networks in that it will always give a true or false result given enough time.
Depending on the benchmark, different verification modes and domains can be selected. For smaller networks and problems, PyRAT can leverage branch and bound strategies on the inputs with heuristics like ReCIPH~\cite{durand2022reciph}. While on larger ReLU networks, PyRAT will use branch and bound strategies on the ReLU neurons in conjunction with fast GPU computation. 

\paragraph*{Link} \href{https://git.frama-c.com/pub/pyrat/}{https://git.frama-c.com/pub/pyrat}
\paragraph*{Commit} 4a9a4f065a623be395ac4b3385a47ea81638dc48 
\paragraph*{Hardware and licenses} CPU and GPU, closed source CEA licence.
\paragraph*{Participated benchmarks} All benchmarks.

\subsection{SobolBox}
\paragraph*{Team}
Sarthak Das (Indian Institute of Technology Gandhinagar)

\paragraph*{Description}
SobolBox is a spiritual successor to the unsubmitted INNVerS \cite{innvers} project, developed by Shubhajit Roy (IIT Gandhinagar) and Avishek Lahiri (IACS Kolkata) for VNNCOMP-2021, under the supervision of Dr. Rajarshi Ray (IACS Kolkata).
SobolBox \cite{sobolbox,das2025saiv_cc} is a Python-based black-box falsification tool for detecting safety violations in neural networks. It accepts ONNX-format models and VNNLIB safety specifications, treating networks as multi-input, multi-output, non-convex systems without architectural assumptions. Input bounds are extracted via the Z3 theorem prover, then quasi-Monte Carlo Sobol sequence sampling \cite{sobol} generates low-discrepancy candidate points. Local refinement of extrema is performed with L-BFGS-B optimization \cite{lbfgsb}, and the resulting bounds are checked against the safety property using Z3. Results are reported as \texttt{sat} with counterexample, \texttt{unsat} (high-confidence but unsound), or \texttt{unknown}. An optional \texttt{deep} mode applies Automatic Differentiation Variational Inference \cite{advi} to explore complex input regions when the first pass is inconclusive. SobolBox includes caching of Sobol samples and computed output bounds for faster incremental runs.

\paragraph*{Links}
\url{https://github.com/dassarthak18/SobolBox}

\paragraph{Commit}
0ec47e58f0b1ff9513667c61b5bd6350e536b77d

\paragraph*{Hardware and licenses}
CPU, no license required.

\paragraph*{Participated benchmarks}
\texttt{acasxu}, 
\texttt{cersyve}, 
\texttt{collins-rul-cnn}, 
\texttt{linearizenn}, 
\texttt{nn4sys}, 
\texttt{safe-nlp},
\texttt{sat-relu},
\texttt{tllverifybench}, 
\texttt{ml4acopf}.

\newpage
\section{Benchmarks}
\label{sec:benchmarks}

In this section, we provide an overview of all scored benchmarks, reproducing the benchmark proposers' descriptions.
Artifacts for all benchmarks are available in the repository\footnote{\url{https://github.com/VNN-COMP/vnncomp2025_benchmarks}}.

               
\begin{table}[h]
    \centering
    \caption{Overview of all scored benchmarks. 
    }
    \label{tab:benchmarks}
    \resizebox{\textwidth}{!}{
    \renewcommand{\arraystretch}{1.4}
    \begin{tabular}{ccccccc}
    \toprule
    Category &
    Benchmark &
    Application &
    Network Types &
    \# Params &
    Effective Input Dim &
    Track
    \\
    \midrule
    \multirow{8}{*}{Complex} 
    & cGAN & \makecell{Image Generation \\ \& Image Prediction} & Conv. + Vision Transformer & 500k - 68M & 5 & regular \\
    & NN4Sys & \makecell{Dataset Indexing \\ \&  Cardinality Prediction}   & ReLU + Sigmoid & 33k - 37M & 1-308 & regular \\
    & LinearizeNN & NN controller approximation & FC. + Conv. + Vision Transformer + Residual + ReLU & 203k & 4 & regular \\
    & ml4acopf & Power System & Complex (ReLU + Trigonometric + Sigmoid) & 4k-680k & 22 - 402 & extended \\
    & ViT & Vision & Conv. + Residual + Softmax + BatchNorm & 68k - 76k & 3072 & extended \\
    & Collins Aerospace & - & FC + Conv. + Residual, LeakyReLU + MaxPool + Square & 1.8M & 1.2M & extended \\
    & LSNC-ReLU & Lyapunov stability of NN controllers & FC, ReLU & 275 & 6 & extended \\
    & CCTSDB & - & FC + Conv. + Residual, ReLU +  MaxPool + Clip & 100k & 2 & extended \\
    \cmidrule(lr){1-7}
    \multirow{7}{*}{\makecell{CNN \\ \& ResNet}} 
    & Collins RUL CNN & Condition Based Maintenance & Conv. + ReLU, Dropout  & 60k - 262k   & 400 - 800 & regular \\
    & VGGNet16 & Image Classification & Conv. + ReLU + MaxPool    & 138M & 150k & extended \\
    & Traffic Signs Recognition & Image Classification & Conv. + Sign + MakPool + BatchNorm & 905k - 1.7M & 2.7k - 12k & extended \\
    & cifar100 & Image Classification & FC + Conv. + Residual, ReLU + BatchNorm & 2.5M - 3.8M & 3072 & regular \\
    & tinyimagenet & Image Classification & FC + Conv. + Residual, ReLU + BatchNorm & 3.6M & 9408 & regular \\
    & Metaroom & - & Conv. + FC, ReLU & 466k - 7.4M & 5376 & regular \\
    & Yolo & - & FC + Conv. + Residual, ReLU + Sigmoid & 22k - 37M & 1 - 308 & extended \\
    & SoundnessBench & Verifier Test & FC + Conv. + ReLU & 1.7M & 12k & regular \\
    & Relusplitter & Meta-Benchmark & FC + Conv. + ReLU & 13k - 625k & 5 - 3k & extended \\
    & MalBeWare & Malware Classification & FC + Conv. + ReLU & 102k - 1.5M & 4k & regular \\
    & cersyve & Neural Certificates for Control Systems & FC + Residual + ReLU & 1k - 4k & 2 - 6 & regular \\
    \cmidrule(lr){1-7} %
    \multirow{5.5}{*}{\makecell{FC}}
    & TLL Verify Bench & Two-Level Lattice NN & \makecell{Two-Level Lattice NN \\(FC. + ReLU)}  & 17k - 67M & 2 & regular \\
    & Acas XU & Collision Detection & FC. + ReLU & 13k & 5 & regular \\
    & Dist Shift & Distribution Shift Detection & FC. + ReLU + Sigmoid & 342k - 855k & 792 & regular \\
    & safeNLP & Sentence classification & FC. + ReLU & 4k & 30 & regular \\
    & CORA & Image Classification & FC. + ReLU & 575k, 1.1M & 784, 3072 & regular \\
    & SAT ReLU & Verifier Test & FC. + ReLU & 100-35k & 2 - 100 & regular \\
    \bottomrule
    \end{tabular}
    }
\end{table}

\subsection{cGAN}
\paragraph*{Proposed by} Feiyang Cai, Ali Arjomandbigdeli, Stanley Bak (Stony Brook University)
\paragraph*{Motivation}
While existing neural network verification benchmarks focus on discriminative models, the exploration of practical and widely used generative networks remains neglected in terms of robustness assessment.
This benchmark introduces a set of image generation networks specifically designed for verifying the robustness of the generative networks.
\paragraph*{Networks}
The generative networks are trained using conditional generative adversarial networks (cGAN), whose objective is to generate camera images that contain a vehicle obstacle located at a specific distance in front of the ego vehicle, where the distance is controlled by the input distance condition.
The network to be verified is the concatenation of a generator and a discriminator.  The generator takes two inputs: 1) a distance condition (1D scalar) and 2) a noise vector controlling the environment (4D vector). The output of the generator is the generated image. The discriminator takes the generated image as input and outputs two values: 1) a real/fake score (1D scalar) and 2) a predicted distance (1D scalar).
Several different models with varying architectures (CNN and vision transformer) and image sizes (32x32, 64x64) are provided for different difficulty levels.
\paragraph*{Specifications}
The verification task is to check whether the generated image aligns with the input distance condition, or in other words, verify whether the input distance condition matches the predicted distance of the generated image.
In each specification, the inputs (condition distance and latent variables) are constrained in small ranges, and the output is the predicted distance with the same center as the condition distance but with slightly larger range.
\paragraph*{Link} \url{https://github.com/feiyang-cai/cgan_benchmark2023}

\subsection{NN4Sys}
\paragraph*{Proposed by} the $\alpha,\!\beta$-CROWN team with collaborations with Cheng Tan, Haoyu He and Shuyi Lin at Northeastern University.
\paragraph*{Application}
The benchmark contains networks for database learned index, video streaming learned adaptive bitrate, and learned cardinality
estimation which map inputs from various dimensions to 1-dimension outputs.

\begin{itemize}

\item \textit{Background}: learned index, learned cardinality, and learned
    adaptive bitrate are all instances in neural networks for computer systems
        (NN4Sys), which are neural network based methods performing system
        operations. These classes of methods show great potential but have one
        drawback---the outputs of an NN4Sys model (a neural network) can be
        arbitrary, which may lead to unexpected issues in systems.

\item \textit{What to verify}: our benchmark provides multiple pairs of (1) trained NN4Sys model
and (2) corresponding specifications. We design these pairs with different parameters such
that they cover a variety of user needs and have varied difficulties for verifiers. 
We describe benchmark details in our NN4SysBench report:
        \url{http://naizhengtan.github.io/doc/papers/nn4sys23lin.pdf}.

\item \textit{Translating NN4Sys applications to a VNN benchmark}: 
the original NN4Sys applications have some sophisticated structures that are hard to verify.
We tailored the neural networks and their specifications to be suitable for VNN-COMP.
For example, learned index~\cite{kraska18case} contains multiple NNs in a tree structure that together serve one purpose.
However, this cascading structure is inconvenient/unsupported to verify
because there is a ``switch" operation---choosing one NN in the second stage
based on the prediction of the first stage's NN.
To convert learned indexes to a standard form, we train a monolithic (larger) NN.

\item \textit{A note on broader impact}: using NNs for systems is a broad topic, but many existing works
lack strict safety guarantees. We believe that NN Verification can help system developers gain confidence
to apply NNs to critical systems. We hope our benchmark can be an early step toward this vision.

\end{itemize}

\paragraph*{Networks}
This benchmark has twelve networks with different parameters: two for learned
indexes, four for learned cardinality estimation and six for learned adaptive bitrate.
The learned index uses fully-connected feed-forward neural networks. The other
two---the learned cardinality and the learned adaptive bitrate---have a
relatively sophisticated internal structure. Please see our NN4SysBench report
(URL listed above) for details

\paragraph*{Specifications}
For learned indexes,
the specification aims to check if the prediction error is bounded.
The specification is a collection of pairs of input and output intervals such that
any input in the input interval should be mapped to the corresponding output interval.
For learned cardinality estimation and learned adaptive bitrate,
the specifications check the prediction error bounds (similar to the learned indexes)
and monotonicity of the networks.
By monotonicity specifications, we mean that for two inputs, the network should produce a larger
output for the larger input, which is required by cardinality estimation or adaptive bitrate.

\paragraph{Link:} \url{https://github.com/Khoury-srg/VNNComp23_NN4Sys}

\subsection{LinearizeNN}
\paragraph*{Proposed by}  Ali Arjomandbigdeli, Stanley Bak (Stony Brook University).
\paragraph*{Motivation}
Assuming having a neural network controller approximation with a piecewise linear model in the form of a set of linear models with added noise to account for local linearization error. The objective of this benchmark is to investigate the neural network output falls within the range we obtain from our linear model output plus some uncertainty. The idea of this benchmark came from one of our recent paper~\cite{ArjomandBigdeli2024} in which we approximated the NN controller with a piecewise linear model, and we wanted to check if the neural network output falls within the range we obtained from our linear model output plus some uncertainty.

\paragraph*{Networks} The neural network controller we used in this benchmark is an image-based controller for an Autonomous Aircraft Taxiing System whose goal is to control an aircraft's taxiing at a steady speed on a taxiway. This network was introduced  in the paper "Verification of Image-based Neural Network Controllers Using Generative Models"~\cite{katz2021veri}. The neural network integrates a concatenation of the cGAN (conditional GAN) and controller, resulting in a unified neural network controller with low-dimensional state inputs. In this problem, the inputs to the neural network consist of two state variables and two latent variables. The aircraft's state is determined by its crosstrack position (p) and heading angle error ($\theta$) with respect to the taxiway center line. Two latent variables with a range of -0.8 to 0.8 are introduced to account for environmental changes. Because in this case the output spec depends on both the input and output and considering the VNN-LIB limitation, we added a skip-connection layer to the neural network to have the input values present in the output space. We also added one linear layer after that to create a linear equation for each local model.

\paragraph*{Specifications} As mentioned earlier, the aim of this benchmark is to examine whether the neural network output stays within the range defined by the linear model's output, including a margin for uncertainty.Given input $x \in X$ and output $Y = f_{NN}(x)$, the query is of the form: $A_{mat}\times X + b + U_{lb} \leq Y \leq A_{mat}\times X + b + U_{ub}$ for each linear model in its abstraction region.

\paragraph*{Link} \url{https://github.com/aliabigdeli/LinearizeNN_benchmark2024}






\subsection{ml4acopf}
\paragraph*{Proposed by} Haoruo Zhao, Michael Klamkin, Mathieu Tanneau, Wenbo Chen, and Pascal Van Hentenryck (Georgia Institute of Technology), and Hassan Hijazi, Juston Moore, and Haydn Jones (Los Alamos National Laboratory).

\paragraph*{Motivation}
Machine learning models are utilized to predict solutions for an optimization model known as AC Optimal Power Flow (ACOPF) in the power system. Since the solutions are continuous, a regression model is employed. The objective is to evaluate the quality of these machine learning model predictions, specifically by determining whether they satisfy the constraints of the optimization model. Given the challenges in meeting some constraints, the goal is to verify whether the worst-case violations of these constraints are within an acceptable tolerance level.

\paragraph*{Networks}
The neural network designed comprises two components. The first component predicts the solutions of the optimization model, while the second evaluates the violation of each constraint that needs checking. The first component consists solely of general matrix multiplication (GEMM) and rectified linear unit (ReLU) operators. However, the second component has a more complex structure, as it involves evaluating the violation of AC constraints using nonlinear functions, including sigmoid, quadratic, and trigonometric functions such as sine and cosine. This complex evaluation component is incorporated into the network due to a limitation of the VNNLIB format, which does not support trigonometric functions. Therefore, these constraints violation evaluation are included in the neural network.

\paragraph*{Specifications}
In this benchmark, four different properties are checked, each corresponding to a type of constraint violation:
\begin{enumerate}
    \item Power balance constraints: the net power at each bus node is equal to the sum of the power flows in the branches connected to that node.
    \item Thermal limit constraints: power flow on a transmission line is within its maximum and minimum limits.
    \item Generation bounds: a generator's active and reactive power output is within its maximum and minimum limits.
    \item Voltage magnitude bounds: a voltage's magnitude output is within its maximum and minimum limits.
\end{enumerate}

The input to the model is the active and reactive load. The chosen input point for perturbation is a load profile for which a corresponding feasible solution to the ACOPF problem is known to exist. For the feasibility check, the input load undergoes perturbation. Although this perturbation does not exactly match physical laws, the objective is to ascertain whether a machine learning-predicted solution with the perturbation can produce a solution that does not significantly violate the constraints.

The scale of the perturbation and the violation threshold are altered by testing whether an adversarial example can be easily found using projected gradient descent with the given perturbation. The benchmark, provided with a fixed random seed, is robust against the simple projected gradient descent that is implemented.

\paragraph*{Link} \url{https://github.com/AI4OPT/ml4acopf_benchmark}

\subsection{ViT}
\paragraph*{Proposed by} the $\alpha,\!\beta$-CROWN team.
\paragraph*{Motivation}
Transformers~\cite{vaswani2017attention} based on the self-attention mechanism have much more complicated architectures and contain more kinds of nonlinerities, compared to simple feedforward networks with relatively simple activation functions. 
It makes verifying Transformers challenging. We aim to encourage the development of verification techniques for Transformer-based models, and we also aim to benchmark neural network verifiers on relatively complicated neural network architectures and more general nonlinearities. Therefore, we propose a new benchmark with Vision Transformers (ViTs)~\cite{dosovitskiy2020image}. This benchmark is developed based on our work on neural network verification for models with general nonlinearities~\cite{shi2024genbab}.

\paragraph*{Networks}
The benchmark contains two ViTs, as shown in \Cref{tab:vits}.
Considering the difficulty of verifying ViTs, we modify the ViTs and make the models relatively shallow and narrow, with significantly reduced number of layers and attention heads.
Following \cite{shi2019robustness}, we also replace the layer normalization with batch normalization.
The models are mainly trained with PGD training~\cite{madry2017towards}, and we also add a weighted IBP~\cite{gowal2018effectiveness,shi2021fast} loss for one of the models as a regularization.

\begin{table}[ht]
\centering
\caption{Networks in the ViT benchmark.}
\label{tab:vits}
\begin{tabular}{ccc}
\toprule 
Model & \texttt{PGD\_2\_3\_16} & \texttt{IBP\_3\_3\_8} \\
\midrule
Layers & 2 & 3\\
Attention heads & 3 & 3\\
Patch size & 16 & 8\\
Weight of IBP loss & 0 & 0.01\\
Training $\epsilon$ & $\frac{2}{255}$ & $\frac{1}{255}$\\
Clean accuracy & 59.78\% & 62.21\%\\
\bottomrule
\end{tabular}
\end{table}

\paragraph*{Specifications} 
The specifications are generated from the robustness verification problem with $\ell_\infty$ perturbation. 
We use the CIFAR-10 dataset with perturbation size $\epsilon=\frac{1}{255}$ at test time.
We have filtered the CIFAR-10 test set to exclude instances where either adversarial examples can be found (by PGD attack~\cite{madry2017towards} with 100 steps and 1000 restarts) or the vanilla CROWN-like method~\cite{zhang2018efficient,shi2019robustness} can already easily verify. 
We randomly keep 100 instances for each model, with a timeout threshold of 100 seconds. 
Note that since instances with adversarial examples have mostly been excluded during the filtering process, this version of the benchmark may not be able to reflect soundness issues in verifiers, and we refer readers to \cite{zhou2024testing} for discussions on testing soundness with models including ViT.

\paragraph*{Link} \url{https://github.com/shizhouxing/ViT_vnncomp2023}

\subsection{LSNC-ReLU}
\paragraph*{Proposed by} the $\alpha,\!\beta$-CROWN team.
\paragraph*{Motivation}
We develop a benchmark for the problem of verifying the Lyapunov stability of NN controllers in nonlinear dynamical systems within a region-of-intrest and a region-of-attraction. This is important for providing stability guarantees that are essential for safety-critical applications with NN controllers. It is also a useful application of neural network verification as recently demonstrated in \cite{yang2024lyapunov,shi2024certified}, and we refer readers to those works for more details on the problem.

\paragraph*{Networks and Specifications}
The NN controller and Lyapunov function are trained using the technique presented in CT-BaB~\cite{shi2024certified}. Compared to a general nonlinear dynamics as in \cite{yang2024lyapunov}, we utilized a neural-network dynamics. More specifically, the dynamics of this model is given by a shallow neural network with ReLU activations that are trained to approximate the 2D quadrotor dynamics. In this way, only ReLU nonlinearity is presented in the network (no trig functions, leakyrelu, etc.). The verification goal is to certify the Lyapunov stability of the NN controller in a nonlinear dynamical system within a certain Lyapunov sublevel set $(V < c)$. Specifications for the benchmark are randomly generated and consist of random rings of the form $x: c_1 < V(x) < c_2$.

\paragraph*{Link} \url{https://github.com/xiangruzh/LSNC-ReLU}

\subsection{Collins-RUL-CNN}
\paragraph*{Proposed by} Collins Aerospace, Applied Research \& Technology (\href{https://www.collinsaerospace.com/what-we-do/capabilities/technology-and-innovation/applied-research-and-technology}{website}).

\paragraph*{Motivation} Machine Learning (ML) is a disruptive technology for the aviation industry. This particularly concerns safety-critical aircraft functions, where high-assurance design and verification methods have to be used in order to obtain approval from certification authorities for the new ML-based products. Assessment of correctness and robustness of trained models, such as neural networks, is a crucial step for demonstrating the absence of unintended functionalities~\cite{ForMuLA, kirov2023formal}. The key motivation for providing this benchmark is to strengthen the interaction between the VNN community and the aerospace industry by providing a realistic use case for neural networks in future avionics systems~\cite{kirov2023benchmark}.

\paragraph*{Application} Remaining Useful Life (RUL) is a widely used metric in Prognostics and Health Management (PHM) that manifests the remaining lifetime of a component (e.g., mechanical bearing, hydraulic pump, aircraft engine). RUL is used for Condition-Based Maintenance (CBM) to support aircraft maintenance and flight preparation. It contributes to such tasks as augmented manual inspection of components and scheduling of maintenance cycles for components, such as repair or replacement, thus moving from preventive maintenance to \emph{predictive} maintenance (do maintenance only when needed, based on component’s current condition and estimated future condition). This could allow to eliminate or extend service operations and inspection periods, optimize component servicing (e.g., lubricant replacement), generate inspection and maintenance schedules, and obtain significant cost savings. Finally, RUL function can also be used in airborne (in-flight) applications to dynamically inform pilots on the health state of aircraft components during flight. Multivariate time series data is often used as RUL function input, for example, measurements from a set of sensors monitoring the component state, taken at several subsequent time steps (within a time window). Additional inputs may include information about the current flight phase, mission, and environment. Such highly multi-dimensional input space motivates the use of Deep Learning (DL) solutions with their capabilities of performing automatic feature extraction from raw data.

\paragraph*{Networks} The benchmark includes 3 convolutional neural networks (CNNs) of different complexity: different numbers of filters and different sizes of the input space. All networks contain only convolutional and fully connected layers with ReLU activations. All CNNs perform the regression function. They have been trained on the same dataset (time series data for mechanical component degradation during flight).

\paragraph*{Specifications} We propose 3 properties for the NN-based RUL estimation function. First, two properties (robustness and monotonicity) are local, i.e., defined around a given point. We provide a script with an adjustable random seed that can generate these properties around input points randomly picked from a test dataset. For robustness properties, the input perturbation (delta) is varied between 5\% and 40\%, while the number of perturbed inputs varies between 2 and 16. For monotonicity properties, monotonic shifts between 5\% and 20\% from a given point are considered. Properties of the last type ("if-then") require the output (RUL) to be in an expected value range given certain input ranges. Several if-then properties of different complexity are provided (depending on range widths).

\paragraph*{Link} \url{https://github.com/loonwerks/vnncomp2022}

\paragraph*{Paper} Available in~\cite{kirov2023benchmark} or on request.
\subsection{VGGNET16}
\paragraph*{Proposed by} Stanley Bak, Stony Brook University

\paragraph*{Motivation} This benchmark tries to scale up the size of networks being analyzed by using the well-studied VGGNET-16 architecture~\cite{simonyan2014very} that runs on ImageNet. Input-output properties are proposed on pixel-level perturbations that can lead to image misclassification. 

\paragraph*{Networks} All properties are run on the same network, which includes 138 million parameters. The network features convolution layers, ReLU activation functions, as well as max pooling layers.

\paragraph*{Specifications} Properties analyzed ranged from single-pixel perturbations to perturbations on all 150528 pixles (L-infinity perturbations). A subset of the images was used to create the specifications, one from each category, which was randomly chosen to attack. Pixels to perturb were also randomly selected according to a random seed.

\paragraph*{Link} \url{https://github.com/stanleybak/vggnet16_benchmark2022/}

\subsection{Traffic Signs Recognition}
\paragraph*{Proposed by} M\u{a}d\u{a}lina Era\c{s}cu and Andreea Postovan (West University of Timisoara, Romania)
\paragraph*{Motivation} Traffic signs play a crucial role in ensuring road safety and managing traffic flow in both city and highway driving. The recognition of these signs, a vital component of autonomous driving vision systems, faces challenges such as susceptibility to adversarial examples~\cite{szegedy2013intriguing} and occlusions~\cite{zhang2020lightweight}, stemming from diverse traffic scene conditions.

\paragraph*{Networks} Binary neural networks (BNNs) show promise in computationally limited and energy-constrained environments within the realm of autonomous driving~\cite{hubara2016binarized}. BNNs, where weights and/or activations are binarized to $\pm 1$, offer reduced model size and simplified convolution operations for image recognition compared to traditional neural networks (NNs).

We trained and tested various BNN architectures using the German Traffic Sign Recognition Benchmark (GTSRB) dataset~\cite{GTSRB}. This multi-class dataset, containing images of German road signs across 43 classes, poses challenges for both humans and models due to factors like perspective change, shade, color degradation, and lighting conditions. The dataset was also tested using the Belgian Traffic Signs \cite{BelgianTrafficSignDatabase} and Chinese Traffic Signs \cite{ChineseTrafficSignDatabase} datasets. The Belgium Traffic Signs dataset, with 62 classes, had 23 overlapping classes with GTSRB. The Chinese Traffic Signs dataset, with 58 classes, shared 15 classes with GTSRB. Pre-processing steps involved relabeling classes in the Belgium and Chinese datasets to match those in GTSRB and eliminating non-overlapping classes (see \cite{postovan2023architecturing} for details).

\begin{figure}[h]
  \centering
    \includegraphics[width=0.7\textwidth]{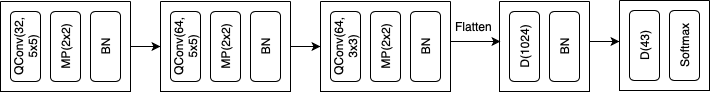}
    \caption{Accuracy Efficient Architecture for GTSRB and Belgium dataset}
    \label{fig:Acc-Efficient-Arch-GTSRB-Belgium}
\end{figure}

\begin{figure}[h]
  \centering
    \includegraphics[width=0.7\textwidth]{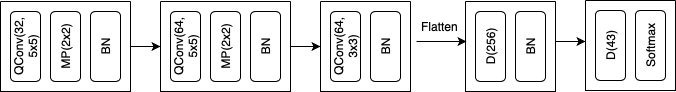}
    \caption{Accuracy Efficient Architecture for Chinese dataset}
    \label{fig:Acc-Efficient-Arch-Chinese}
\end{figure}

\begin{figure}[h]
  \centering
    \includegraphics[width=0.3\textwidth]{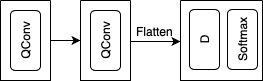}
    \caption{XNOR(QConv) architecture}
    \label{fig:XNOR(QConv)-arch}
\end{figure}

We provide three models with the structure in Figures \ref{fig:Acc-Efficient-Arch-GTSRB-Belgium}, \ref{fig:Acc-Efficient-Arch-Chinese}, and \ref{fig:XNOR(QConv)-arch}. They contain QConv, Batch Normalization (BN), Max Pooling (ML), Fully Connected/Dense (D) layers.  Note that the QConv layer binarizes the corresponding convolutional layer. All models were trained for 30 epochs. The model from Figure \ref{fig:Acc-Efficient-Arch-GTSRB-Belgium} was trained with images having the dimension 64px x 64 px, the one from Figure \ref{fig:Acc-Efficient-Arch-Chinese} with 48px x 48 px and the one from Figure \ref{fig:XNOR(QConv)-arch} with 30px x 30 px. The two models involving Batch Normalization layers introduce real valued parameters besides the binary ones, while the third one contains only binary parameters (see Table \ref{tab:stats}) for statistics.

\begin{table}[h]
\caption{Training and Testing Statistics}
\label{tab:stats}
\centering
\scriptsize
\begin{tabular}{|c|c|ccc|ccc|}
\hline
\multirow{2}{*}{\textbf{Input size}} & \multirow{2}{*}{\textbf{Model name}} & \multicolumn{3}{c|}{\textbf{Accuracy}}                                      & \multicolumn{3}{c|}{\textbf{\#Params}}                                      \\ \cline{3-8} 
                            &                             & \multicolumn{1}{c|}{\textbf{German}} & \multicolumn{1}{c|}{\textbf{China}} & \textbf{Belgium} & \multicolumn{1}{c|}{\textbf{Binary}}  & \multicolumn{1}{c|}{\textbf{Real}} & \textbf{Total}   \\ \hline
64px $\times$ 64px          & Figure \ref{fig:Acc-Efficient-Arch-GTSRB-Belgium}                  & \multicolumn{1}{c|}{96.45}  & \multicolumn{1}{c|}{81.50} & 88.17   & \multicolumn{1}{c|}{1772896} & \multicolumn{1}{c|}{2368} & 1775264 \\ \hline
48px $\times$ 48px          & Figure \ref{fig:Acc-Efficient-Arch-Chinese}                  & \multicolumn{1}{c|}{95.28}  & \multicolumn{1}{c|}{83.90} & 87.78   & \multicolumn{1}{c|}{904288}  & \multicolumn{1}{c|}{832}  & 905120  \\ \hline
30px $\times$ 30px          & Figure \ref{fig:XNOR(QConv)-arch}                  & \multicolumn{1}{c|}{81.54}  & \multicolumn{1}{c|}{N/A}   & N/A     & \multicolumn{1}{c|}{1005584} & \multicolumn{1}{c|}{0}    & 1005584 \\ \hline
\end{tabular}
\end{table}
\paragraph*{Specifications} To evaluate the \emph{adversarial robustness} of the networks above, we assessed perturbations within the infinity norm around zero, with the radius denoted as $\epsilon = \{1, 3, 5, 10, 15\}$. This involved randomly selecting three distinct images from the GTSRB dataset's test set for each model and generating \textsc{VNNLIB} files for each epsilon in the set. In total, we created 45 \textsc{VNNLIB} files. Due to a 6-hour total timeout constraint for solving all instances, each instance had a maximum timeout of 480 seconds. To review the generated \textsc{VNNLIB} specification files submitted to VNNCOMP 2023, as well as to generate new ones, please refer to \url{https://github.com/apostovan21/vnncomp2023}.

\paragraph*{Link} \url{https://github.com/apostovan21/vnncomp2023}

\subsection{CIFAR100}

\paragraph*{Proposed by} the $\alpha,\!\beta$-CROWN team.
\paragraph*{Motivation} This benchmark is reused from VNN-COMP 2022 with a reduced complexity (only two out of the four models with medium sizes are retained). 
See details in Section 4.5 of the report of VNN-COMP 2022~\cite{muller2022vnncomp}.

\paragraph*{Networks} We provide two ResNet models on CIFAR-100 with different model widths and depths (input dimension $32 \times 32 \times 3$, 100 classes):
\begin{itemize}
    \item \texttt{CIFAR100-ResNet-medium}: 8 residual blocks, 17 convolutional layers + 2 linear layers
    \item \texttt{CIFAR100-ResNet-large}: 8 residual blocks, 19 convolutional layers + 2 linear layers (almost identical to standard ResNet-18 architecture)
\end{itemize}

\paragraph*{Specifications} 
We randomly select 100 images from the CIFAR-100 test set with a verification timeout of 100 seconds for each of the two models. 
We filtered out the samples which can be verified by vanilla CROWN (which is used during training) to make the benchmark more challenging. The filtering process is done offline on a machine with a GPU due to the large sizes of these models. 
A small proportion of instances (around 18\%) with adversarial examples have been retained for potentially identifying unsound results. 

\paragraph*{Link} \url{https://github.com/huanzhang12/vnncomp2024_cifar100_benchmark}

\subsection{TinyImagenet}

\paragraph*{Proposed by} the $\alpha,\!\beta$-CROWN team.
\paragraph*{Motivation} This benchmark is reused from VNN-COMP 2022. See details in Section 4.5 of the report of VNN-COMP 2022~\cite{muller2022vnncomp}.

\paragraph*{Networks} We provide a ResNet for TinyImageNet (input dimension $64 \times 64 \times 3$, 200 classes):
\begin{itemize}
    \item \texttt{TinyImageNet-ResNet-medium}: 8 residual blocks, 17 convolutional layers + 2 linear layers
\end{itemize}

\paragraph*{Specifications} 
We randomly select 200 images from the TinyImageNet test set with a verification timeout of 100 seconds for each of the two models. A filtering procedure has been adopted similar to the CIFAR100 benchmark.

\paragraph*{Link} \url{https://github.com/huanzhang12/vnncomp2024_tinyimagenet_benchmark}

\subsection{Yolo}
\paragraph*{Proposed by} the $\alpha,\!\beta$-CROWN team.

\paragraph*{Motivation} Object detection is central to many safety-critical applications, yet verification research has had limited benchmarks that capture its unique challenges. Even in a simplified form, YOLO highlights key aspects of detection tasks: multiple outputs, spatial reasoning over bounding boxes, and the need to ensure reliable perception under perturbations. A YOLO benchmark therefore provides a concrete step toward advancing verification methods for complex perception tasks such as object detection.

\paragraph*{Networks} The model is a modified version of YOLOv2. For simplicity of verification, the backbone of the original YOLOv2 model is replaced with a much smaller network. The network takes in an image ($3\times 52\times 52$) and outputs a tensor of a shape $125\times 13\times 13$, which contains the confidence scores, the classification result, and the positions of 5 bounding boxes for each of the $13 \times 13$ grid cells. The network was trained using adversarial training, incorporating an $\ell_{1}$ regularization term in the loss.

\paragraph*{Specifications} Robustness properties are defined on the inputs and raw output tensors of the backbone network. For each instance, the property is verified if there is at least one bounding box successfully predicting the same object (with a confidence score greater than a preset threshold) given any bounded random perturbations on the input image.

\paragraph*{Link} \url{https://github.com/xiangruzh/Yolo-Benchmark}

\subsection{SoundnessBench}
\paragraph*{Proposed by} the $\alpha,\!\beta$-CROWN team.
\paragraph*{Motivation}
We introduce SoundnessBench, a novel benchmark designed to evaluate the  soundness of neural network (NN) verifiers, inspired by the work~\cite{zhou2024testing}. While recent years have seen significant progress in developing both NN verifiers and benchmarks, existing benchmarks typically lack ground truth for challenging instances. As a result, it remains difficult to determine whether a verification claim is sound when no verifier can falsify the property under consideration.

\paragraph*{Networks} SoundnessBench includes a simple feedforward neural network with minimal architectures and standard operators widely supported by mainstream verifiers. These operators include fully connected (linear) layers, 2D convolutional layers, ReLU activations, as well as reshaping and flattening operations between 2D and 1D formats.

\paragraph*{Specifications} The verification tasks are synthetic, with some specifications intentionally designed to embed counterexamples that are difficult for standard falsification methods—such as projected gradient descent (PGD)—to discover. Specifications are expressed as conjunctions of clauses. \textbf{All instances are SAT}, and any UNSAT reports likely indicate a verifier bug.

\paragraph*{Link} \url{https://github.com/keyis2/SoundnessBench_vnncomp2025}

\subsection{Relusplitter}

\paragraph{Proposed by} Linhan Li and ThanhVu Nguyen (George Mason University)
\paragraph{Motivation}
Many existing DNN verification benchmarks are becoming ineffective at evaluating modern verifiers because their instances are often too easy or too hard and lack architectural diversity. To address these issues and allow for the reuse of established benchmarks, we propose Relusplitter \cite{Li2025destabilizing}. It takes a DNN verification instance (network and property) and generates a semantically equivalent network on which the property is harder to verify. Relusplitter achieves this by systematically destabilizing stable neurons in the network.

\paragraph{Description}
This benchmark uses selected instances from benchmarks used in previous VNNCOMP iterations as seed inputs to generate more challenging instances. The networks include both fully connected (FC) and convolutional (Conv) architectures. All networks use a standard feed-forward structure. The timeout for the instances was reduced based on results from prior iterations, and the generated instances have 3x timeout. 
Benchmarks used as seed inputs include (1) ACAS Xu, (2) MNIST\_FC, (3) Oval21, and (4) Cifar Biasfield.

\paragraph{Networks} The seed instances use the original network, and the generated instances use a destabilized network generated from the original network.
\paragraph{Specifications} The original specification from the seed instance was used for each generated instance. 
\paragraph*{Link} \url{https://github.com/dynaroars/relusplitter/tree/VNNCOMP-25}

\subsection{TLL Verify Bench}
\paragraph*{Proposed by} James Ferlez (University of California, Irvine)

\paragraph*{Motivation} This benchmark consists of Two-Level Lattice (TLL) NNs, which have been shown to be amenable to fast verification algorithms (e.g. \cite{FerlezKS22}). Thus, this benchmark was proposed as a means of comparing TLL-specific verification algorithms with general-purpose NN verification algorithms (i.e. algorithms that can verify arbitrary deep, fully-connected ReLU NNs).

\paragraph*{Networks}  The networks in this benchmark are a subset of the ones used in \cite[Experiment 3]{FerlezKS22}. Each of these TLL NNs has $n=2$ inputs and $m=1$ output. The architecture of a TLL NN is further specified by two parameters: $N$, the number of local linear functions, and $M$, the number of selector sets. This benchmark contains TLLs of sizes $N = M = 8, 16, 24, 32, 40, 48, 56, 64$, with $30$ randomly generated examples of each (the generation procedure is described in \cite[Section 6.1.1]{FerlezKS22}). At runtime, the specified verification timeout determines how many of these networks are included in the benchmark so as to achieve an overall 6-hour run time; this selection process is deterministic. Finally, a TLL NN has a natural representation using multiple computation paths \cite[Figure 1]{FerlezKS22}, but many tools are only compatible with fully-connected networks. Hence, the ONNX models in this benchmark implement TLL NNs by ``stacking'' these computation paths to make a fully connected NN (leading to sparse weight matrices: i.e. with many zero weights and biases). The \texttt{TLLnet} class (\url{https://github.com/jferlez/TLLnet}) contains the code necessary to generate these implementations via the \texttt{exportONNX} method.

\paragraph*{Specifications}  All specifications have as input constraints the hypercube $[-2,2]^2$. Since all networks have only a single output, the output properties consist of a randomly generated real number and a randomly generated inequality direction. Random output samples from the network are used to roughly ensure that the real number property has an equal likelihood of being within the output range of the NN and being outside of it (either above or below all NN outputs on the input constraint set). The inequality direction is generated independently and with each direction having an equal probability. This scheme biases the benchmark towards verification problems for which counterexamples exist. 

\paragraph*{Link} \url{https://github.com/jferlez/TLLVerifyBench}
\paragraph*{Commit}
199d2c26d0ec456e62906366b694a875a21ff7ef

\subsection{ACAS Xu}
\paragraph{Networks} The ACASXu benchmark consists of ten properties defined over 45 neural networks used to issue turn advisories to aircraft to avoid collisions. The neural networks have 300 neurons arranged in 6 layers, with ReLU activation functions. There are five inputs corresponding to the aircraft states, and five network outputs, where the minimum output is used as the turn advisory the system ultimately produces.

\paragraph{Specifications} We use the original 10 properties~\cite{katz2017reluplex}, where properties 1-4 are checked on all 45 networks as was done in later work by the original authors~\cite{katz2019marabou}. Properties 5-10 are checked on a single network. The total number of benchmarks is therefore 186. The original verification times ranged from seconds to days---including some benchmark instances that did not finish. This year we used a timeout of around two minutes (116 seconds) for each property, in order to fit within a total maximum runtime of six hours.





\subsection{Real-world distribution shifts}
\paragraph*{Proposed by} the Marabou team.
\paragraph*{Motivation}
While robustness against handcrafted perturbations (e.g., norm-bounded) for perception networks are more commonly investigated, robustness against real-world distribution shifts~\cite{wu2022toward} are less studied but of practical interests. This benchmark set contains queries for verifying the latter type of robustness.  
\paragraph*{Networks} The network is a concatenation of a generative model and a MNIST classifier. The generative model is trained to take in an unperturbed image and an embedding of a particular type of distribution shifts in latent space, and produce a perturbed image. The distribution shift captured in this case is the "shear" perturbation. 
\paragraph*{Specifications} The verification task is to certify that a classifier correctly classifies all images in a perturbation set, which is a set of images generated by the generative model given a fixed image and a ball centering the mean perturbations on this image (in the latent space). This mean perturbation is computed by a prior network.
\paragraph*{Link} \url{https://github.com/wu-haoze/dist-shift-vnn-comp}

\subsection{safeNLP}

\begin{figure}[htbp]
    \centering
    \includegraphics[width=0.9\columnwidth]{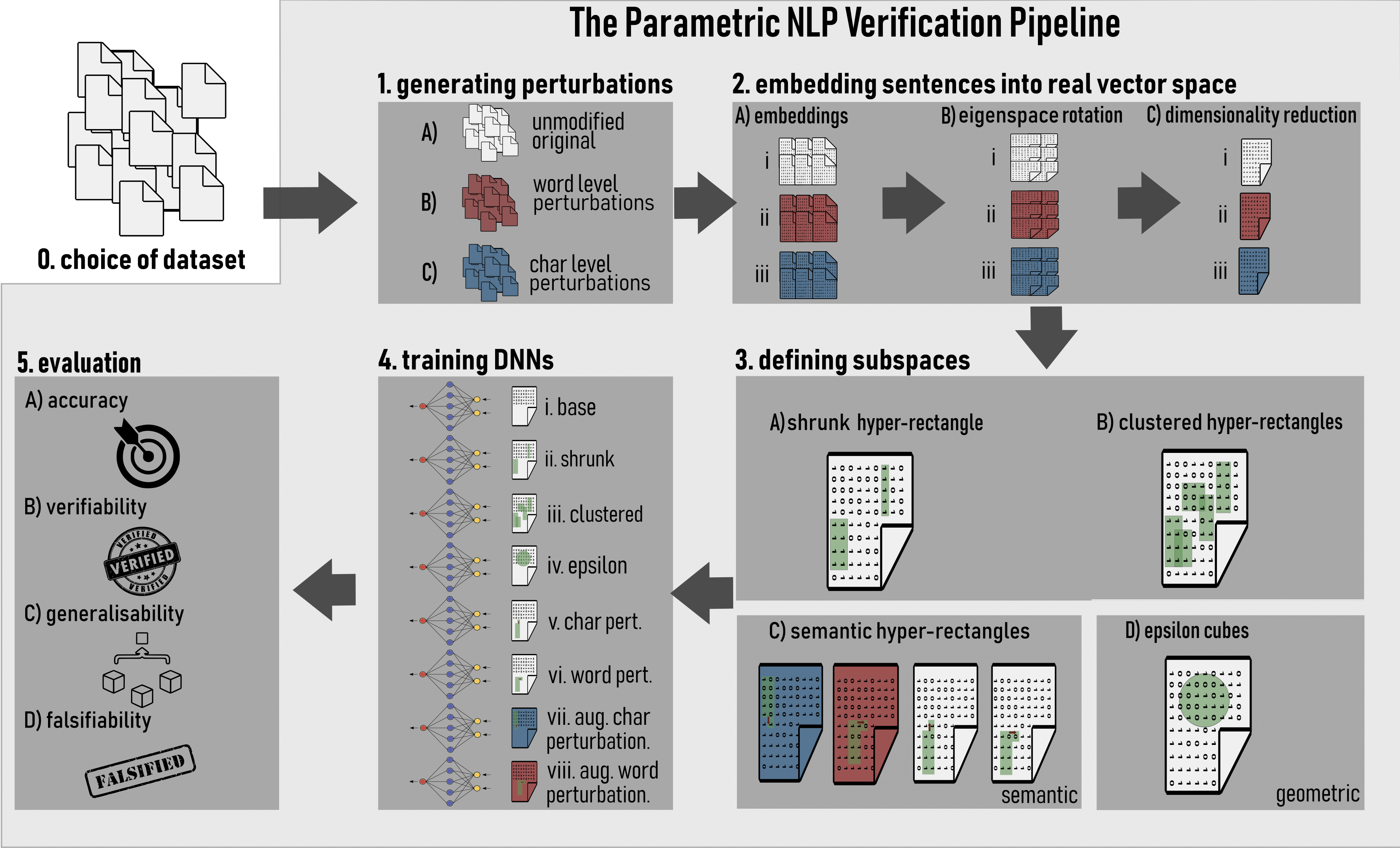}
    \caption{\small\emph{Generic approach to generating the NLP verification pipelines~\cite{casadio2023antonio,casadio2024nlp} deployed to obtain the safeNLP benchmark.}}
   \label{fig:antonio}
\end{figure}

\paragraph*{Proposed by} Marco Casadio, Ekaterina Komendantskaya, Luca Arnaboldi, Tanvi Dinkar.

\paragraph*{Motivation}
While considerable research has been dedicated to the verification of DNN-based systems in domains such as computer vision, there has been a notable lack of focus on the verification of natural language processing (NLP) systems. This is particularly critical given the rise of conversational agents across various domains, where inaccurate or misleading responses can cause real-world harm. For example, recent EU legislation~\cite{EUlaw} requires chatbots to disclose their non-human nature when queried, and developers of the chatbots should provide firm, and if possible, formal, guarantees that such disclosure will be given in an accurate manner. Medical assistants give another example where formal guarantees about the conversational agent responses are needed in order to safeguard against chatbots  generating harmful medical advice~\cite{bickmore2018patient}. While some initial work has been done in this area of NLP verification~\cite{jia2019certified,huang2019achieving,welbl2020towards,zhang2021certified,wang2023robustness,ko2019popqorn,du2021cert,shi2020robustness,bonaert2021fast}, no agreement on commonly  accepted benchmarks has been reached in this domain. To address this gap, we introduce safeNLP, the first such benchmark. 

\paragraph*{Application}
In~\cite{casadio2024nlp}, we have undertaken a large-scale study of the existing literature on  NLP verification, and distilled common patterns among the existing approaches. Usually, given a dataset consisting of sentences divided into classes, Large Language Models (LLMs) are used to embed these sentences into real-vector spaces, after which smaller neural networks are trained to classify the embedded vectors (relative to the originally given classes). For verification, one can generate meaning-preserving sentence perturbations, again embed them into vector spaces, and verify that subspaces that contain the (embeddings of) the perturbed sentences are classified correctly. Also, in line with classical verification pipelines~\cite{CasadioKDKKAR22}, one can use these input subspaces to train the neural network to be robust on them.  The problem was that each of the existing approaches~\cite{jia2019certified,huang2019achieving,welbl2020towards,zhang2021certified,wang2023robustness,ko2019popqorn,du2021cert,shi2020robustness,bonaert2021fast} used parts of this pipeline in different ways, which made it difficult to compare or audit the results. In~\cite{casadio2023antonio,casadio2024nlp}, we made a generic implementation of this pipeline, 
where each of the components of the pipeline is implemented in a modular and transparent way. For example, we can choose  and vary embedding functions, training modes, algorithms for sentence perturbations and algorithms for robust training, independently and modularly; as shown in Figure~\ref{fig:antonio}. This implementation was used to generate the presented VNNCOMP benchmark.

\begin{itemize}
    \item \emph{Datasets:} Although there was no clear consensus in~\cite{jia2019certified,huang2019achieving,welbl2020towards,zhang2021certified,wang2023robustness,ko2019popqorn,du2021cert,shi2020robustness,bonaert2021fast}, the most frequently used dataset in prior works was the IMDB dataset containing film reviews. Its disadvantage is unclear relation to safety critical domains that usually motivate verification efforts. On the other hand, none of the previously used datasets concerned safety-critical applications of NLP. We decided to address this problem, and therefore applied our generic NLP verification pipeline on two safety-critical datasets:  R-U-A-Robot~\cite{gros2021ruarobot}, which focuses on the chatbot disclosure problem, and Medical~\cite{abercrombie2022risk}, which addresses the issue of harmful advice provided by medical chatbots. Both datasets are pre-processed into two classes, positive and negative, to simplify the verification task. For further details on the pre-processing steps and datasets, see~\cite{casadio2024nlp} and the \href{https://github.com/ANTONIONLP/safeNLP}{benchmark GitHub repository}.
    \item \emph{Input Space:} In both datasets, sentences are transformed into fixed-size vector representations, i.e. embeddings, which serve as the inputs to the neural networks. For this VNNCOMP benchmark, we used Sentence-BERT~\cite{reimers-gurevych-2019-sentence}. 
    \item \emph{What to Verify:} For each dataset, we generated meaning-preseving sentence perturbations at character and word level as in Moradi et al.~\cite{moradi2021evaluating} and at sentence level with Vicuna~\cite{vicuna2023}. For each positive sentence in the dataset, the smallest hypercube containing the embeddings of all of its obtained perturbations formed one input subspace for verification. Such subspaces were obtained for all positive sentences from the given data set, and were subject to VNCCOMP verification challenge. 
    \item \emph{A note on broader impact:} Verified models can serve as filters for larger NLP systems: e.g. to screen inputs to ensure they meet safety criteria before being passed on to more complex models.
\end{itemize}

\paragraph*{Networks} The safeNLP benchmark includes two neural networks, each corresponding to a different dataset (R-U-A-Robot and Medical). Both networks share the same architecture, consisting of two fully-connected layers. The hidden layer has 128 units with a ReLU activation function, while the output layer has 2 units representing the two classification classes (positive/negative). To enhance the robustness of the networks to the specified safety requirements, they are trained using a custom PGD (Projected Gradient Descent)~\cite{madry2018towards} adversarial training technique. In particular, the PGD attack explores the above-mentioned subspaces of the input space (cf. also Figure~\ref{fig:antonio}). 

\paragraph*{Specifications} The benchmark uses hyper-rectangles in the 30-dimensional embedding space as the subspaces of choice, offering a computationally efficient way to define more precise and adaptable regions compared to the traditional $\epsilon$-cubes. 
The specifications require verifying that, for a given network and hyper-rectangle, every point within the hyper-rectangle is classified as the positive class by the network. To meet time constraints, we randomly select 1,080 such specifications, each linked to one of the two networks and a corresponding hyper-rectangle, with a timeout of 20 seconds per specification.

\paragraph*{Link} \url{https://github.com/ANTONIONLP/safeNLP}

\subsection{CORA Benchmark}
\paragraph*{Proposed by} the CORA team.
\paragraph*{Motivation}
The verification of neural networks can be quite slow, i.e., the verification of a single instance can take multiple days -- which is often hard to justify, particularly in safety-critical scenarios. To encourage the fast verification of neural networks, our benchmark focuses on the verification time by setting a small timeout and testing three different (adversarial) training techniques that aim to ease the verifiability.
\paragraph*{Networks} The benchmark consists of one ReLU-neural network architecture (7x250 + ReLU), which was trained on three datasets, (MNIST, SVHN, and CIFAR10), using three different (adversarial) training methods, i.e., standard (point), interval-bound propagation, and set-based. Both interval-bound propagation and set-based training are training methods that improve the robustness of the trained neural network and aim to ease later verification. The neural networks are taken from the first evaluation run of~\cite{koller_et_al_2025settraining}; please refer to~\cite{koller_et_al_2025settraining} for the training details.
\paragraph*{Specifications} All networks are trained on classification tasks. The goal is to verify that no image within a given input set is incorrectly classified.
\paragraph*{Link} \url{https://github.com/kollerlukas/cora-vnncomp2024-benchmark}

\subsection{SAT ReLU}

\paragraph{Proposed by} Edoardo Manino (The University of Manchester)

\paragraph{Motivation} Neural network verifiers are fundamentally not different from any other piece of software in that they may contain bugs. As such, it is necessary to test their behavior for correctness. To this end, the SAT ReLU benchmark is constructed with the additional requirement of having its verification verdicts (safe or unsafe) known a priori. With it, we can test all neural network verifiers for correctness.

\paragraph{Description} The SAT ReLU benchmark is entirely synthetic. Each instance is built by generating a random 3-SAT boolean formula with known satisfiability verdict. In total, we generate $50$ satisfiable formulae and $50$ unsatisfiable ones in CNF format. We vary the number of variables in $v=[2,100]$ and the number of clauses in $c=[v,5v]$. Then, we encode the formulae as neural network verification problems.

\paragraph{Networks} We encode each CNF formula as a fully-connected ReLU network with one hidden layer. To do so, we take inspiration from the ReLU gadgets introduced by \cite{katz2017reluplex}. First, we assume that each element of the input $x\in\mathbb{R}^v$ represents a boolean variable $x_i\in\{0,1\}$. Half of the network encodes the boolean arithmetic necessary to compute the CNF formula under this assumption. Second, we let each variable vary in $x_i\in[0,1]$ and use the other half of the network to encode the necessary binarization constraints. Finally, we merge the two halves into a single network. The final network has $v$ inputs, $c+2v$ hidden activations, and $2$ outputs.

\paragraph{Specifications} Given the nature of the networks we construct, the original CNF formula is satisfiable if and only if there exists some input $x\in\{0,1\}^n$ that produces an output of $y_1\geq1$. Non-binary inputs cause the second output to be positive, and thus are rejected by setting a binarization constraint of $y_2\leq0$.

\paragraph*{Link} \url{https://github.com/emanino/neurocodebench}

\subsection{MetaRoom}
\paragraph*{Proposed by} Hanjiang Hu (CMU), Zuxin Liu (CMU), Linyi Li (UIUC), Jiacheng Zhu (CMU), Ding Zhao (CMU).

\paragraph*{Motivation} MetaRoom Benchmark is based on the MetaRoom dataset, which aims at the robustness certification and verification of deep learning based vision models against camera motion perturbation in the indoor robotics application. Specifically, this benchmark focuses on the classification of 20 labels against camera moves within tiny perturbation radii of translation along the z-axis and rotation along the y-axis (e.g. 1e-5 m, 2.5e-4 degree). The MetaRoom dataset is a realistic indoor object recognition dataset with dense point cloud maps, which enables image projections from different camera poses and simulates object-centric semantic robustness in real-world robotic applications. More details about the dataset can be found in \cite{hu2022robustness}.

\paragraph*{Networks} The models of \texttt{cnn\_4layer} and \texttt{cnn\_6layer} are 4-layer and 6-layer convolutional neural networks with a last layer of a fully connected layer for multi-class classification. The input shape is \texttt{1*3*32*56}, which is projected from  one-dimensional relative camera rotation or translation around 0 (e.g. [-1e-5, 1e-5]) from a very dense point cloud based on direct Z-buffering. And the output is for the classification of 20 labels. The pretrained models give correct predictions at camera motion movement origins. 

\paragraph*{Specifications} Simplified from the 1-dim semantic perturbation studied in \cite{hu2023robustness,hu2024pixel}, the benchmark for VNN-COMP adopts the pre-projected image specifications from a very dense point cloud, where the lower and upper bounds are all pixels of the image during the projection within the camera perturbation radius of translation along the z-axis or rotation along the y-axis. The output specifications will be the classification labels at camera movement origins.

\paragraph*{Link} \url{https://github.com/HanjiangHu/metaroom_vnn_comp2023}

\subsection{cersyve}
\paragraph*{Proposed by} Yujie Yang (THU, CMU), Hanjiang Hu (CMU), Tianhao Wei (CMU), Shengbo Eben Li (THU), Changliu Liu (CMU).

\paragraph*{Motivation} Neural safety certificate verification is a special type of verification problems that require neural networks to satisfy certain properties everywhere in the state space, i.e., the input set of the verification problem is the entire state space. This is different from most existing benchmarks, which only consider verification in either a small disturbance set around data samples or part of the state space, such as MNIST, CIFAR, and ACAS Xu. Another distinct feature of safety certificate verification is that it not only involves a single safety certificate network but also needs system dynamics and control policies for verification, and certain conversions are required before these components can be formulated as a standard verification problem.

\paragraph*{Description} While this version of cersyve only contains necessary networks and specifications for verification, our full version also includes a set of neural safety certificate synthesis tools, including pre-training, adversarial training, and verification-guided training modules, as well as evaluation tools for synthesized certificates. For a detailed description of cersyve, please refer to our paper~\cite{yang2025scalable}.

\paragraph*{Networks} The cersyve benchmark contains nine commonly used control tasks with state dimensions ranging from two to six. These tasks include both linear and nonlinear dynamics and safety constraints. We provide two ONNX models for each task. One is a pre-trained safety certificate, and the other is fine-tuned. The models are already integrated with all necessary elements for verification, such as system dynamics, constraints, and control policies. Users can view each of them as a single neural network for standard verification.

\paragraph*{Specifications} The cersyve benchmark requires to verify that within a bounded state space, whether the constraint satisfaction and forward invariance properties of a neural safety certificate is satisfied for all states. For a detailed description of the verification problem formulation, please refer to our paper~\cite{yang2025scalable}. The input specifications are lower and upper bounds on each dimension of the state. The output specifications are always two inequalities corresponding to the two properties.

\paragraph*{Link} \url{https://github.com/intelligent-control-lab/cersyve.jl/tree/VNN-COMP-2025}

\subsection{Additional Benchmarks}

We have not yet obtained benchmark descriptions for other benchmark sets reported in Table \ref{tab:benchmarks}. We will update the report when these descriptions are available.
Artifacts of the benchmarks are available in the repository\footnote{\url{https://github.com/VNN-COMP/vnncomp2025_benchmarks}}. 

\newpage
\section{Results}
\label{sec:results}

Each tool was run on each of the benchmarks and produced a \texttt{csv} result file, that was provided as feedback to the tool authors using the online execution platform.
The final \texttt{csv} files for each tool as well as scoring scripts are available online: \url{https://github.com/VNN-COMP/vnncomp2025_results}.
The results were automatically analyzed to compute scores and create the statistics presented in this section. 

We consider two sets of results based on two different criteria of accepting counter-examples as valid ones. They are reported in Section~\ref{subsec:zerotol} and  Section~\ref{subsec:smalltol}, respectively. The former deems a counter-example valid if executing the original counter-example in the evaluation environment results in the violation of the specification. The latter allows a small tolerance (1e-6) when checking the violation of the specification. The latter criterion may accept more counter-examples. The reason to allow such tolerance is that we have discovered cases where replaying the same counter-examples on different hardware can result in different outputs due to floating-point rounding errors, and thus the counter-example may not violate the specification in the evaluation environment even though it does so during the verification process. This is an issue that we have not discovered in previous competitions and it calls for a discussion on how to handle it in future competitions.


%
%


\subsection{Zero Tolerance}
\label{subsec:zerotol}
\subsubsection{Regular Track}
The regular track contained all the benchmarks that were voted for by at least half of all participants.


\begin{table}[h]
\begin{center}
\caption{Overall Score} \label{tab:score}
{\setlength{\tabcolsep}{2pt}
\begin{tabular}[h]{@{}lll@{}}
\toprule
\textbf{\# ~} & \textbf{Tool} & \textbf{Score}\\
\midrule
1 & $\alpha$-$\beta$-CROWN & 1566.9 \\
2 & NeuralSAT & 1430.2 \\
3 & PyRAT & 1228.4 \\
4 & CORA & 987.2 \\
5 & NNV & 796.4 \\
6 & nnenum & 740.3 \\
7 & SobolBox & 593.3 \\
\bottomrule
\end{tabular}
}
\end{center}
\end{table}

\begin{figure}[h]
\centerline{\includegraphics[width=\textwidth]{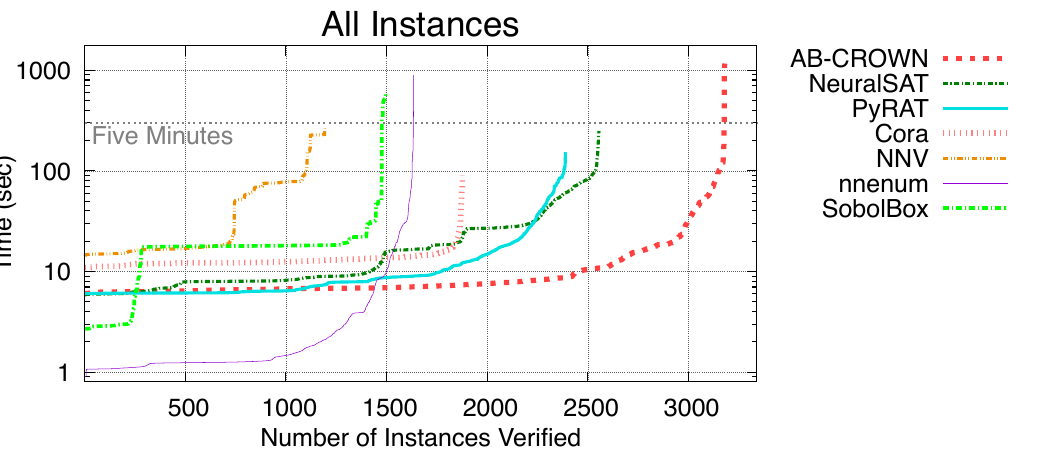}}
\caption{Cactus Plot for All Instances.}
\label{fig:quantPic}
\end{figure}

\begin{figure}[h]
\centerline{\includegraphics[width=\textwidth]{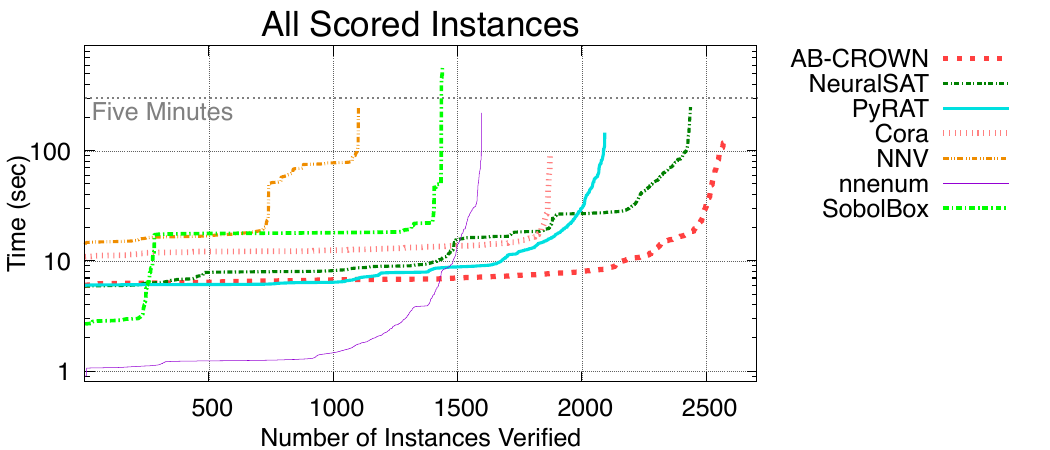}}
\caption{Cactus Plot for All Scored Instances.}
\label{fig:quantPic_as}
\end{figure}

\subsubsection{Extended Track}
All benchmarks that were voted for by at least one team and did not make it into the regular track were part of the extended track.
Every benchmark was voted for at least once, so no benchmark was unscored.

\clearpage

\subsubsection{Other Stats}

This section presents other statistics related to the measurements that are interesting but did not play a direct role in scoring this year.
The results reflect the tool performance on the regular track.



\begin{table}[h]
\begin{center}
\caption{Overhead} \label{tab:overhead}
{\setlength{\tabcolsep}{2pt}
\begin{tabular}[h]{@{}llr@{}}
\toprule
\textbf{\# ~} & \textbf{Tool} & \textbf{Seconds}\\
\midrule
1 & nnenum & 0.9 \\
2 & SobolBox & 2.7 \\
3 & NeuralSAT & 5.9 \\
4 & $\alpha$-$\beta$-CROWN & 6.0 \\
5 & PyRAT & 6.0 \\
6 & CORA & 10.9 \\
7 & NNV & 14.2 \\
\bottomrule
\end{tabular}
}
\end{center}
\end{table}


\begin{table}[h]
\begin{center}
\caption{Num Benchmarks Participated} \label{tab:stats0}
{\setlength{\tabcolsep}{2pt}
\begin{tabular}[h]{@{}llr@{}}
\toprule
\textbf{\# ~} & \textbf{Tool} & \textbf{Count}\\
\midrule
1 & NeuralSAT & 16 \\
2 & $\alpha$-$\beta$-CROWN & 16 \\
3 & PyRAT & 15 \\
4 & NNV & 15 \\
5 & CORA & 14 \\
6 & nnenum & 11 \\
7 & SobolBox & 9 \\
\bottomrule
\end{tabular}
}
\end{center}
\end{table}


\begin{table}[h]
\begin{center}
\caption{Num Instances Verified} \label{tab:stats1}
{\setlength{\tabcolsep}{2pt}
\begin{tabular}[h]{@{}llr@{}}
\toprule
\textbf{\# ~} & \textbf{Tool} & \textbf{Count}\\
\midrule
1 & $\alpha$-$\beta$-CROWN & 2575 \\
2 & NeuralSAT & 2437 \\
3 & PyRAT & 2092 \\
4 & CORA & 1872 \\
5 & nnenum & 1596 \\
6 & SobolBox & 1439 \\
7 & NNV & 1101 \\
\bottomrule
\end{tabular}
}
\end{center}
\end{table}


\begin{table}[h]
\begin{center}
\caption{Num SAT} \label{tab:stats2}
{\setlength{\tabcolsep}{2pt}
\begin{tabular}[h]{@{}llr@{}}
\toprule
\textbf{\# ~} & \textbf{Tool} & \textbf{Count}\\
\midrule
1 & $\alpha$-$\beta$-CROWN & 1082 \\
2 & NeuralSAT & 1055 \\
3 & PyRAT & 1018 \\
4 & CORA & 946 \\
5 & nnenum & 786 \\
6 & NNV & 354 \\
7 & SobolBox & 317 \\
\bottomrule
\end{tabular}
}
\end{center}
\end{table}


\begin{table}[h]
\begin{center}
\caption{Num UNSAT} \label{tab:stats3}
{\setlength{\tabcolsep}{2pt}
\begin{tabular}[h]{@{}llr@{}}
\toprule
\textbf{\# ~} & \textbf{Tool} & \textbf{Count}\\
\midrule
1 & $\alpha$-$\beta$-CROWN & 1493 \\
2 & NeuralSAT & 1382 \\
3 & SobolBox & 1122 \\
4 & PyRAT & 1074 \\
5 & CORA & 926 \\
6 & nnenum & 810 \\
7 & NNV & 747 \\
\bottomrule
\end{tabular}
}
\end{center}
\end{table}


\begin{table}[h]
\begin{center}
\caption{Incorrect Results (or Missing CE)} \label{tab:stats4}
{\setlength{\tabcolsep}{2pt}
\begin{tabular}[h]{@{}llr@{}}
\toprule
\textbf{\# ~} & \textbf{Tool} & \textbf{Count}\\
\midrule
1 & SobolBox & 55 \\
2 & NeuralSAT & 5 \\
3 & CORA & 2 \\
\bottomrule
\end{tabular}
}
\end{center}
\end{table}

\clearpage
\subsection{Small Tolerance}
\label{subsec:smalltol}
\subsubsection{Regular Track}
The regular track contained all the benchmarks that were voted for by at least half of all participants.


\begin{table}[h]
\begin{center}
\caption{Overall Score} \label{tab:score_tol}
{\setlength{\tabcolsep}{2pt}
\begin{tabular}[h]{@{}lll@{}}
\toprule
\textbf{\# ~} & \textbf{Tool} & \textbf{Score}\\
\midrule
1 & $\alpha$-$\beta$-CROWN & 1600.0 \\
2 & NeuralSAT & 1360.1 \\
3 & PyRAT & 1218.5 \\
4 & CORA & 954.6 \\
5 & nnenum & 740.8 \\
6 & NNV & 697.3 \\
7 & SobolBox & 529.0 \\
\bottomrule
\end{tabular}
}
\end{center}
\end{table}

\begin{figure}[h]
\centerline{\includegraphics[width=\textwidth]{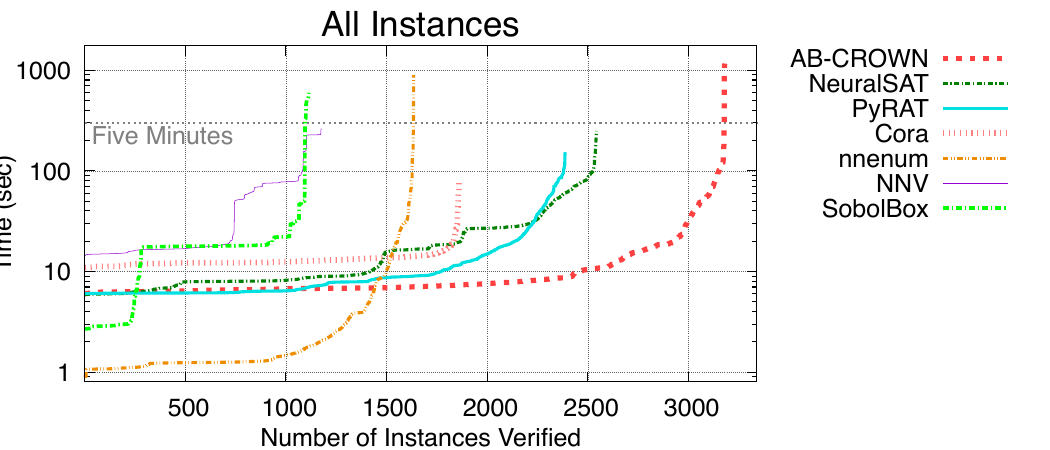}}
\caption{Cactus Plot for All Instances.}
\label{fig:quantPic_tol}
\end{figure}

\begin{figure}[h]
\centerline{\includegraphics[width=\textwidth]{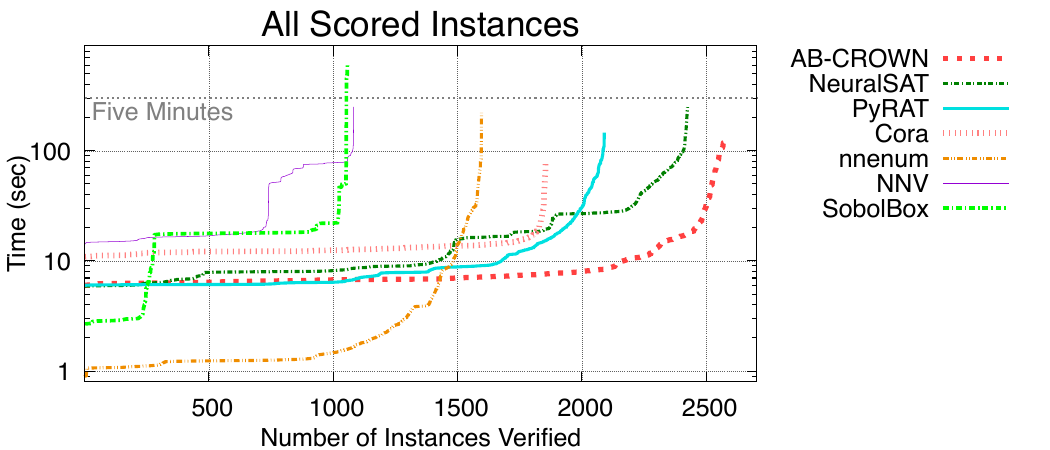}}
\caption{Cactus Plot for All Scored Instances.}
\label{fig:quantPic_tol_as} 
\end{figure}

\subsubsection{Extended Track}
All benchmarks that were voted for by at least one team and did not make it into the regular track were part of the extended track.
Every benchmark was voted for at least once, so no benchmark was unscored.

\clearpage

\subsubsection{Other Stats}

This section presents other statistics related to the measurements that are interesting but did not play a direct role in scoring this year.
The results reflect the tool performance on the regular track.



\begin{table}[h]
\begin{center}
\caption{Overhead} \label{tab:overhead_tol}
{\setlength{\tabcolsep}{2pt}
\begin{tabular}[h]{@{}llr@{}}
\toprule
\textbf{\# ~} & \textbf{Tool} & \textbf{Seconds}\\
\midrule
1 & nnenum & 0.9 \\
2 & SobolBox & 2.7 \\
3 & NeuralSAT & 5.9 \\
4 & $\alpha$-$\beta$-CROWN & 6.0 \\
5 & PyRAT & 6.0 \\
6 & CORA & 10.9 \\
7 & NNV & 14.2 \\
\bottomrule
\end{tabular}
}
\end{center}
\end{table}


\begin{table}[h]
\begin{center}
\caption{Num Benchmarks Participated} \label{tab:stats0_tol}
{\setlength{\tabcolsep}{2pt}
\begin{tabular}[h]{@{}llr@{}}
\toprule
\textbf{\# ~} & \textbf{Tool} & \textbf{Count}\\
\midrule
1 & NeuralSAT & 16 \\
2 & $\alpha$-$\beta$-CROWN & 16 \\
3 & PyRAT & 15 \\
4 & NNV & 15 \\
5 & CORA & 14 \\
6 & nnenum & 11 \\
7 & SobolBox & 9 \\
\bottomrule
\end{tabular}
}
\end{center}
\end{table}


\begin{table}[h]
\begin{center}
\caption{Num Instances Verified} \label{tab:stats1_tol}
{\setlength{\tabcolsep}{2pt}
\begin{tabular}[h]{@{}llr@{}}
\toprule
\textbf{\# ~} & \textbf{Tool} & \textbf{Count}\\
\midrule
1 & $\alpha$-$\beta$-CROWN & 2575 \\
2 & NeuralSAT & 2425 \\
3 & PyRAT & 2090 \\
4 & CORA & 1854 \\
5 & nnenum & 1596 \\
6 & NNV & 1082 \\
7 & SobolBox & 1057 \\
\bottomrule
\end{tabular}
}
\end{center}
\end{table}


\begin{table}[h]
\begin{center}
\caption{Num SAT} \label{tab:stats2_tol}
{\setlength{\tabcolsep}{2pt}
\begin{tabular}[h]{@{}llr@{}}
\toprule
\textbf{\# ~} & \textbf{Tool} & \textbf{Count}\\
\midrule
1 & $\alpha$-$\beta$-CROWN & 1082 \\
2 & NeuralSAT & 1055 \\
3 & PyRAT & 1018 \\
4 & CORA & 946 \\
5 & nnenum & 786 \\
6 & NNV & 354 \\
7 & SobolBox & 317 \\
\bottomrule
\end{tabular}
}
\end{center}
\end{table}


\begin{table}[h]
\begin{center}
\caption{Num UNSAT} \label{tab:stats3_tol}
{\setlength{\tabcolsep}{2pt}
\begin{tabular}[h]{@{}llr@{}}
\toprule
\textbf{\# ~} & \textbf{Tool} & \textbf{Count}\\
\midrule
1 & $\alpha$-$\beta$-CROWN & 1493 \\
2 & NeuralSAT & 1370 \\
3 & PyRAT & 1072 \\
4 & CORA & 908 \\
5 & nnenum & 810 \\
6 & SobolBox & 740 \\
7 & NNV & 728 \\
\bottomrule
\end{tabular}
}
\end{center}
\end{table}


\begin{table}[h]
\begin{center}
\caption{Incorrect Results (or Missing CE)} \label{tab:stats4_tol}
{\setlength{\tabcolsep}{2pt}
\begin{tabular}[h]{@{}llr@{}}
\toprule
\textbf{\# ~} & \textbf{Tool} & \textbf{Count}\\
\midrule
1 & SobolBox & 437 \\
2 & CORA & 20 \\
3 & NNV & 19 \\
4 & NeuralSAT & 17 \\
5 & PyRAT & 2 \\
\bottomrule
\end{tabular}
}
\end{center}
\end{table}

\section{Conclusion and Ideas for Future Competitions}
\label{sec:conclusion}
This report summarizes the 6$^\text{th}$ Verification of Neural Networks Competition (VNN-COMP), held in 2025.
While we observed a significant increase in the diversity, complexity, and scale of the proposed benchmarks, the best-performing tools seem to converge to GPU-enabled linear bound propagation methods using a branch-and-bound framework.
In addition to the standardization of input formats (\texttt{onnx} and \texttt{vnnlib}) and evaluation hardware, introduced for VNN-COMP 2021, VNN-COMP 2025 also continued the standardized format for counter-examples and fully automated evaluation pipeline introduced in VNN-COMP 2022, requiring authors to provide complete installation scripts.
We hope that this increased standardization and automatization does not only simplify the evaluation during the competition but also enables practitioners and researchers to more easily apply a range of state-of-the-art verification methods to their individual problems.

VNN-COMP 2025 successfully implemented a range of improvement opportunities identified during the previous iteration. These included requiring witnesses of found counter-examples to disambiguate tool disagreement, increasing automatization to enable a smoother final evaluation, and making a broader range of AWS instances available to allow for a better fit with tools' requirements. Additionally, we introduced the Competition Contribution papers included in the SAIV proceedings~\cite{saiv2025} for participants to summarize their approaches.
However, some issues were identified in the evaluation particularly with respect to the criteria of accepting a counter-example as valid, as we observed cases where a counter-example violates the property on some hardware but does not violate the property on some other. This calls for a discussion of how to handle such cases for future iterations. In addition, this might also motivate research to find counter-examples that are robust across different hardware platforms. 
In addition, in cases where the answer to a benchmark is known a priori, we will explore using the known correct results as the ground truth.
Further ideas for future competitions include the use of scored benchmarks specifically designed for year-on-year progress tracking, the reduction of tool tuning, a batch-processing mode, and more rigorous soundness evaluation. Finally, as research on proof generation for neural network verification grows~\cite{isac2022neural,elboher2025abstraction,duong2025generating}, a longer term goal is to require solvers to produce UNSAT proofs.

\section*{Acknowledgements}
%

The 2025 competition was supported by CEA-List and MathWorks, which provided funds for computation costs.

Additionally, this research was supported in part by the Air Force Research Laboratory Information Directorate, through the Air Force Office of Scientific Research Summer Faculty Fellowship
Program, Contract Numbers FA8750-15-3-6003, FA9550-15-0001 and FA9550-20-F-0005.
This material is based upon work supported by the Air Force Office of Scientific Research under award numbers FA9550-19-1-0288, FA9550-21-1-0121, and FA9550-22-1-0019, the National Science Foundation (NSF) under grant numbers 2107035, 2220401, 2220418, 2220426, and 2238133, and the Defense Advanced Research Projects Agency (DARPA) Assured Neuro Symbolic Learning and Reasoning (ANSR)
programs through contract number and FA8750-23-C-0518.
Any opinions, findings, and conclusions or recommendations expressed in this material are those of the author(s) and do not necessarily reflect the views of the United States Air Force, DARPA, nor NSF.
This research was additionally supported by a
grant from the project FAI (No. 286525601) funded
by the German Research Foundation (Deutsche Forschungsgemeinschaft, DFG).

Tool and benchmark authors listed in \Cref{sec:participants} and \Cref{sec:benchmarks} participated in the preparation and review of this report.

\clearpage
\label{sect:bib}
\bibliographystyle{plain}
\bibliography{bib/nnv, bib/nnenum, bib/peregriNN, bib/verinet, bib/oval,bib/venus,bib/MIPVerify, bib/mnbab, bib/alpha-beta-CROWN, bib/collins,bib/dnnf,bib/nvjl,bib/nn4sys,bib/Marabou,bib/RPM,bib/AVeriNN,bib/VeRAPAk,bib/general,bib/traffic-signs-recognition,bib/neuralsat, bib/pyrat, bib/vit, bib/cora, bib/never2, bib/safeNLP, bib/SobolBox, bib/relusplitter,bib/metaroom,bib/cersyve}

@inproceedings{Tjeng2019EvaluatingRO,
  title={Evaluating Robustness of Neural Networks with Mixed Integer Programming},
  author={Vincent Tjeng and Kai Y. Xiao and Russ Tedrake},
  booktitle={ICLR},
  year={2019}
}

@article{wu2022toward,
  title={Toward Certified Robustness Against Real-World Distribution Shifts},
  author={Wu, Haoze and Tagomori, Teruhiro and Robey, Alexander and Yang, Fengjun and Matni, Nikolai and Pappas, George and Hassani, Hamed and Pasareanu, Corina and Barrett, Clark},
  journal={arXiv preprint arXiv:2206.03669},
  year={2022}
}

@inproceedings{isac2022neural,
  author       = {Omri Isac and
                  Clark W. Barrett and
                  Min Zhang and
                  Guy Katz},
  editor       = {Alberto Griggio and
                  Neha Rungta},
  title        = {Neural Network Verification with Proof Production},
  booktitle    = {22nd Formal Methods in Computer-Aided Design, {FMCAD} 2022, Trento,
                  Italy, October 17-21, 2022},
  pages        = {38--48},
  publisher    = {{IEEE}},
  year         = {2022},
  url          = {https://doi.org/10.34727/2022/isbn.978-3-85448-053-2\_9},
  doi          = {10.34727/2022/ISBN.978-3-85448-053-2\_9},
}

@inproceedings{elboher2025abstraction,
  title={Abstraction-Based Proof Production in Formal Verification of Neural Networks},
  author={Elboher, Yizhak Yisrael and Isac, Omri and Katz, Guy and Ladner, Tobias and Wu, Haoze},
  booktitle={International Symposium on AI Verification},
  pages={203--220},
  year={2025},
  organization={Springer}
}

@misc{sobolbox,
  title = {{SobolBox}},
  howpublished = {\url{https://github.com/dassarthak18/SobolBox}},
  note = {VNNCOMP-2025 branch}
}

@misc{innvers,
  title = {{IACS Neural Network Verification System}},
  howpublished = {\url{https://github.com/iacs-csu-2020/INNVerS}}
}

@article{sobol,
  title={{O}n the {D}istribution of {P}oints in a {C}ube and the {A}pproximate {E}valuation of {I}ntegrals},
  author={Ilya M. Sobol'},
  journal={Ussr Computational Mathematics and Mathematical Physics},
  year={1967},
  volume={7},
  pages={86-112},
  doi={10.1016/0041-5553(67)90144-9}
}

@article{lbfgsb,
author = {Byrd, Richard H. and Lu, Peihuang and Nocedal, Jorge and Zhu, Ciyou},
title = {{A} {L}imited {M}emory {A}lgorithm for {B}ound {C}onstrained {O}ptimization},
journal = {SIAM Journal on Scientific Computing},
volume = {16},
number = {5},
pages = {1190-1208},
year = {1995},
doi = {10.1137/0916069}
}

@article{advi,
author = {Kucukelbir, Alp and Tran, Dustin and Ranganath, Rajesh and Gelman, Andrew and Blei, David M.},
title = {Automatic differentiation variational inference},
year = {2017},
issue_date = {January 2017},
publisher = {JMLR.org},
volume = {18},
number = {1},
issn = {1532-4435},
journal = {J. Mach. Learn. Res.},
month = jan,
pages = {430–474},
numpages = {45}
}

@article{li2025two,
  title={Two-Stage Learning of Stabilizing Neural Controllers via Zubov Sampling and Iterative Domain Expansion},
  author={Li, Haoyu and Zhong, Xiangru and Hu, Bin and Zhang, Huan},
  journal={arXiv preprint arXiv:2506.01356},
  year={2025}
}

@inproceedings{zhou2024scalable,
  title={Scalable Neural Network Verification with Branch-and-bound Inferred Cutting Planes},
  author={Zhou, Duo and Brix, Christopher and Hanasusanto, Grani A and Zhang, Huan},
  booktitle={The Thirty-eighth Annual Conference on Neural Information Processing Systems},
  year={2024}
}

@article{shi2024genbab,
  title={Neural Network Verification with Branch-and-Bound for General Nonlinearities},
  author={Shi, Zhouxing and Jin, Qirui and Kolter, Zico and Jana, Suman and Hsieh, Cho-Jui and Zhang, Huan},
  journal={arXiv preprint arXiv:2405.21063},
  year={2024}
}

@article{zhang2022general,
  title={General cutting planes for bound-propagation-based neural network verification},
  author={Zhang*, Huan and Wang*, Shiqi and Xu*, Kaidi and Li, Linyi and Li, Bo and Jana, Suman and Hsieh, Cho-Jui and Kolter, J Zico},
  journal={Advances in Neural Information Processing Systems (NeurIPS)},
  year={2022}
}

@article{madry2017towards,
  title={Towards deep learning models resistant to adversarial attacks},
  author={Madry, Aleksander and Makelov, Aleksandar and Schmidt, Ludwig and Tsipras, Dimitris and Vladu, Adrian},
  journal={arXiv preprint arXiv:1706.06083},
  year={2017}
}

@inproceedings{xu2021fast,
    title={{Fast and Complete}: Enabling Complete Neural Network Verification with Rapid and Massively Parallel Incomplete Verifiers},
    author={Kaidi Xu and Huan Zhang and Shiqi Wang and Yihan Wang and Suman Jana and Xue Lin and Cho-Jui Hsieh},
    booktitle={International Conference on Learning Representations},
    year={2021},
    url={https://openreview.net/forum?id=nVZtXBI6LNn}
}

@article{wang2021betacrown,
  title={{Beta-CROWN}: Efficient Bound Propagation with Per-neuron Split Constraints for Complete and Incomplete Neural Network Verification},
  author={Wang, Shiqi and Zhang, Huan and Xu, Kaidi and Lin, Xue and Jana, Suman and Hsieh, Cho-Jui and Kolter, Zico},
  journal={arXiv preprint arXiv:2103.06624},
  year={2021}
}

@article{zhang2018efficient,
  title={Efficient Neural Network Robustness Certification with General Activation Functions},
  author={Zhang, Huan and Weng, Tsui-Wei and Chen, Pin-Yu and Hsieh, Cho-Jui and Daniel, Luca},
  journal={Advances in Neural Information Processing Systems},
  volume={31},
  pages={4939--4948},
  year={2018},
  url={https://arxiv.org/pdf/1811.00866.pdf}
}

@article{xu2020automatic,
  title={Automatic perturbation analysis for scalable certified robustness and beyond},
  author={Xu, Kaidi and Shi, Zhouxing and Zhang, Huan and Wang, Yihan and Chang, Kai-Wei and Huang, Minlie and Kailkhura, Bhavya and Lin, Xue and Hsieh, Cho-Jui},
  journal={Advances in Neural Information Processing Systems},
  volume={33},
  year={2020}
}

@article{bunelunified2018,
  title={A Unified View of Piecewise Linear Neural Network Verification},
  author={Bunel, Rudy and Turkaslan, Ilker and Torr, Philip HS and Kohli, Pushmeet and Kumar, M Pawan},
  journal={Advances in Neural Information Processing Systems},
  year={2018}
}

@article{zhou2024testing,
  title={Testing Neural Network Verifiers: A Soundness Benchmark with Hidden Counterexamples},
  author={Zhou, Xingjian and Xu, Hongji and Xu, Andy and Shi, Zhouxing and Hsieh, Cho-Jui and Zhang, Huan},
  journal={arXiv preprint arXiv:2412.03154},
  year={2024}
}

@article{shi2024certified,
  title={Certified Training with Branch-and-Bound: A Case Study on Lyapunov-stable Neural Control},
  author={Shi, Zhouxing and Hsieh, Cho-Jui and Zhang, Huan},
  journal={arXiv preprint arXiv:2411.18235},
  year={2024}
}

@inproceedings{yang2024lyapunov,
  title={Lyapunov-stable Neural Control for State and Output Feedback: A Novel Formulation},
  author={Yang, Lujie and Dai, Hongkai and Shi, Zhouxing and Hsieh, Cho-Jui and Tedrake, Russ and Zhang, Huan},
  booktitle={Forty-first International Conference on Machine Learning},
  year={2024}
}

@article{yang2025scalable,
  title={Scalable synthesis of formally verified neural value function for hamilton-jacobi reachability analysis},
  author={Yang, Yujie and Hu, Hanjiang and Wei, Tianhao and Li, Shengbo Eben and Liu, Changliu},
  journal={Journal of Artificial Intelligence Research},
  volume={83},
  year={2025}
}

@inproceedings{kirov2023formal,
  title={Formal verification of a neural network based prognostics system for aircraft equipment},
  author={Kirov, Dmitrii and Rollini, Simone Fulvio and Di Guglielmo, Luigi and Cofer, Darren},
  booktitle={International Conference on Bridging the Gap between AI and Reality},
  pages={225--240},
  year={2023},
  organization={Springer}
}

@inproceedings{kirov2023benchmark,
  title={Benchmark: remaining useful life predictor for aircraft equipment},
  author={Kirov, Dmitrii and Rollini, Simone Fulvio},
  booktitle={International Conference on Bridging the Gap between AI and Reality},
  pages={299--304},
  year={2023},
  organization={Springer}
}

@techreport{ForMuLA,
	author = {{EASA and Collins Aerospace}},
	title = {{Formal Methods use for Learning Assurance (ForMuLA)}},
	month = {April},
	year = {2023}
}

@inproceedings{althoff_2015,
  author    = {Althoff, Matthias},
  booktitle = {Proc. of the Workshop on Applied Verification for Continuous and Hybrid Systems (ARCH)},
  pages     = {120--151},
  title     = {An Introduction to {CORA} 2015},
  year      = {2015}
}

@inproceedings{kochdumper_et_al_2023,
  author    = {Kochdumper, Niklas and Schilling, Christian and Althoff, Matthias and Bak, Stanley},
  booktitle = {NASA Formal Methods},
  pages     = {16--36},
  title     = {Open- and Closed-Loop Neural Network Verification Using Polynomial Zonotopes},
  year      = {2023}
}

@inproceedings{ladner_althoff_2023,
  author    = {Ladner, Tobias and Althoff, Matthias},
  booktitle = {Proc. of the Int. Conf. on Hybrid Systems: Computation and Control (HSCC)},
  pages     = {1--13},
  title     = {Automatic Abstraction Refinement in Neural Network Verification Using Sensitivity Analysis},
  year      = {2023}
}

@article{koller_et_al_2025settraining,
  title={Set-Based Training for Neural Network Verification},
  author={Koller, Lukas and Ladner, Tobias and Althoff, Matthias},
  journal={TMLR},
  year={2025}
}

@article{wendl2024training,
  title={Training verifiably robust agents using set-based reinforcement learning},
  author={Wendl, Manuel and Koller, Lukas and Ladner, Tobias and Althoff, Matthias},
  journal={arXiv preprint arXiv:2408.09112},
  year={2024}
}

@article{koller_et_al_2025shadows,
    title={Out of the Shadows: Exploring a Latent Space for Neural Network Verification}, 
    author={Lukas Koller and Tobias Ladner and Matthias Althoff},
    journal={arXiv},
    year={2025}
}

@inproceedings{FerlezKS22,
  author    = {James Ferlez and
               Haitham Khedr and
               Yasser Shoukry},
  editor    = {Ezio Bartocci and
               Sylvie Putot},
  title     = {Fast {BATLLNN:} Fast Box Analysis of Two-Level Lattice Neural Networks},
  booktitle = {{HSCC} '22: 25th {ACM} International Conference on Hybrid Systems:
               Computation and Control, Milan, Italy, May 4 - 6, 2022},
  pages     = {23:1--23:11},
  publisher = {{ACM}},
  year      = {2022},
  url       = {https://doi.org/10.1145/3501710.3519533},
  doi       = {10.1145/3501710.3519533},
  bibsource = {dblp computer science bibliography, https://dblp.org}
}

@inproceedings{vnnlib,
  author    = {Stefano Demarchi and Dario Guidotti and Luca Pulina and Armando Tacchella},
  title     = {Supporting Standardization of Neural Networks Verification with VNNLIB and CoCoNet},
  booktitle = {Proceedings of the 6th Workshop on Formal Methods for ML-Enabled Autonomous Systems},
  editor    = {Nina Narodytska and Guy Amir and Guy Katz and Omri Isac},
  series    = {Kalpa Publications in Computing},
  volume    = {16},
  pages     = {47--58},
  year      = {2023},
  publisher = {EasyChair},
  bibsource = {EasyChair, https://easychair.org},
  issn      = {2515-1762},
  url       = {https://easychair.org/publications/paper/Qgdn},
  doi       = {10.29007/5pdh}}

@misc{bak2021vnncomp,
      title={The Second International Verification of Neural Networks Competition (VNN-COMP 2021): Summary and Results}, 
      author={Stanley Bak and Changliu Liu and Taylor Johnson},
      year={2021},
      journal={arXiv preprint arXiv:2109.00498},
}

@article{muller2022vnncomp,
  title={The third international verification of neural networks competition (VNN-COMP 2022): Summary and results},
  author={M{\"u}ller, Mark Niklas and Brix, Christopher and Bak, Stanley and Liu, Changliu and Johnson, Taylor T},
  journal={arXiv preprint arXiv:2212.10376},
  year={2022}
}

@article{brix2023vnncomp,
      title={The Fourth International Verification of Neural Networks Competition (VNN-COMP 2023): Summary and Results}, 
      author={Christopher Brix and Stanley Bak and Changliu Liu and Taylor T. Johnson},
      year={2023},
      eprint={arXiv preprint arXiv:2312.16760},
}

@article{brix2024vnncomp,
      title={The Fifth International Verification of Neural Networks Competition ({VNN-COMP} 2024): Summary and Results}, 
      author={Christopher Brix and Stanley Bak and Taylor T. Johnson and Haoze Wu},
      year={2024},
      eprint={2412.19985},
      archivePrefix={arXiv},
      primaryClass={cs.LG},
      url={https://arxiv.org/abs/2412.19985}, 
}

@proceedings{saiv2025,
title = {AI Verification: Second International Symposium, SAIV 2025, Zagreb, Croatia, July 21–22, 2025, Proceedings},
year = {2025},
isbn = {978-3-031-99990-1},
publisher = {Springer-Verlag},
address = {Berlin, Heidelberg},
location = {Zagreb, Croatia}
}

@proceedings{cav2025,
title = {Computer Aided Verification: 37th International Conference, CAV 2025, Zagreb, Croatia, July 23-25, 2025, Proceedings, Part II},
year = {2025},
isbn = {978-3-031-98678-9},
publisher = {Springer-Verlag},
address = {Berlin, Heidelberg},
location = {Zagreb, Croatia}
}

@InProceedings{duong2025saiv_cc,
author="Duong, Hai
and Nguyen, ThanhVu",
editor="Giacobbe, Mirco
and Lukina, Anna",
title="NeuralSAT: Scaling Constraint Solving for DNN Verification (Competition Contribution)",
booktitle="AI Verification",
year="2026",
publisher="Springer Nature Switzerland",
address="Cham",
pages="253--259",
isbn="978-3-031-99991-8"
}

@InProceedings{lopez2025saiv_cc,
author="Lopez, Diego Manzanas
and Sasaki, Samuel
and Johnson, Taylor T.",
editor="Giacobbe, Mirco
and Lukina, Anna",
title="NNV: A Star Set Reachability Approach (Competition Contribution)",
booktitle="AI Verification",
year="2026",
publisher="Springer Nature Switzerland",
address="Cham",
pages="260--265",
isbn="978-3-031-99991-8"
}

@InProceedings{lemesle2025saiv_cc,
author="Lemesle, Augustin
and Lehmann, Julien
and Le Gall, Tristan
and Chihani, Zakaria",
editor="Giacobbe, Mirco
and Lukina, Anna",
title="PyRAT: Verifying Neural Networks with Abstract Interpretation (Competition Contribution)",
booktitle="AI Verification",
year="2026",
publisher="Springer Nature Switzerland",
address="Cham",
pages="266--271",
isbn="978-3-031-99991-8"
}

@InProceedings{das2025saiv_cc,
author="Das, Sarthak",
editor="Giacobbe, Mirco
and Lukina, Anna",
title="SobolBox: Boxed Refinement of Sobol Sequence Samples for Neural Network Verification (Competition Contribution)",
booktitle="AI Verification",
year="2026",
publisher="Springer Nature Switzerland",
address="Cham",
pages="272--277",
isbn="978-3-031-99991-8"
}

@inproceedings{hu2022robustness,
  title={Robustness Certification of Visual Perception Models via Camera Motion Smoothing},
  author={Hu, Hanjiang and Liu, Zuxin and Li, Linyi and Zhu, Jiacheng and Zhao, Ding},
  booktitle={Conference on Robot Learning},
  pages={1309--1320},
  year={2022},
  organization={PMLR}
}

@inproceedings{hu2024pixel,
  title={Pixel-wise Smoothing for Certified Robustness against Camera Motion Perturbations},
  author={Hu, Hanjiang and Liu, Zuxin and Li, Linyi and Zhu, Jiacheng and Zhao, Ding},
  booktitle={International Conference on Artificial Intelligence and Statistics},
  pages={217--225},
  year={2024},
  organization={PMLR}
}

@inproceedings{hu2023robustness,
  title={Robustness verification for perception models against camera motion perturbations},
  author={Hu, Hanjiang and Liu, Changliu and Zhao, Ding},
  booktitle={International conference on machine learning (ICML) Workshop on Formal Verification of Machine Learning (WFVML)},
  year={2023}
}

@misc{duong2023dpllt,
      title={{A DPLL(T) Framework for Verifying Deep Neural Networks}}, 
      author={Hai Duong and Linhan Li and ThanhVu Nguyen and Matthew Dwyer},
      year={2023},
      note={arXiv, 25 pages},
      eprint={2307.10266},
      archivePrefix={arXiv},
      primaryClass={cs.LG}
}

@article{duong2024harnessing,
  title={Harnessing neuron stability to improve dnn verification},
  author={Duong, Hai and Xu, Dong and Nguyen, ThanhVu and Dwyer, Matthew B},
  journal={Proceedings of the ACM on Software Engineering},
  volume={1},
  number={FSE},
  pages={859--881},
  year={2024},
  publisher={ACM New York, NY, USA}
}

@inproceedings{duong2025neuralsat,
  title={NeuralSAT: A High-Performance Verification Tool for Deep Neural Networks},
  author={Duong, Hai and Nguyen, ThanhVu and Dwyer, Matthew B},
  booktitle={International Conference on Computer Aided Verification},
  pages={409--423},
  year={2025},
  organization={Springer}
}

@inproceedings{duong2025neuralsat2,
  title={NeuralSAT: Scaling Constraint Solving for DNN Verification (Competition Contribution)},
  author={Duong, Hai and Nguyen, ThanhVu},
  booktitle={International Symposium on AI Verification},
  pages={253--259},
  year={2025},
  organization={Springer}
}

@article{duong2025compositional,
  title={Compositional Neural Network Verification via Assume-Guarantee Reasoning},
  author={Duong, Hai and Shriver, David and Nguyen, ThanhVu and Dwyer, Matthew},
  journal={Advances in Neural Information Processing Systems},
  volume={},
  pages={to appear},
  year={2025}
}

@article{duong2025generating,
  title={Generating and Checking DNN Verification Proofs},
  author={Duong, Hai and Nguyen, ThanhVu and Dwyer, Matthew},
  journal={Advances in Neural Information Processing Systems},
  volume={},
  pages={to appear},
  year={2025}
}

@inproceedings{kraska18case,
  title={The case for learned index structures},
  author={Kraska, Tim and Beutel, Alex and Chi, Ed H and Dean, Jeffrey and Polyzotis, Neoklis},
  booktitle={Proceedings of the 2018 International Conference on Management of Data},
  year={2018}
}

@inproceedings{bak2021nnenum,
  title={nnenum: Verification of ReLU Neural Networks with Optimized Abstraction Refinement},
  author={Bak, Stanley},
  booktitle={NASA Formal Methods Symposium},
  pages={19--36},
  year={2021},
  organization={Springer}
}

@article{simonyan2014very,
  title={Very deep convolutional networks for large-scale image recognition},
  author={Simonyan, Karen and Zisserman, Andrew},
  journal={arXiv preprint arXiv:1409.1556},
  year={2014}
}

@misc{bak2020vnn,
  title={Execution-Guided Overapproximation (EGO) for Improving Scalability of Neural Network Verification},
  author={Stanley Bak},
  booktitle={3rd International Workshop on Verification of Neural Networks},
  year={2020}
}

@inproceedings{katz2017reluplex,
  title={Reluplex: An efficient SMT solver for verifying deep neural networks},
  author={Katz, Guy and Barrett, Clark and Dill, David L and Julian, Kyle and Kochenderfer, Mykel J},
  booktitle={International Conference on Computer Aided Verification},
  pages={97--117},
  year={2017},
  organization={Springer}
}

@inproceedings{katz2019marabou,
  title={The marabou framework for verification and analysis of deep neural networks},
  author={Katz, Guy and Huang, Derek A and Ibeling, Duligur and Julian, Kyle and Lazarus, Christopher and Lim, Rachel and Shah, Parth and Thakoor, Shantanu and Wu, Haoze and Zelji{\'c}, Aleksandar and others},
  booktitle={International Conference on Computer Aided Verification},
  pages={443--452},
  year={2019},
  organization={Springer}
}

@misc{katz2021veri,
      title={Verification of Image-based Neural Network Controllers Using Generative Models}, 
      author={Sydney M. Katz and Anthony L. Corso and Christopher A. Strong and Mykel J. Kochenderfer},
      year={2021},
      eprint={2105.07091},
      archivePrefix={arXiv},
      primaryClass={cs.LG},
      url={https://arxiv.org/abs/2105.07091}, 
}

@inbook{ArjomandBigdeli2024,
   title={Verification of Neural Network Control Systems in Continuous Time},
   ISBN={9783031651120},
   ISSN={1611-3349},
   url={http://dx.doi.org/10.1007/978-3-031-65112-0_5},
   DOI={10.1007/978-3-031-65112-0_5},
   booktitle={AI Verification},
   publisher={Springer Nature Switzerland},
   author={ArjomandBigdeli, Ali and Mata, Andrew and Bak, Stanley},
   year={2024},
   pages={100–115} }

@article{hashemi2025neurips,
  title={Scaling Data-Driven Probabilistic Robustness Analysis for Semantic Segmentation Neural Networks},
  author={Navid Hashemi and Samuel Sasaki · Ipek Oguz and Meiyi Ma and Taylor Johnsonk},
  journal={NeurIPS},
  year={2025}
}

@InProceedings{tran2020cav,
  author = "Hoang-Dung Tran and Stanley Bak and Weiming Xiang and Taylor T. Johnson",
  title = "Verification of Deep Convolutional Neural Networks Using ImageStars",
  booktitle = "32nd International Conference on Computer-Aided Verification (CAV)",
  year = "2020",
  month = "July",
  publisher = "Springer",
}

@inproceedings{tran2020cav_tool,
author = "Hoang-Dung Tran and Xiaodong Yang and Diego Manzanas Lopez and Patrick Musau and Luan Viet Nguyen and Weiming Xiang and Stanley Bak and Taylor T. Johnson",
title = "{NNV}: The Neural Network Verification Tool for Deep Neural Networks and Learning-Enabled Cyber-Physical Systems",
booktitle = "32nd International Conference on Computer-Aided Verification (CAV)",
year = "2020",
month = "July",
}

@inproceedings{bak2020cav,
author = "Stanley Bak and Hoang-Dung Tran and Kerianne Hobbs and Taylor T. Johnson",
title = "Improved Geometric Path Enumeration for Verifying {ReLU} Neural Networks",
booktitle = "32nd International Conference on Computer-Aided Verification (CAV)",
year = "2020",
month = "July",
}

@InProceedings{tran2019emsoft,
author="Tran, Hoang-Dung
and Feiyang Cei 
and Diego Manzanas Lopez 
and Taylor T. Johnson
and Xenofon Koutsoukos",
title="Safety Verification of Cyber-Physical Systems with Reinforcement Learning Control",
booktitle="ACM SIGBED International Conference on Embedded Software (EMSOFT'19)",
year="2019",
month="October",
publisher="ACM",
}

@InProceedings{tran2019fm,
author="Tran, Hoang-Dung
and Musau, Patrick
and Diego Manzanas Lopez 
and Xiaodong Yang 
and Nguyen, Luan Viet
and Xiang, Weiming
and Taylor T. Johnson",
editor="",
title="Star-Based Reachability Analysis for Deep Neural Networks",
booktitle="23rd International Symposium on Formal Methods (FM'19)",
year="2019",
month="October",
publisher="Springer International Publishing",
}

@inproceedings{manzanas2023cav,
author = "Diego Manzanas Lopez and Sung Woo Choi and Hoang-Dung Tran and Taylor T. Johnson",
title = "{NNV 2.0}: The Neural Network Verification Tool",
booktitle = "35th International Conference on Computer-Aided Verification (CAV)",
year = "2023",
month = "July"
}

@InProceedings{tran2021cav,
  author = "Hoang-Dung Tran and Neelanjana Pal and Patrick Musau and Xiaodong Yang and Nathaniel P. Hamilton and Diego Manzanas Lopez and Stanley Bak and Taylor T. Johnson",
  title = "Robustness Verification of Semantic Segmentation Neural Networks using Relaxed Reachability",
  booktitle = "33rd International Conference on Computer-Aided Verification (CAV)",
  year = "2021",
  month = "July",
  publisher = "Springer"
}

@InProceedings{tran2023hscc,
author = {Hoang Dung Tran and SungWoo Choi and Tomoya Yamaguchi and Bardh Hoxha and Danil Prokhorov},
title = {Verification of Recurrent Neural Networks using Star Reachability},
booktitle = {The 26th ACM International Conference on Hybrid Systems: Computation and Control (HSCC)},
year = {2023},
month = {May}
}

@inproceedings{manzanas2022formats,
author = {Manzanas Lopez, Diego and Musau, Patrick and Hamilton, Nathaniel and Johnson, Taylor},
title = {Reachability Analysis of a General Class of Neural Ordinary Differential Equation},
year = {2022},
month = {September},
address = {Warsaw, Poland},
booktitle = {Proceedings of the 20th International Conference on Formal Modeling and Analysis of Timed Systems (FORMATS 2022), Co-Located with CONCUR, FMICS, and QEST as part of CONFEST 2022.}
}

@article{tran2021fac,
  title={Verification of Piecewise Deep Neural Networks: A Star Set Approach with Zonotope Pre-Filter},
  author={H. D. Tran and N. Pal and D. Lopez and P. Musau and X. Yang and L. Nguyen, W. Xiang and S. Bak and and T. T. Johnson},
  journal={Formal aspects of computing},
  year={2021},
  publisher={Springer}
}

@inproceedings{durand2022reciph,
  title={ReCIPH: Relational Coefficients for Input Partitioning Heuristic},
  author={Durand, Serge and Lemesle, Augustin and Chihani, Zakaria and Urban, Caterina and Terrier, Fran{\c{c}}ois},
  booktitle={1st Workshop on Formal Verification of Machine Learning (WFVML 2022)},
  year={2022}
}

@article{pyrat2024,
    title={{Neural Network Verification with PyRAT}},
	author={Lemesle, Augustin and Lehmann, Julien and Le Gall Tristan},
	journal={arXiv preprint arXiv:2410.23903},
	year={2024}
}

@inproceedings{Li2025destabilizing,
  author    = {Li, Linhan and Nguyen, ThanhVu},
  title     = {Destabilizing Neurons to Generate Challenging Neural Network Verification Benchmarks},
  booktitle = {Proceedings of the 40th IEEE/ACM International Conference on Automated Software Engineering (ASE)},
  year      = {2025},
  series    = {ASE '25},
  note      = {To appear}
}

@inproceedings{CasadioKDKKAR22,
  author       = {Marco Casadio and
                  Ekaterina Komendantskaya and
                  Matthew L. Daggitt and
                  Wen Kokke and
                  Guy Katz and
                  Guy Amir and
                  Idan Refaeli},
  editor       = {Sharon Shoham and
                  Yakir Vizel},
  title        = {Neural Network Robustness as a Verification Property: {A} Principled
                  Case Study},
  booktitle    = {Computer Aided Verification - 34th International Conference, {CAV}
                  2022, Haifa, Israel, August 7-10, 2022, Proceedings, Part {I}},
  series       = {Lecture Notes in Computer Science},
  volume       = {13371},
  pages        = {219--231},
  publisher    = {Springer},
  year         = {2022},
  url          = {https://doi.org/10.1007/978-3-031-13185-1\_11},
  doi          = {10.1007/978-3-031-13185-1\_11},
  bibsource    = {dblp computer science bibliography, https://dblp.org}
}

@article{bickmore2018patient,
  title={Patient and consumer safety risks when using conversational assistants for medical information: an observational study of Siri, Alexa, and Google Assistant},
  author={Bickmore, Timothy W and Trinh, Ha and Olafsson, Stefan and O'Leary, Teresa K and Asadi, Reza and Rickles, Nathaniel M and Cruz, Ricardo},
  journal={Journal of medical Internet research},
  volume={20},
  number={9},
  pages={e11510},
  year={2018},
  publisher={JMIR Publications Inc., Toronto, Canada}
}

@misc{EUlaw,
      title = {EU Artificial Intelligence Act: The European Approach to AI},
      author = {Mauritz Kop},
      year = {2021},
      url = {https://futurium.ec.europa.eu/sites/default/files/2021-10/Kop\_EU Artificial Intelligence Act - The European Approach to AI\_21092021\_0.pdf}
}

@inproceedings{jia2019certified,
  title={Certified Robustness to Adversarial Word Substitutions},
  author={Jia, Robin and Raghunathan, Aditi and G{\"o}ksel, Kerem and Liang, Percy},
  booktitle={Proceedings of the 2019 Conference on Empirical Methods in Natural Language Processing and the 9th International Joint Conference on Natural Language Processing (EMNLP-IJCNLP)},
  pages={4129--4142},
  year={2019}
}

@inproceedings{huang2019achieving,
  title={Achieving Verified Robustness to Symbol Substitutions via Interval Bound Propagation},
  author={Huang, Po-Sen and Stanforth, Robert and Welbl, Johannes and Dyer, Chris and Yogatama, Dani and Gowal, Sven and Dvijotham, Krishnamurthy and Kohli, Pushmeet},
  booktitle={Proceedings of the 2019 Conference on Empirical Methods in Natural Language Processing and the 9th International Joint Conference on Natural Language Processing (EMNLP-IJCNLP)},
  pages={4083--4093},
  year={2019}
}

@article{welbl2020towards,
  title={Towards verified robustness under text deletion interventions},
  author={Welbl, Johannes and Huang, Po-Sen and Stanforth, Robert and Gowal, Sven and Dvijotham, Krishnamurthy Dj and Szummer, Martin and Kohli, Pushmeet},
  year={2020}
}

@inproceedings{zhang2021certified,
  title={Certified Robustness to Programmable Transformations in LSTMs},
  author={Zhang, Yuhao and Albarghouthi, Aws and D’Antoni, Loris},
  booktitle={Proceedings of the 2021 Conference on Empirical Methods in Natural Language Processing},
  pages={1068--1083},
  year={2021}
}

@inproceedings{wang2023robustness,
  title={Robustness-Aware Word Embedding Improves Certified Robustness to Adversarial Word Substitutions},
  author={Wang, Yibin and Yang, Yichen and He, Di and He, Kun},
  booktitle={Findings of the Association for Computational Linguistics: ACL 2023},
  pages={673--687},
  year={2023}
}

@inproceedings{ko2019popqorn,
  title={POPQORN: Quantifying robustness of recurrent neural networks},
  author={Ko, Ching-Yun and Lyu, Zhaoyang and Weng, Lily and Daniel, Luca and Wong, Ngai and Lin, Dahua},
  booktitle={International Conference on Machine Learning},
  pages={3468--3477},
  year={2019},
  organization={PMLR}
}

@article{du2021cert,
  title={Cert-RNN: Towards Certifying the Robustness of Recurrent Neural Networks.},
  author={Du, Tianyu and Ji, Shouling and Shen, Lujia and Zhang, Yao and Li, Jinfeng and Shi, Jie and Fang, Chengfang and Yin, Jianwei and Beyah, Raheem and Wang, Ting},
  journal={CCS},
  volume={21},
  number={2021},
  pages={15--19},
  year={2021}
}

@misc{shi2020robustness,
      title={Robustness Verification for Transformers}, 
      author={Zhouxing Shi and Huan Zhang and Kai-Wei Chang and Minlie Huang and Cho-Jui Hsieh},
      year={2020},
      eprint={2002.06622},
      archivePrefix={arXiv},
      primaryClass={cs.LG}
}

@inproceedings{bonaert2021fast,
  title={Fast and precise certification of transformers},
  author={Bonaert, Gregory and Dimitrov, Dimitar I and Baader, Maximilian and Vechev, Martin},
  booktitle={Proceedings of the 42nd ACM SIGPLAN International Conference on Programming Language Design and Implementation},
  pages={466--481},
  year={2021}
}

@article{casadio2024nlp,
  title={NLP Verification: Towards a General Methodology for Certifying Robustness},
  author={Casadio, Marco and Dinkar, Tanvi and Komendantskaya, Ekaterina and Arnaboldi, Luca and Isac, Omri and Daggitt, Matthew L and Katz, Guy and Rieser, Verena and Lemon, Oliver},
  journal={arXiv preprint arXiv:2403.10144},
  year={2024}
}

@inproceedings{casadio2023antonio,
  title={ANTONIO: Towards a Systematic Method for Generating NLP Benchmarks for Verification},
  author={Casadio, Marco and Arnaboldi, Luca and Daggitt, Matthew L and Isac, Omri and Dinkar, Tanvi and Kienitz, Daniel and Rieser, Verena and Komendantskaya, Ekaterina},
  booktitle={Proceedings of the 6th Workshop on Formal},
  volume={16},
  pages={59--70},
  year={2023}
}

@inproceedings{gros2021ruarobot,
  title={The RUA-Robot Dataset: Helping Avoid Chatbot Deception by Detecting User Questions About Human or Non-Human Identity},
  author={Gros, David and Li, Yu and Yu, Zhou},
  booktitle={Proceedings of the 59th Annual Meeting of the Association for Computational Linguistics and the 11th International Joint Conference on Natural Language Processing (Volume 1: Long Papers)},
  pages={6999--7013},
  year={2021}
}

@inproceedings{abercrombie2022risk,
  title={Risk-graded Safety for Handling Medical Queries in Conversational AI},
  author={Abercrombie, Gavin and Rieser, Verena},
  booktitle={Proceedings of the 2nd Conference of the Asia-Pacific Chapter of the Association for Computational Linguistics and the 12th International Joint Conference on Natural Language Processing},
  pages={234--243},
  year={2022}
}

@inproceedings{reimers-gurevych-2019-sentence,
    title = "Sentence-{BERT}: Sentence Embeddings using {S}iamese {BERT}-Networks",
    author = "Reimers, Nils  and
      Gurevych, Iryna",
    booktitle = "Proceedings of the 2019 Conference on Empirical Methods in Natural Language Processing and the 9th International Joint Conference on Natural Language Processing (EMNLP-IJCNLP)",
    month = nov,
    year = "2019",
    address = "Hong Kong, China",
    publisher = "Association for Computational Linguistics",
    url = "https://aclanthology.org/D19-1410",
    doi = "10.18653/v1/D19-1410",
    pages = "3982--3992",
}

@inproceedings{moradi2021evaluating,
  title={Evaluating the Robustness of Neural Language Models to Input Perturbations},
  author={Moradi, Milad and Samwald, Matthias},
  booktitle={Proceedings of the 2021 Conference on Empirical Methods in Natural Language Processing},
  pages={1558--1570},
  year={2021}
}

@misc{vicuna2023,
    title = {Vicuna: An Open-Source Chatbot Impressing GPT-4 with 90\%* ChatGPT Quality},
    url = {https://lmsys.org/blog/2023-03-30-vicuna/},
    author = {Chiang, Wei-Lin and Li, Zhuohan and Lin, Zi and Sheng, Ying and Wu, Zhanghao and Zhang, Hao and Zheng, Lianmin and Zhuang, Siyuan and Zhuang, Yonghao and Gonzalez, Joseph E. and Stoica, Ion and Xing, Eric P.},
    month = {March},
    year = {2023}
}

@inproceedings{madry2018towards,
      author = {Madry, Aleksander and Makelov, Aleksandar and Schmidt, Ludwig and Tsipras, Dimitris and Vladu, Adrian},
      booktitle = {International Conference on Learning Representations},
      title = {Towards Deep Learning Models Resistant to Adversarial Attacks},
      year = {2018}
}

@misc{GTSRB,
  title = {{German Traffic Sign Recognition Benchmark}},
  howpublished = {\url{https://www.kaggle.com/datasets/meowmeowmeowmeowmeow/gtsrb-german-traffic-sign?datasetId=82373&language=Python}},
  note = {Accessed: March 25th, 2023}
}

@misc{ChineseTrafficSignDatabase,
  title = {{Chinese Traffic Sign Database}},
  howpublished = {\url{https://www.kaggle.com/datasets/dmitryyemelyanov/chinese-traffic-signs}},
  note = {Accessed: March 25th, 2023}
}

@misc{BelgianTrafficSignDatabase,
  title = {{Belgian Traffic Sign Database}},
  howpublished = {\url{https://www.kaggle.com/datasets/shazaelmorsh/trafficsigns}},
  note = {Accessed: March 25th, 2023}
}

@article{hubara2016binarized,
  title={{Binarized Neural Networks}},
  author={Hubara, Itay and Courbariaux, Matthieu and Soudry, Daniel and El-Yaniv, Ran and Bengio, Yoshua},
  journal="Advances in Neural Information Processing Systems",
  volume={29},
  year={2016}
}

@article{szegedy2013intriguing,
  title={{Intriguing Properties of Neural Networks}},
  author={Szegedy, Christian and Zaremba, Wojciech and Sutskever, Ilya and Bruna, Joan and Erhan, Dumitru and Goodfellow, Ian and Fergus, Rob},
  journal={arXiv preprint arXiv:1312.6199},
  year={2013},
  doi = {10.48550/arXiv.1312.6199}
}

@article{zhang2020lightweight,
  title={{Lightweight Deep Network for Traffic Sign Classification}},
  author={Zhang, Jianming and Wang, Wei and Lu, Chaoquan and Wang, Jin and Sangaiah, Arun Kumar},
  journal={Annals of Telecommunications},
  volume={75},
  pages={369--379},
  year={2020},
  publisher={Springer},
  doi = {10.1007/s12243-019-00731-9}
}

@misc{brix2023years,
      title={First Three Years of the International Verification of Neural Networks Competition (VNN-COMP)}, 
      author={Christopher Brix and Mark Niklas Müller and Stanley Bak and Taylor T. Johnson and Changliu Liu},
      year={2023},
      eprint={2301.05815},
      archivePrefix={arXiv},
      primaryClass={cs.LG}
}

@article{postovan2023architecturing,
  title={Architecturing Binarized Neural Networks for Traffic Sign Recognition},
  author={Postovan, Andreea and Era{\c{s}}cu, M{\u{a}}d{\u{a}}lina},
  journal={arXiv preprint arXiv:2303.15005},
  year={2023},
  doi = {10.48550/arXiv.2303.15005}
}

@inproceedings{dosovitskiy2020image,
  title={An Image is Worth 16x16 Words: Transformers for Image Recognition at Scale},
  author={Dosovitskiy, Alexey and Beyer, Lucas and Kolesnikov, Alexander and Weissenborn, Dirk and Zhai, Xiaohua and Unterthiner, Thomas and Dehghani, Mostafa and Minderer, Matthias and Heigold, Georg and Gelly, Sylvain and others},
  booktitle={International Conference on Learning Representations},
  year={2020}
}

@inproceedings{shi2019robustness,
  title={Robustness Verification for Transformers},
  author={Shi, Zhouxing and Zhang, Huan and Chang, Kai-Wei and Huang, Minlie and Hsieh, Cho-Jui},
  booktitle={International Conference on Learning Representations},
  year={2019}
}

@article{shi2021fast,
  title={Fast certified robust training with short warmup},
  author={Shi, Zhouxing and Wang, Yihan and Zhang, Huan and Yi, Jinfeng and Hsieh, Cho-Jui},
  journal={Advances in Neural Information Processing Systems},
  volume={34},
  pages={18335--18349},
  year={2021}
}

@article{gowal2018effectiveness,
  title={On the effectiveness of interval bound propagation for training verifiably robust models},
  author={Gowal, Sven and Dvijotham, Krishnamurthy and Stanforth, Robert and Bunel, Rudy and Qin, Chongli and Uesato, Jonathan and Arandjelovic, Relja and Mann, Timothy and Kohli, Pushmeet},
  journal={arXiv preprint arXiv:1810.12715},
  year={2018}
}

@article{vaswani2017attention,
  title={Attention is all you need},
  author={Vaswani, Ashish and Shazeer, Noam and Parmar, Niki and Uszkoreit, Jakob and Jones, Llion and Gomez, Aidan N and Kaiser, {\L}ukasz and Polosukhin, Illia},
  journal={Advances in neural information processing systems},
  volume={30},
  year={2017}
}


\appendix

\clearpage
\section{Benchmark Results}
In this section, we provide more fine-grained results.

\subsection{Zero Tolerance}
\subsubsection{Regular Track}
\label{sec:benchmark_results_regular}


\begin{table}[h]
\begin{center}
\caption{Benchmark \texttt{2025-acasxu-2023}} \label{tab:cat_2025_acasxu_2023}
{\setlength{\tabcolsep}{2pt}
\begin{tabular}[h]{@{}llllllrrr@{}}
\toprule
\textbf{\# ~} & \textbf{Tool} & \textbf{Verified} & \textbf{Falsified} & \textbf{Fastest} & \textbf{Penalty} & \textbf{Points} & \textbf{Score} & \textbf{Solved}\\
\midrule
1 & nnenum & 139 & 47 & 0 & 0 & 1860 & 100.0 & 100.0\% \\
2 & NeuralSAT & 139 & 47 & 0 & 0 & 1860 & 100.0 & 100.0\% \\
3 & $\alpha$-$\beta$-CROWN & 139 & 47 & 0 & 0 & 1860 & 100.0 & 100.0\% \\
4 & PyRAT & 139 & 46 & 0 & 0 & 1850 & 99.5 & 99.5\% \\
5 & CORA & 137 & 46 & 0 & 0 & 1830 & 98.4 & 98.4\% \\
6 & SobolBox & 118 & 43 & 0 & 1 & 1460 & 78.5 & 86.6\% \\
7 & NNV & 71 & 39 & 0 & 0 & 1100 & 59.1 & 59.1\% \\
\bottomrule
\end{tabular}
}
\end{center}
\end{table}

\begin{figure}[h]
\centerline{\includegraphics[width=\textwidth]{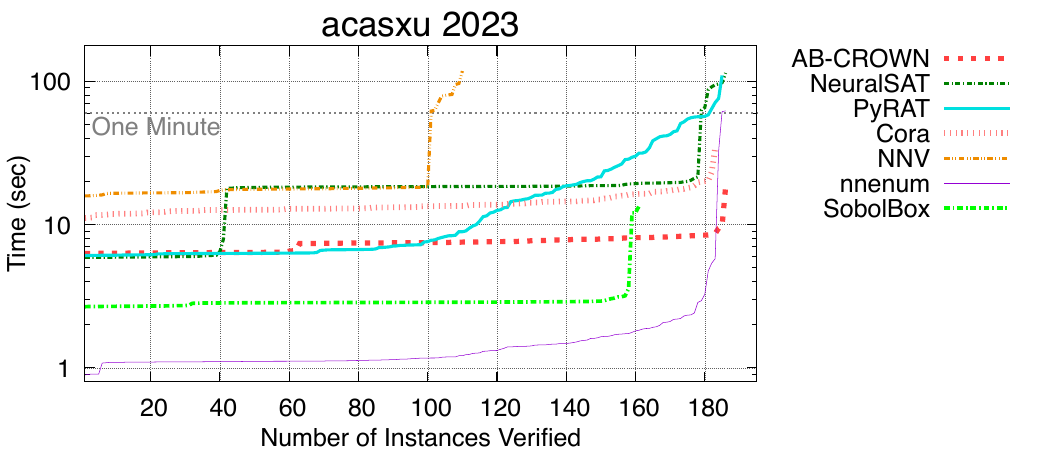}}
\caption{Cactus Plot for acasxu 2023.}
\label{fig:quantPic_acasxu}
\end{figure}

\clearpage

\begin{table}[h]
\begin{center}
\caption{Benchmark \texttt{2025-cersyve}} \label{tab:cat_2025_cersyve}
{\setlength{\tabcolsep}{2pt}
\begin{tabular}[h]{@{}llllllrrr@{}}
\toprule
\textbf{\# ~} & \textbf{Tool} & \textbf{Verified} & \textbf{Falsified} & \textbf{Fastest} & \textbf{Penalty} & \textbf{Points} & \textbf{Score} & \textbf{Solved}\\
\midrule
1 & $\alpha$-$\beta$-CROWN & 6 & 6 & 0 & 0 & 120 & 100.0 & 100.0\% \\
2 & NeuralSAT & 4 & 6 & 0 & 0 & 100 & 83.3 & 83.3\% \\
3 & PyRAT & 2 & 6 & 0 & 0 & 80 & 66.7 & 66.7\% \\
4 & SobolBox & 0 & 6 & 0 & 0 & 60 & 50.0 & 50.0\% \\
5 & NNV & 0 & 3 & 0 & 0 & 30 & 25.0 & 25.0\% \\
6 & CORA & 6 & 5 & 0 & 1 & -40 & 0 & 91.7\% \\
\bottomrule
\end{tabular}
}
\end{center}
\end{table}

\begin{figure}[h]
\centerline{\includegraphics[width=\textwidth]{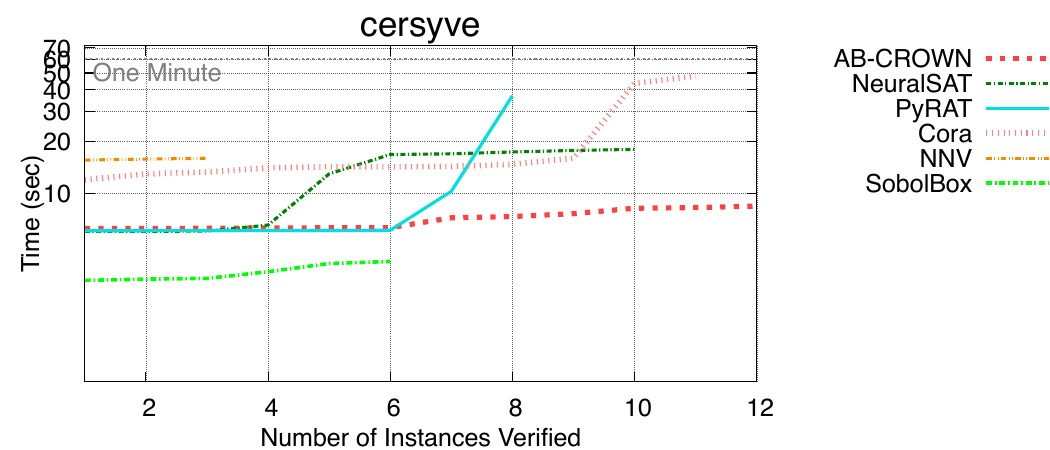}}
\caption{Cactus Plot for cersyve.}
\label{fig:quantPic_cersyve}
\end{figure}

\clearpage

\begin{table}[h]
\begin{center}
\caption{Benchmark \texttt{2025-cgan-2023}} \label{tab:cat_2025_cgan_2023}
{\setlength{\tabcolsep}{2pt}
\begin{tabular}[h]{@{}llllllrrr@{}}
\toprule
\textbf{\# ~} & \textbf{Tool} & \textbf{Verified} & \textbf{Falsified} & \textbf{Fastest} & \textbf{Penalty} & \textbf{Points} & \textbf{Score} & \textbf{Solved}\\
\midrule
1 & PyRAT & 9 & 12 & 0 & 0 & 210 & 100.0 & 100.0\% \\
2 & NeuralSAT & 9 & 12 & 0 & 0 & 210 & 100.0 & 100.0\% \\
3 & $\alpha$-$\beta$-CROWN & 9 & 12 & 0 & 0 & 210 & 100.0 & 100.0\% \\
4 & SobolBox & 9 & 10 & 0 & 0 & 190 & 90.5 & 90.5\% \\
5 & nnenum & 7 & 10 & 0 & 0 & 170 & 81.0 & 81.0\% \\
6 & NNV & 5 & 11 & 0 & 0 & 160 & 76.2 & 76.2\% \\
\bottomrule
\end{tabular}
}
\end{center}
\end{table}

\begin{figure}[h]
\centerline{\includegraphics[width=\textwidth]{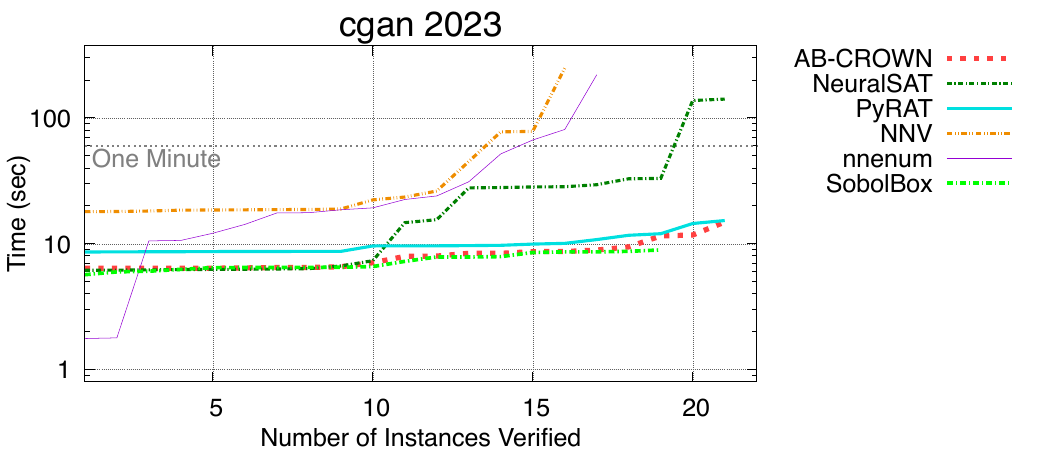}}
\caption{Cactus Plot for cgan 2023.}
\label{fig:quantPic_cgan}
\end{figure}

\clearpage

\begin{table}[h]
\begin{center}
\caption{Benchmark \texttt{2025-cifar100-2024}} \label{tab:cat_2025_cifar100_2024}
{\setlength{\tabcolsep}{2pt}
\begin{tabular}[h]{@{}llllllrrr@{}}
\toprule
\textbf{\# ~} & \textbf{Tool} & \textbf{Verified} & \textbf{Falsified} & \textbf{Fastest} & \textbf{Penalty} & \textbf{Points} & \textbf{Score} & \textbf{Solved}\\
\midrule
1 & NNV & 190 & 0 & 0 & 0 & 1900 & 100.0 & 95.0\% \\
2 & $\alpha$-$\beta$-CROWN & 100 & 29 & 0 & 0 & 1290 & 67.9 & 64.5\% \\
3 & PyRAT & 61 & 25 & 0 & 0 & 860 & 45.3 & 43.0\% \\
4 & NeuralSAT & 87 & 25 & 0 & 4 & 520 & 27.4 & 56.0\% \\
5 & CORA & 0 & 10 & 0 & 0 & 100 & 5.3 & 5.0\% \\
\bottomrule
\end{tabular}
}
\end{center}
\end{table}

\begin{figure}[h]
\centerline{\includegraphics[width=\textwidth]{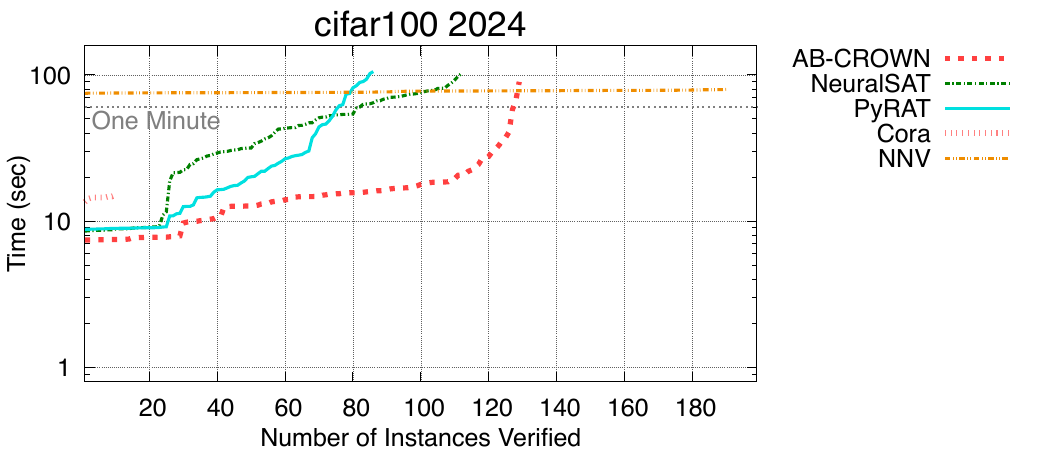}}
\caption{Cactus Plot for cifar100 2024.}
\label{fig:quantPic_cifar100}
\end{figure}

\clearpage

\begin{table}[h]
\begin{center}
\caption{Benchmark \texttt{2025-collins-rul-cnn-2022}} \label{tab:cat_2025_collins_rul_cnn_2022}
{\setlength{\tabcolsep}{2pt}
\begin{tabular}[h]{@{}llllllrrr@{}}
\toprule
\textbf{\# ~} & \textbf{Tool} & \textbf{Verified} & \textbf{Falsified} & \textbf{Fastest} & \textbf{Penalty} & \textbf{Points} & \textbf{Score} & \textbf{Solved}\\
\midrule
1 & nnenum & 39 & 23 & 0 & 0 & 620 & 100.0 & 100.0\% \\
2 & PyRAT & 39 & 23 & 0 & 0 & 620 & 100.0 & 100.0\% \\
3 & NeuralSAT & 39 & 23 & 0 & 0 & 620 & 100.0 & 100.0\% \\
4 & NNV & 39 & 23 & 0 & 0 & 620 & 100.0 & 100.0\% \\
5 & CORA & 39 & 23 & 0 & 0 & 620 & 100.0 & 100.0\% \\
6 & $\alpha$-$\beta$-CROWN & 39 & 23 & 0 & 0 & 620 & 100.0 & 100.0\% \\
7 & SobolBox & 19 & 15 & 0 & 0 & 340 & 54.8 & 54.8\% \\
\bottomrule
\end{tabular}
}
\end{center}
\end{table}

\begin{figure}[h]
\centerline{\includegraphics[width=\textwidth]{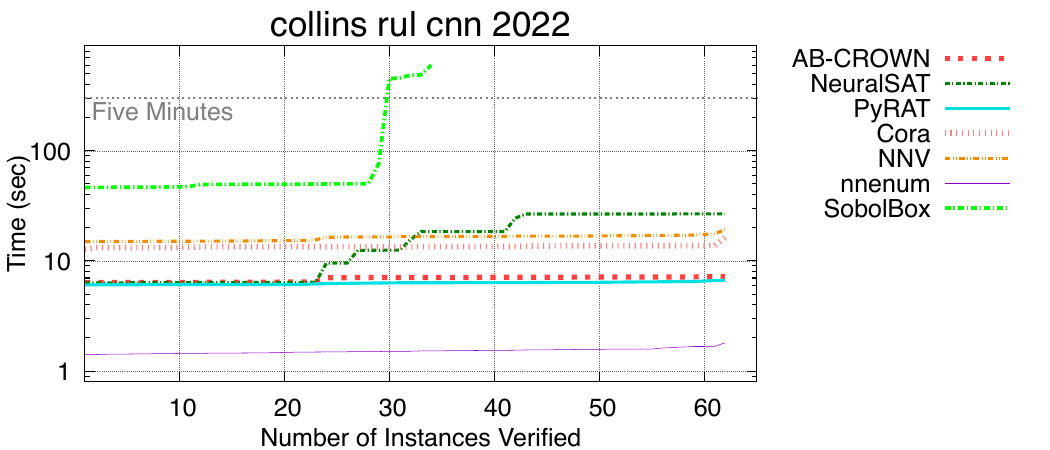}}
\caption{Cactus Plot for collins rul cnn 2022.}
\label{fig:quantPic_rul}
\end{figure}

\clearpage

\begin{table}[h]
\begin{center}
\caption{Benchmark \texttt{2025-cora-2024}} \label{tab:cat_2025_cora_2024}
{\setlength{\tabcolsep}{2pt}
\begin{tabular}[h]{@{}llllllrrr@{}}
\toprule
\textbf{\# ~} & \textbf{Tool} & \textbf{Verified} & \textbf{Falsified} & \textbf{Fastest} & \textbf{Penalty} & \textbf{Points} & \textbf{Score} & \textbf{Solved}\\
\midrule
1 & $\alpha$-$\beta$-CROWN & 22 & 131 & 0 & 0 & 1530 & 100.0 & 85.0\% \\
2 & NeuralSAT & 21 & 131 & 0 & 0 & 1520 & 99.3 & 84.4\% \\
3 & PyRAT & 20 & 128 & 0 & 0 & 1480 & 96.7 & 82.2\% \\
4 & CORA & 19 & 124 & 0 & 0 & 1430 & 93.5 & 79.4\% \\
5 & NNV & 19 & 57 & 0 & 0 & 760 & 49.7 & 42.2\% \\
6 & nnenum & 19 & 4 & 0 & 0 & 230 & 15.0 & 12.8\% \\
\bottomrule
\end{tabular}
}
\end{center}
\end{table}

\begin{figure}[h]
\centerline{\includegraphics[width=\textwidth]{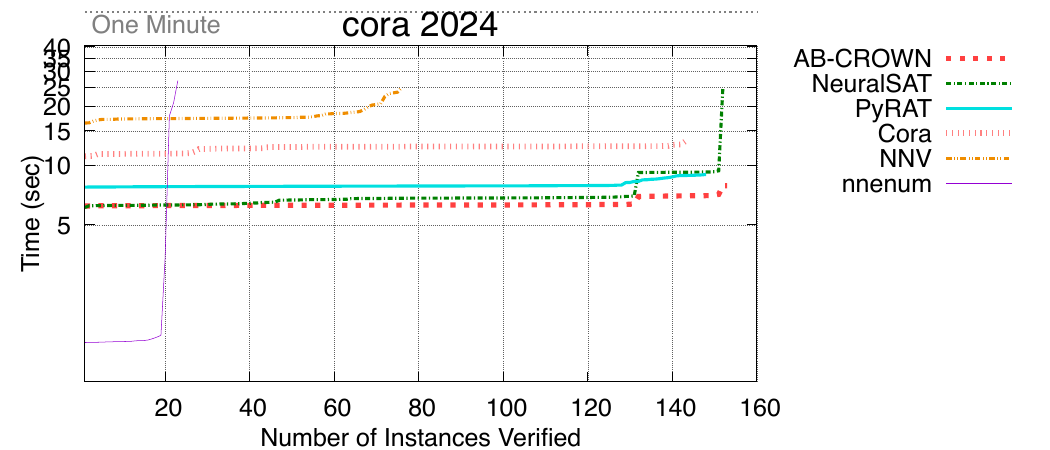}}
\caption{Cactus Plot for cora 2024.}
\label{fig:quantPic_cora}
\end{figure}

\clearpage

\begin{table}[h]
\begin{center}
\caption{Benchmark \texttt{2025-dist-shift-2023}} \label{tab:cat_2025_dist_shift_2023}
{\setlength{\tabcolsep}{2pt}
\begin{tabular}[h]{@{}llllllrrr@{}}
\toprule
\textbf{\# ~} & \textbf{Tool} & \textbf{Verified} & \textbf{Falsified} & \textbf{Fastest} & \textbf{Penalty} & \textbf{Points} & \textbf{Score} & \textbf{Solved}\\
\midrule
1 & NeuralSAT & 65 & 7 & 0 & 0 & 720 & 100.0 & 100.0\% \\
2 & CORA & 65 & 7 & 0 & 0 & 720 & 100.0 & 100.0\% \\
3 & $\alpha$-$\beta$-CROWN & 65 & 7 & 0 & 0 & 720 & 100.0 & 100.0\% \\
4 & PyRAT & 64 & 7 & 0 & 0 & 710 & 98.6 & 98.6\% \\
5 & NNV & 50 & 4 & 0 & 0 & 540 & 75.0 & 75.0\% \\
\bottomrule
\end{tabular}
}
\end{center}
\end{table}

\begin{figure}[h]
\centerline{\includegraphics[width=\textwidth]{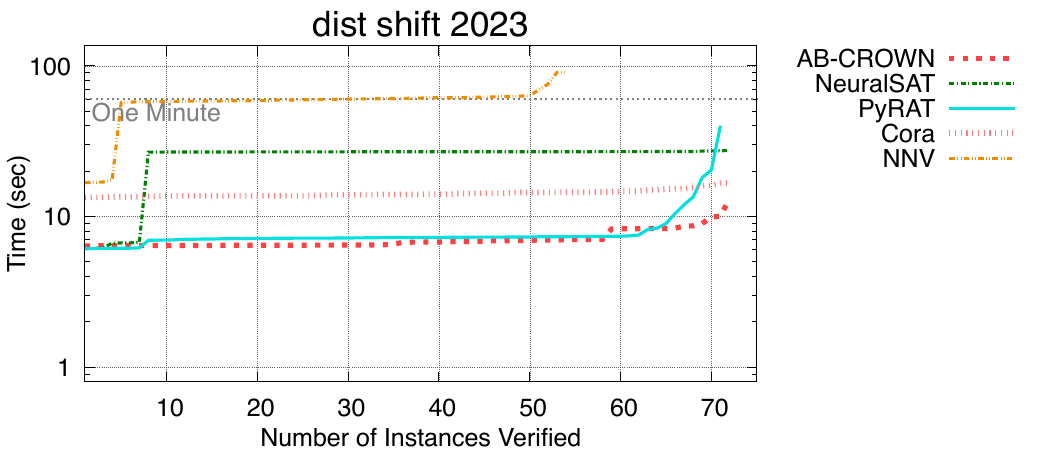}}
\caption{Cactus Plot for dist shift 2023.}
\label{fig:quantPic_dist_shift}
\end{figure}

\clearpage

\begin{table}[h]
\begin{center}
\caption{Benchmark \texttt{2025-linearizenn-2024}} \label{tab:cat_2025_linearizenn_2024}
{\setlength{\tabcolsep}{2pt}
\begin{tabular}[h]{@{}llllllrrr@{}}
\toprule
\textbf{\# ~} & \textbf{Tool} & \textbf{Verified} & \textbf{Falsified} & \textbf{Fastest} & \textbf{Penalty} & \textbf{Points} & \textbf{Score} & \textbf{Solved}\\
\midrule
1 & nnenum & 59 & 1 & 0 & 0 & 600 & 100.0 & 100.0\% \\
2 & SobolBox & 59 & 1 & 0 & 0 & 600 & 100.0 & 100.0\% \\
3 & PyRAT & 59 & 1 & 0 & 0 & 600 & 100.0 & 100.0\% \\
4 & NeuralSAT & 59 & 1 & 0 & 0 & 600 & 100.0 & 100.0\% \\
5 & CORA & 59 & 1 & 0 & 0 & 600 & 100.0 & 100.0\% \\
6 & $\alpha$-$\beta$-CROWN & 59 & 1 & 0 & 0 & 600 & 100.0 & 100.0\% \\
7 & NNV & 40 & 1 & 0 & 0 & 410 & 68.3 & 68.3\% \\
\bottomrule
\end{tabular}
}
\end{center}
\end{table}

\begin{figure}[h]
\centerline{\includegraphics[width=\textwidth]{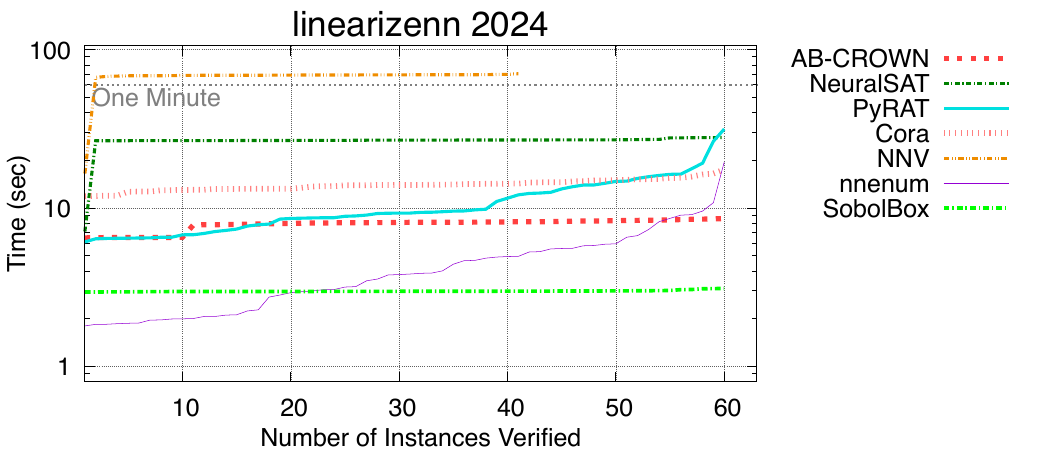}}
\caption{Cactus Plot for linearizenn 2024.}
\label{fig:quantPic_linearizenn}
\end{figure}

\clearpage

\begin{table}[h]
\begin{center}
\caption{Benchmark \texttt{2025-malbeware}} \label{tab:cat_2025_malbeware}
{\setlength{\tabcolsep}{2pt}
\begin{tabular}[h]{@{}llllllrrr@{}}
\toprule
\textbf{\# ~} & \textbf{Tool} & \textbf{Verified} & \textbf{Falsified} & \textbf{Fastest} & \textbf{Penalty} & \textbf{Points} & \textbf{Score} & \textbf{Solved}\\
\midrule
1 & $\alpha$-$\beta$-CROWN & 131 & 19 & 0 & 0 & 1500 & 100.0 & 100.0\% \\
2 & NeuralSAT & 127 & 19 & 0 & 0 & 1460 & 97.3 & 97.3\% \\
3 & PyRAT & 121 & 18 & 0 & 0 & 1390 & 92.7 & 92.7\% \\
4 & nnenum & 125 & 12 & 0 & 0 & 1370 & 91.3 & 91.3\% \\
5 & CORA & 88 & 9 & 0 & 0 & 970 & 64.7 & 64.7\% \\
6 & NNV & 49 & 4 & 0 & 0 & 530 & 35.3 & 35.3\% \\
\bottomrule
\end{tabular}
}
\end{center}
\end{table}

\begin{figure}[h]
\centerline{\includegraphics[width=\textwidth]{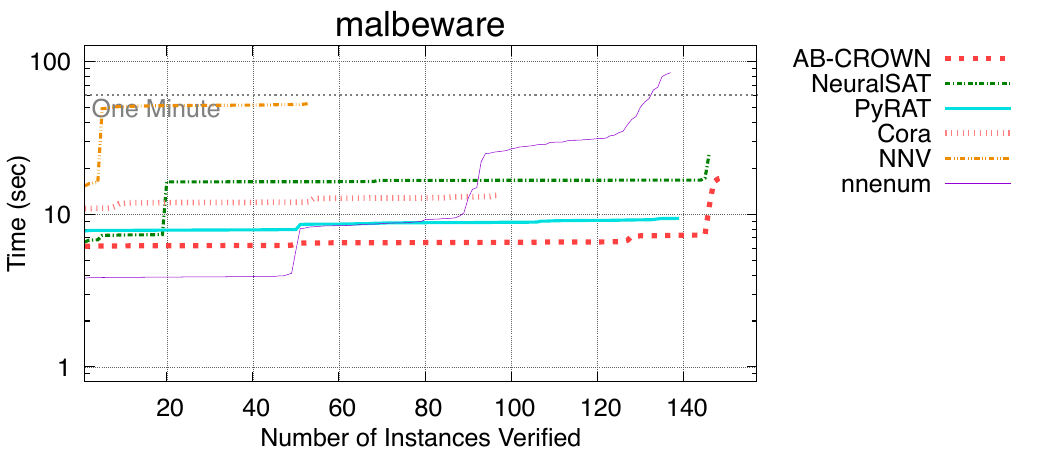}}
\caption{Cactus Plot for malbeware.}
\label{fig:quantPic_malbeware}
\end{figure}

\clearpage

\begin{table}[h]
\begin{center}
\caption{Benchmark \texttt{2025-metaroom-2023}} \label{tab:cat_2025_metaroom_2023}
{\setlength{\tabcolsep}{2pt}
\begin{tabular}[h]{@{}llllllrrr@{}}
\toprule
\textbf{\# ~} & \textbf{Tool} & \textbf{Verified} & \textbf{Falsified} & \textbf{Fastest} & \textbf{Penalty} & \textbf{Points} & \textbf{Score} & \textbf{Solved}\\
\midrule
1 & PyRAT & 97 & 3 & 0 & 0 & 1000 & 100.0 & 100.0\% \\
2 & NeuralSAT & 94 & 5 & 0 & 0 & 990 & 99.0 & 99.0\% \\
3 & $\alpha$-$\beta$-CROWN & 94 & 5 & 0 & 0 & 990 & 99.0 & 99.0\% \\
4 & CORA & 92 & 5 & 0 & 0 & 970 & 97.0 & 97.0\% \\
5 & NNV & 93 & 2 & 0 & 0 & 950 & 95.0 & 95.0\% \\
6 & nnenum & 50 & 1 & 0 & 0 & 510 & 51.0 & 51.0\% \\
\bottomrule
\end{tabular}
}
\end{center}
\end{table}

\begin{figure}[h]
\centerline{\includegraphics[width=\textwidth]{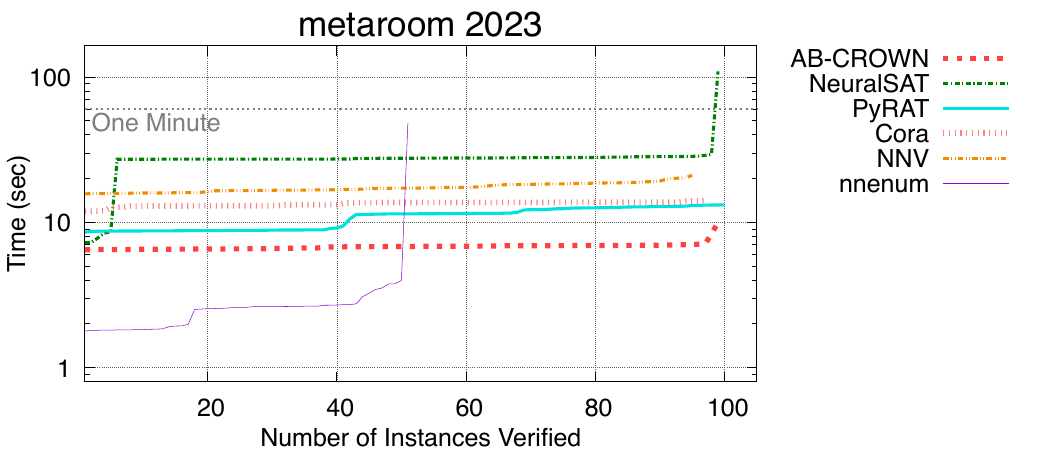}}
\caption{Cactus Plot for metaroom 2023.}
\label{fig:quantPic_metaroom}
\end{figure}

\clearpage

\begin{table}[h]
\begin{center}
\caption{Benchmark \texttt{2025-nn4sys}} \label{tab:cat_2025_nn4sys}
{\setlength{\tabcolsep}{2pt}
\begin{tabular}[h]{@{}llllllrrr@{}}
\toprule
\textbf{\# ~} & \textbf{Tool} & \textbf{Verified} & \textbf{Falsified} & \textbf{Fastest} & \textbf{Penalty} & \textbf{Points} & \textbf{Score} & \textbf{Solved}\\
\midrule
1 & $\alpha$-$\beta$-CROWN & 194 & 0 & 0 & 0 & 1940 & 100.0 & 100.0\% \\
2 & NeuralSAT & 120 & 0 & 0 & 0 & 1200 & 61.9 & 61.9\% \\
3 & SobolBox & 107 & 0 & 0 & 0 & 1070 & 55.2 & 55.2\% \\
4 & PyRAT & 40 & 0 & 0 & 0 & 400 & 20.6 & 20.6\% \\
5 & nnenum & 22 & 0 & 0 & 0 & 220 & 11.3 & 11.3\% \\
6 & NNV & 17 & 0 & 0 & 0 & 170 & 8.8 & 8.8\% \\
\bottomrule
\end{tabular}
}
\end{center}
\end{table}

\begin{figure}[h]
\centerline{\includegraphics[width=\textwidth]{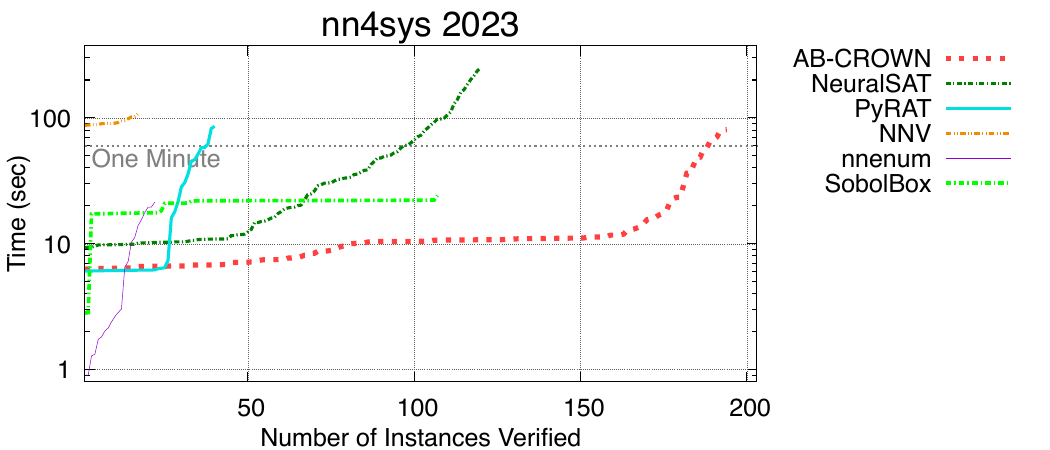}}
\caption{Cactus Plot for nn4sys 2023.}
\label{fig:quantPic_nn4sys}
\end{figure}

\clearpage

\begin{table}[h]
\begin{center}
\caption{Benchmark \texttt{2025-safenlp-2024}} \label{tab:cat_2025_safenlp_2024}
{\setlength{\tabcolsep}{2pt}
\begin{tabular}[h]{@{}llllllrrr@{}}
\toprule
\textbf{\# ~} & \textbf{Tool} & \textbf{Verified} & \textbf{Falsified} & \textbf{Fastest} & \textbf{Penalty} & \textbf{Points} & \textbf{Score} & \textbf{Solved}\\
\midrule
1 & $\alpha$-$\beta$-CROWN & 433 & 647 & 0 & 0 & 10800 & 100.0 & 100.0\% \\
2 & NeuralSAT & 425 & 645 & 0 & 0 & 10700 & 99.1 & 99.1\% \\
3 & PyRAT & 331 & 647 & 0 & 0 & 9780 & 90.6 & 90.6\% \\
4 & nnenum & 340 & 636 & 0 & 0 & 9760 & 90.4 & 90.4\% \\
5 & CORA & 338 & 644 & 0 & 1 & 9670 & 89.5 & 90.9\% \\
6 & SobolBox & 795 & 215 & 0 & 21 & 6950 & 64.4 & 93.5\% \\
7 & NNV & 172 & 176 & 0 & 0 & 3480 & 32.2 & 32.2\% \\
\bottomrule
\end{tabular}
}
\end{center}
\end{table}

\begin{figure}[h]
\centerline{\includegraphics[width=\textwidth]{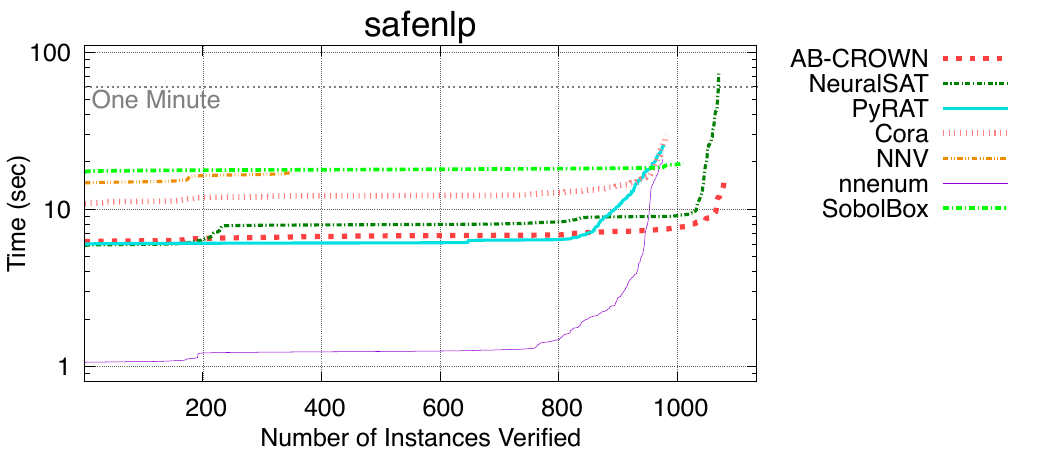}}
\caption{Cactus Plot for safenlp.}
\label{fig:quantPic_safenlp}
\end{figure}

\clearpage

\begin{table}[h]
\begin{center}
\caption{Benchmark \texttt{2025-sat-relu}} \label{tab:cat_2025_sat_relu}
{\setlength{\tabcolsep}{2pt}
\begin{tabular}[h]{@{}llllllrrr@{}}
\toprule
\textbf{\# ~} & \textbf{Tool} & \textbf{Verified} & \textbf{Falsified} & \textbf{Fastest} & \textbf{Penalty} & \textbf{Points} & \textbf{Score} & \textbf{Solved}\\
\midrule
1 & NeuralSAT & 50 & 50 & 0 & 0 & 1000 & 100.0 & 100.0\% \\
2 & CORA & 50 & 50 & 0 & 0 & 1000 & 100.0 & 100.0\% \\
3 & $\alpha$-$\beta$-CROWN & 50 & 50 & 0 & 0 & 1000 & 100.0 & 100.0\% \\
4 & PyRAT & 9 & 50 & 0 & 0 & 590 & 59.0 & 59.0\% \\
5 & nnenum & 9 & 35 & 0 & 0 & 440 & 44.0 & 44.0\% \\
6 & NNV & 2 & 16 & 0 & 0 & 180 & 18.0 & 18.0\% \\
7 & SobolBox & 0 & 10 & 0 & 33 & -4850 & 0 & 10.0\% \\
\bottomrule
\end{tabular}
}
\end{center}
\end{table}

\begin{figure}[h]
\centerline{\includegraphics[width=\textwidth]{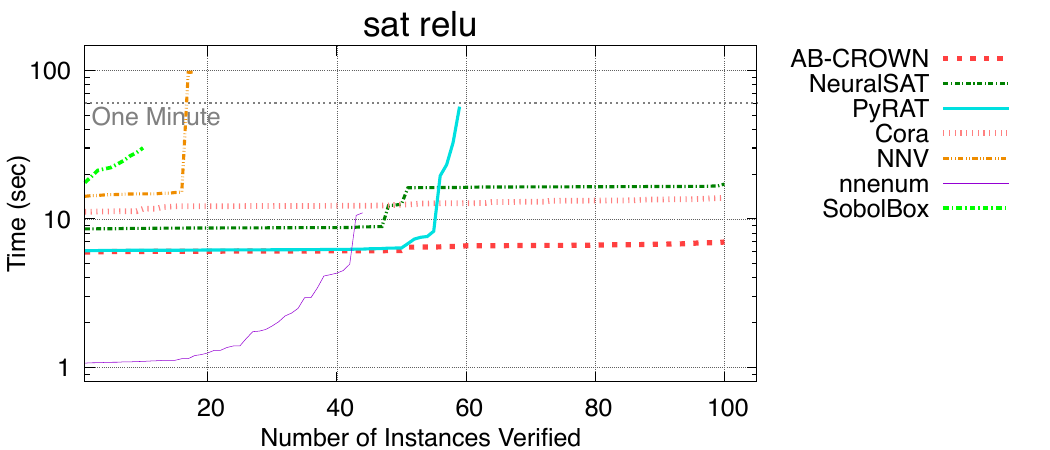}}
\caption{Cactus Plot for sat relu.}
\label{fig:quantPic_sat_relu}
\end{figure}

\clearpage

\begin{table}[h]
\begin{center}
\caption{Benchmark \texttt{2025-soundnessbench}} \label{tab:cat_2025_soundnessbench}
{\setlength{\tabcolsep}{2pt}
\begin{tabular}[h]{@{}llllllrrr@{}}
\toprule
\textbf{\# ~} & \textbf{Tool} & \textbf{Verified} & \textbf{Falsified} & \textbf{Fastest} & \textbf{Penalty} & \textbf{Points} & \textbf{Score} & \textbf{Solved}\\
\midrule
1 & $\alpha$-$\beta$-CROWN & 0 & 50 & 0 & 0 & 500 & 100.0 & 100.0\% \\
2 & NeuralSAT & 12 & 30 & 0 & 0 & 420 & 84.0 & 84.0\% \\
3 & CORA & 18 & 0 & 0 & 0 & 180 & 36.0 & 36.0\% \\
\bottomrule
\end{tabular}
}
\end{center}
\end{table}

\begin{figure}[h]
\centerline{\includegraphics[width=\textwidth]{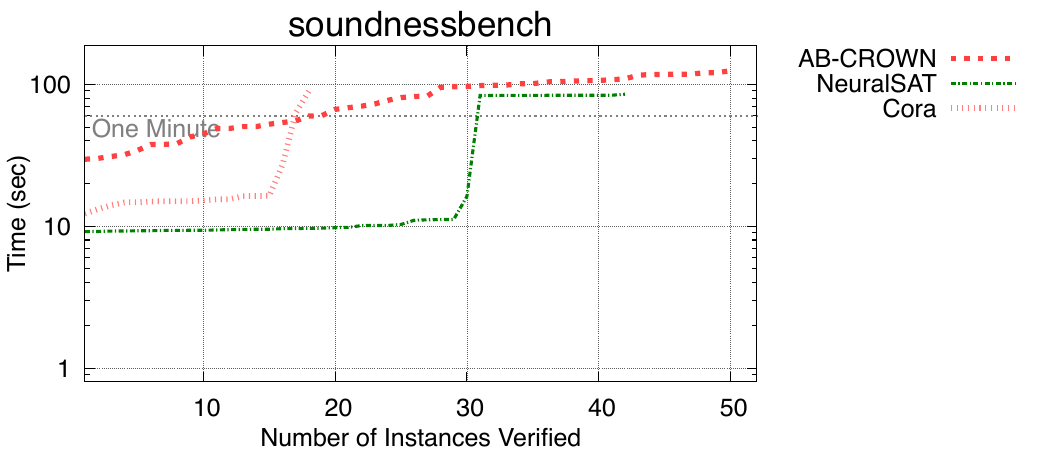}}
\caption{Cactus Plot for soundnessbench.}
\label{fig:quantPic_soundnessbench}
\end{figure}

\clearpage

\begin{table}[h]
\begin{center}
\caption{Benchmark \texttt{2025-tinyimagenet-2024}} \label{tab:cat_2025_tinyimagenet_2024}
{\setlength{\tabcolsep}{2pt}
\begin{tabular}[h]{@{}llllllrrr@{}}
\toprule
\textbf{\# ~} & \textbf{Tool} & \textbf{Verified} & \textbf{Falsified} & \textbf{Fastest} & \textbf{Penalty} & \textbf{Points} & \textbf{Score} & \textbf{Solved}\\
\midrule
1 & $\alpha$-$\beta$-CROWN & 137 & 38 & 0 & 0 & 1750 & 100.0 & 87.5\% \\
2 & NeuralSAT & 116 & 37 & 0 & 1 & 1380 & 78.9 & 76.5\% \\
3 & PyRAT & 68 & 35 & 0 & 0 & 1030 & 58.9 & 51.5\% \\
4 & CORA & 0 & 5 & 0 & 0 & 50 & 2.9 & 2.5\% \\
5 & NNV & 0 & 1 & 0 & 0 & 10 & 0.6 & 0.5\% \\
\bottomrule
\end{tabular}
}
\end{center}
\end{table}

\begin{figure}[h]
\centerline{\includegraphics[width=\textwidth]{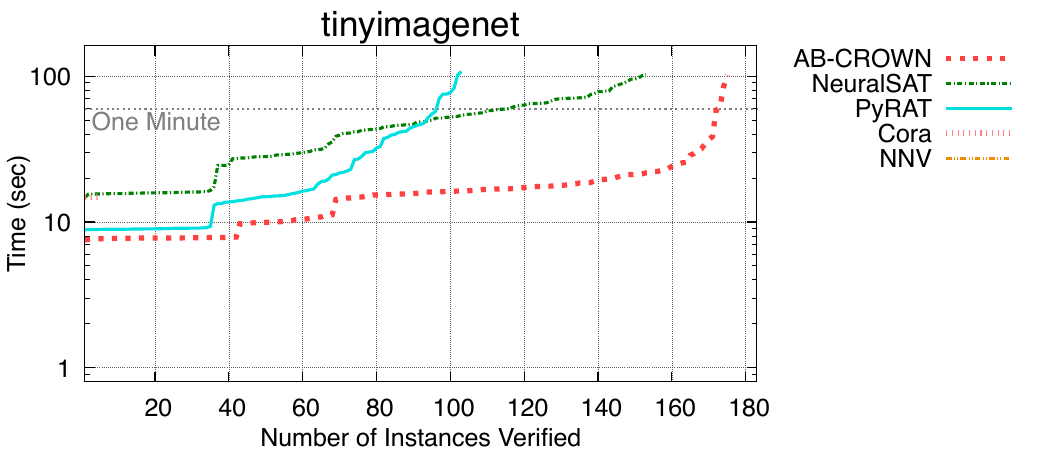}}
\caption{Cactus Plot for tinyimagenet.}
\label{fig:quantPic_tinyimagenet}
\end{figure}

\clearpage

\begin{table}[h]
\begin{center}
\caption{Benchmark \texttt{2025-tllverifybench-2023}} \label{tab:cat_2025_tllverifybench_2023}
{\setlength{\tabcolsep}{2pt}
\begin{tabular}[h]{@{}llllllrrr@{}}
\toprule
\textbf{\# ~} & \textbf{Tool} & \textbf{Verified} & \textbf{Falsified} & \textbf{Fastest} & \textbf{Penalty} & \textbf{Points} & \textbf{Score} & \textbf{Solved}\\
\midrule
1 & SobolBox & 15 & 17 & 0 & 0 & 320 & 100.0 & 100.0\% \\
2 & PyRAT & 15 & 17 & 0 & 0 & 320 & 100.0 & 100.0\% \\
3 & NeuralSAT & 15 & 17 & 0 & 0 & 320 & 100.0 & 100.0\% \\
4 & CORA & 15 & 17 & 0 & 0 & 320 & 100.0 & 100.0\% \\
5 & $\alpha$-$\beta$-CROWN & 15 & 17 & 0 & 0 & 320 & 100.0 & 100.0\% \\
6 & nnenum & 1 & 17 & 0 & 0 & 180 & 56.2 & 56.2\% \\
7 & NNV & 0 & 17 & 0 & 0 & 170 & 53.1 & 53.1\% \\
\bottomrule
\end{tabular}
}
\end{center}
\end{table}

\begin{figure}[h]
\centerline{\includegraphics[width=\textwidth]{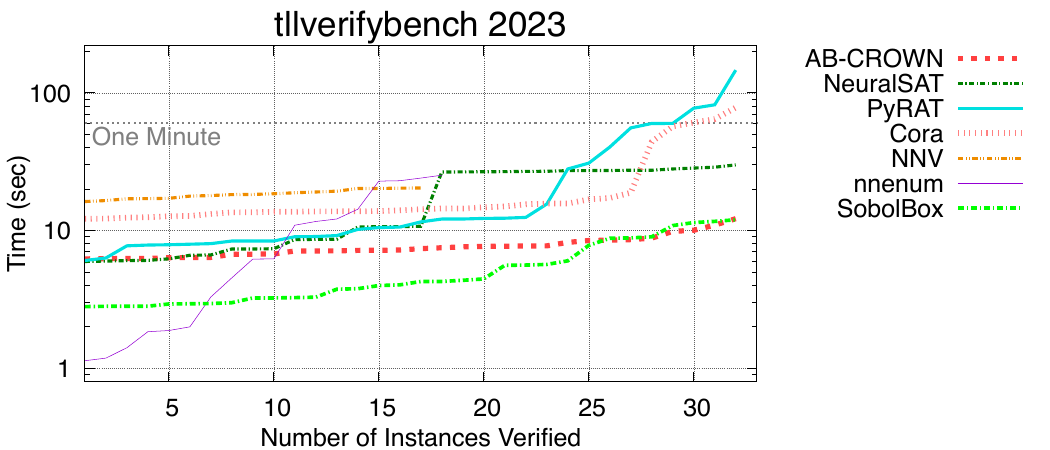}}
\caption{Cactus Plot for tllverifybench 2023.}
\label{fig:quantPic_tll}
\end{figure}

\clearpage
\subsubsection{Extended Track}
\label{sec:benchmark_results_extended}

\clearpage

\begin{table}[h]
\begin{center}
\caption{Benchmark \texttt{2025-cctsdb-yolo-2023}} \label{tab:cat_2025_cctsdb_yolo_2023}
{\setlength{\tabcolsep}{2pt}
\begin{tabular}[h]{@{}llllllrrr@{}}
\toprule
\textbf{\# ~} & \textbf{Tool} & \textbf{Verified} & \textbf{Falsified} & \textbf{Fastest} & \textbf{Penalty} & \textbf{Points} & \textbf{Score} & \textbf{Solved}\\
\midrule
1 & PyRAT & 11 & 28 & 0 & 0 & 390 & 100.0 & 100.0\% \\
2 & $\alpha$-$\beta$-CROWN & 11 & 28 & 0 & 0 & 390 & 100.0 & 100.0\% \\
\bottomrule
\end{tabular}
}
\end{center}
\end{table}

\begin{figure}[h]
\centerline{\includegraphics[width=\textwidth]{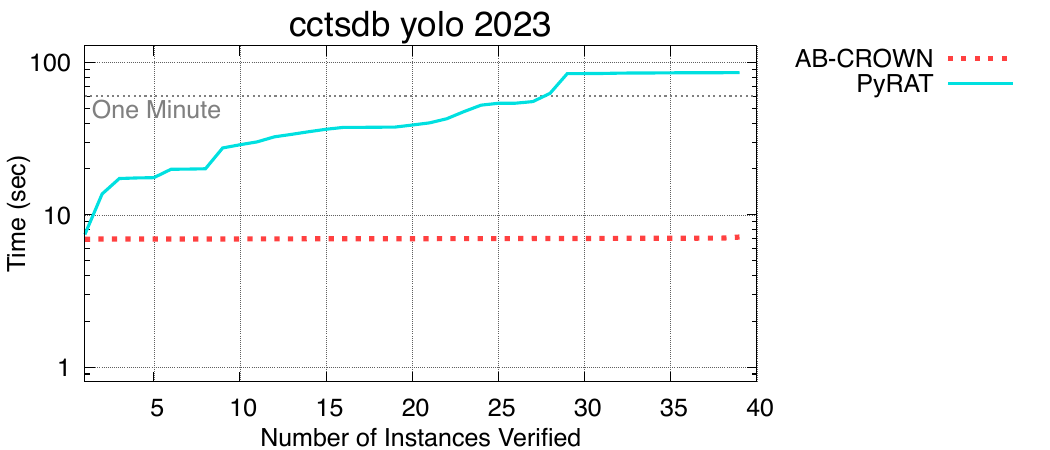}}
\caption{Cactus Plot for cctsdb yolo 2023.}
\label{fig:quantPic_cctsdb}
\end{figure}

\clearpage

\begin{table}[h]
\begin{center}
\caption{Benchmark \texttt{2025-collins-aerospace-benchmark}} \label{tab:cat_2025_collins_aerospace_benchmark}
{\setlength{\tabcolsep}{2pt}
\begin{tabular}[h]{@{}llllllrrr@{}}
\toprule
\textbf{\# ~} & \textbf{Tool} & \textbf{Verified} & \textbf{Falsified} & \textbf{Fastest} & \textbf{Penalty} & \textbf{Points} & \textbf{Score} & \textbf{Solved}\\
\midrule
1 & PyRAT & 0 & 6 & 0 & 0 & 60 & 100.0 & 100.0\% \\
2 & $\alpha$-$\beta$-CROWN & 0 & 6 & 0 & 0 & 60 & 100.0 & 100.0\% \\
\bottomrule
\end{tabular}
}
\end{center}
\end{table}

\begin{figure}[h]
\centerline{\includegraphics[width=\textwidth]{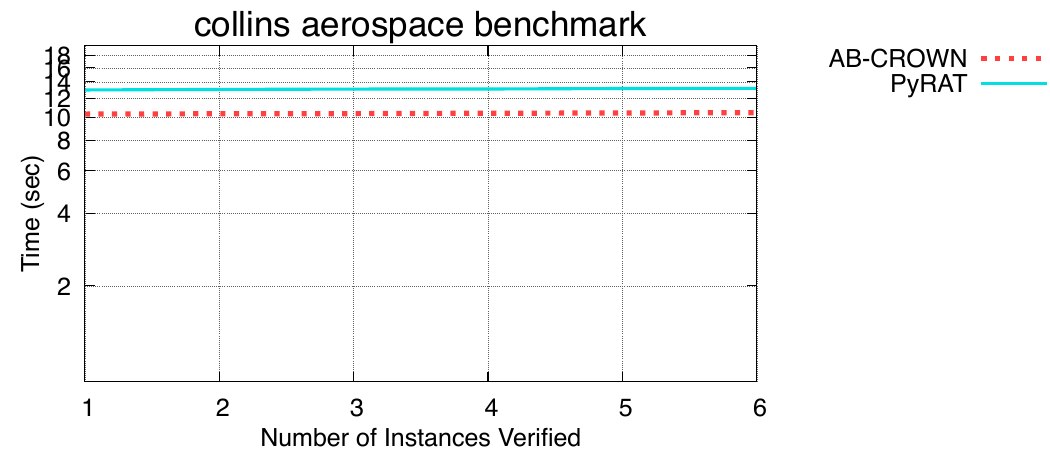}}
\caption{Cactus Plot for collins aerospace benchmark.}
\label{fig:quantPic_collins}
\end{figure}

\clearpage

\begin{table}[h]
\begin{center}
\caption{Benchmark \texttt{2025-lsnc-relu}} \label{tab:cat_2025_lsnc_relu}
{\setlength{\tabcolsep}{2pt}
\begin{tabular}[h]{@{}llllllrrr@{}}
\toprule
\textbf{\# ~} & \textbf{Tool} & \textbf{Verified} & \textbf{Falsified} & \textbf{Fastest} & \textbf{Penalty} & \textbf{Points} & \textbf{Score} & \textbf{Solved}\\
\midrule
1 & $\alpha$-$\beta$-CROWN & 68 & 12 & 0 & 0 & 800 & 100.0 & 100.0\% \\
\bottomrule
\end{tabular}
}
\end{center}
\end{table}

\begin{figure}[h]
\centerline{\includegraphics[width=\textwidth]{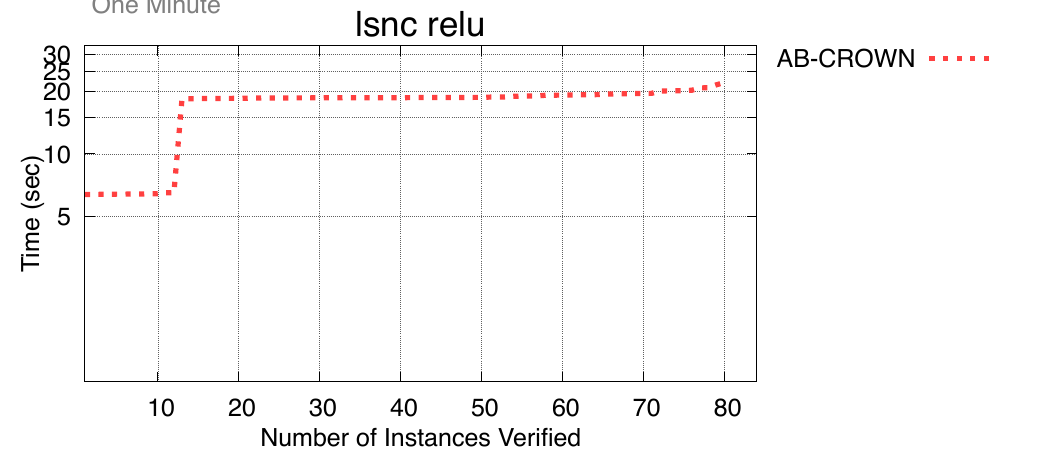}}
\caption{Cactus Plot for lsnc relu.}
\label{fig:quantPic_lsnc}
\end{figure}

\clearpage

\begin{table}[h]
\begin{center}
\caption{Benchmark \texttt{2025-ml4acopf-2024}} \label{tab:cat_2025_ml4acopf_2024}
{\setlength{\tabcolsep}{2pt}
\begin{tabular}[h]{@{}llllllrrr@{}}
\toprule
\textbf{\# ~} & \textbf{Tool} & \textbf{Verified} & \textbf{Falsified} & \textbf{Fastest} & \textbf{Penalty} & \textbf{Points} & \textbf{Score} & \textbf{Solved}\\
\midrule
1 & $\alpha$-$\beta$-CROWN & 58 & 5 & 0 & 0 & 630 & 100.0 & 91.3\% \\
2 & SobolBox & 54 & 3 & 0 & 0 & 570 & 90.5 & 82.6\% \\
3 & PyRAT & 36 & 3 & 0 & 0 & 390 & 61.9 & 56.5\% \\
4 & NNV & 17 & 0 & 0 & 0 & 170 & 27.0 & 24.6\% \\
\bottomrule
\end{tabular}
}
\end{center}
\end{table}

\begin{figure}[h]
\centerline{\includegraphics[width=\textwidth]{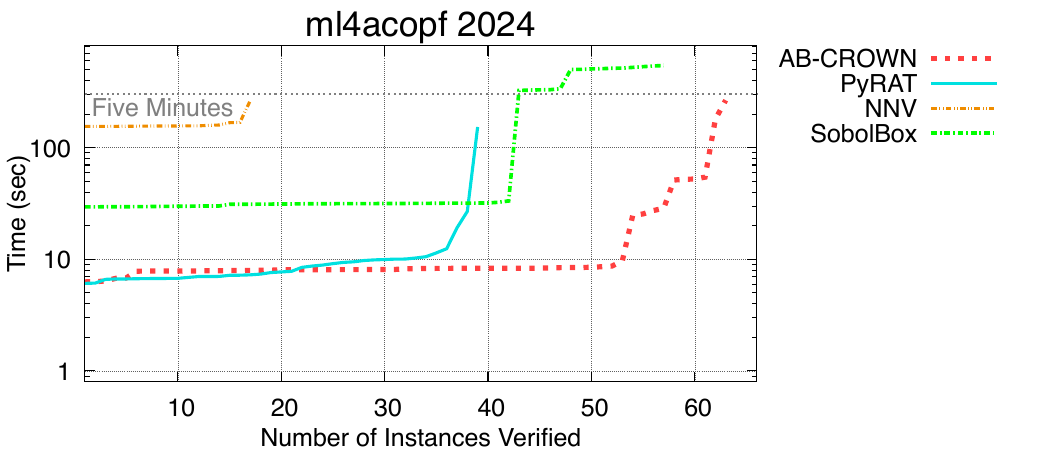}}
\caption{Cactus Plot for ml4acopf 2024.}
\label{fig:quantPic_ml4acopf}
\end{figure}

\clearpage

\begin{table}[h]
\begin{center}
\caption{Benchmark \texttt{2025-relusplitter}} \label{tab:cat_2025_relusplitter}
{\setlength{\tabcolsep}{2pt}
\begin{tabular}[h]{@{}llllllrrr@{}}
\toprule
\textbf{\# ~} & \textbf{Tool} & \textbf{Verified} & \textbf{Falsified} & \textbf{Fastest} & \textbf{Penalty} & \textbf{Points} & \textbf{Score} & \textbf{Solved}\\
\midrule
1 & $\alpha$-$\beta$-CROWN & 151 & 20 & 0 & 0 & 1710 & 100.0 & 77.7\% \\
2 & PyRAT & 61 & 16 & 0 & 0 & 770 & 45.0 & 35.0\% \\
3 & nnenum & 22 & 2 & 0 & 0 & 240 & 14.0 & 10.9\% \\
4 & CORA & 6 & 0 & 0 & 0 & 60 & 3.5 & 2.7\% \\
5 & NNV & 0 & 4 & 0 & 0 & 40 & 2.3 & 1.8\% \\
\bottomrule
\end{tabular}
}
\end{center}
\end{table}

\begin{figure}[h]
\centerline{\includegraphics[width=\textwidth]{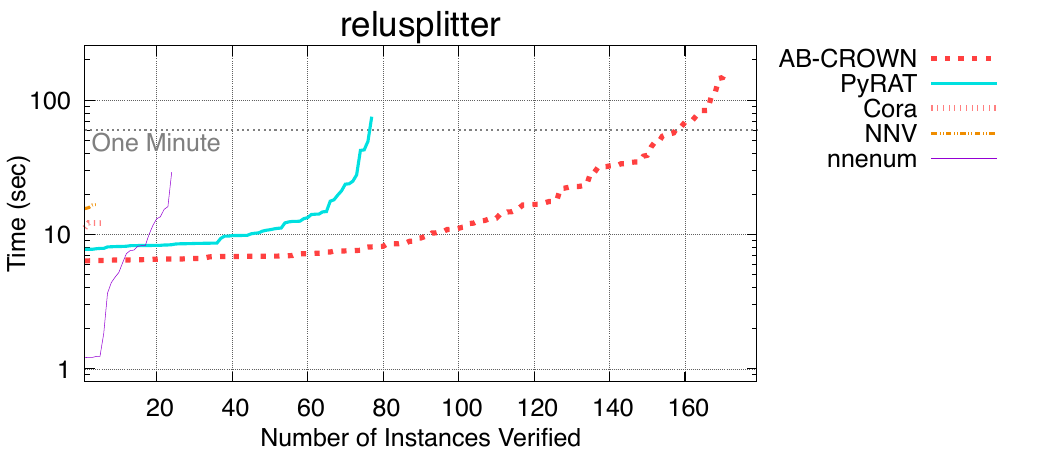}}
\caption{Cactus Plot for relusplitter.}
\label{fig:quantPic_relusplitter}
\end{figure}

\clearpage

\begin{table}[h]
\begin{center}
\caption{Benchmark \texttt{2025-traffic-signs-recognition-2023}} \label{tab:cat_2025_traffic_signs_recognition_2023}
{\setlength{\tabcolsep}{2pt}
\begin{tabular}[h]{@{}llllllrrr@{}}
\toprule
\textbf{\# ~} & \textbf{Tool} & \textbf{Verified} & \textbf{Falsified} & \textbf{Fastest} & \textbf{Penalty} & \textbf{Points} & \textbf{Score} & \textbf{Solved}\\
\midrule
1 & $\alpha$-$\beta$-CROWN & 0 & 43 & 0 & 0 & 430 & 100.0 & 95.6\% \\
\bottomrule
\end{tabular}
}
\end{center}
\end{table}

\begin{figure}[h]
\centerline{\includegraphics[width=\textwidth]{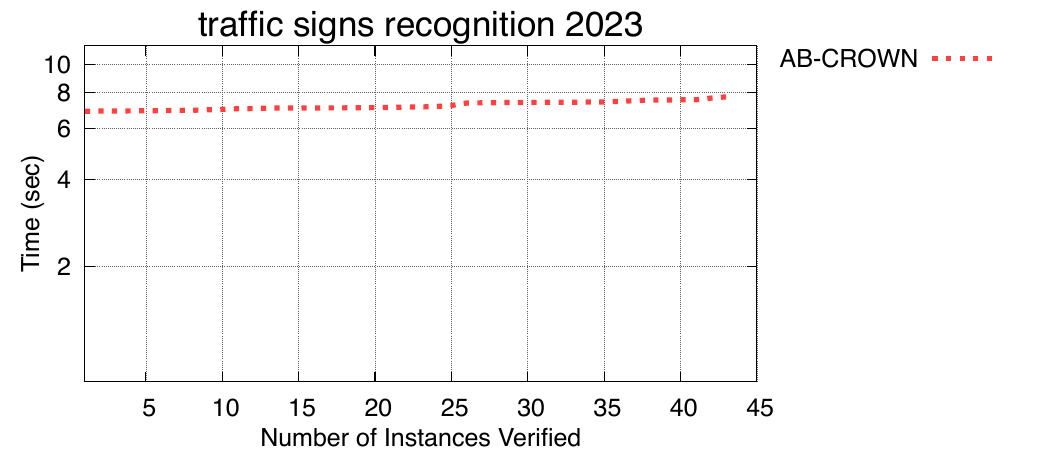}}
\caption{Cactus Plot for traffic signs recognition 2023.}
\label{fig:quantPic_traffic}
\end{figure}

\clearpage

\begin{table}[h]
\begin{center}
\caption{Benchmark \texttt{2025-vggnet16-2022}} \label{tab:cat_2025_vggnet16_2022}
{\setlength{\tabcolsep}{2pt}
\begin{tabular}[h]{@{}llllllrrr@{}}
\toprule
\textbf{\# ~} & \textbf{Tool} & \textbf{Verified} & \textbf{Falsified} & \textbf{Fastest} & \textbf{Penalty} & \textbf{Points} & \textbf{Score} & \textbf{Solved}\\
\midrule
1 & $\alpha$-$\beta$-CROWN & 17 & 1 & 0 & 0 & 180 & 100.0 & 100.0\% \\
2 & nnenum & 14 & 1 & 0 & 0 & 150 & 83.3 & 83.3\% \\
3 & PyRAT & 13 & 1 & 0 & 0 & 140 & 77.8 & 77.8\% \\
4 & NNV & 0 & 1 & 0 & 0 & 10 & 5.6 & 5.6\% \\
\bottomrule
\end{tabular}
}
\end{center}
\end{table}

\begin{figure}[h]
\centerline{\includegraphics[width=\textwidth]{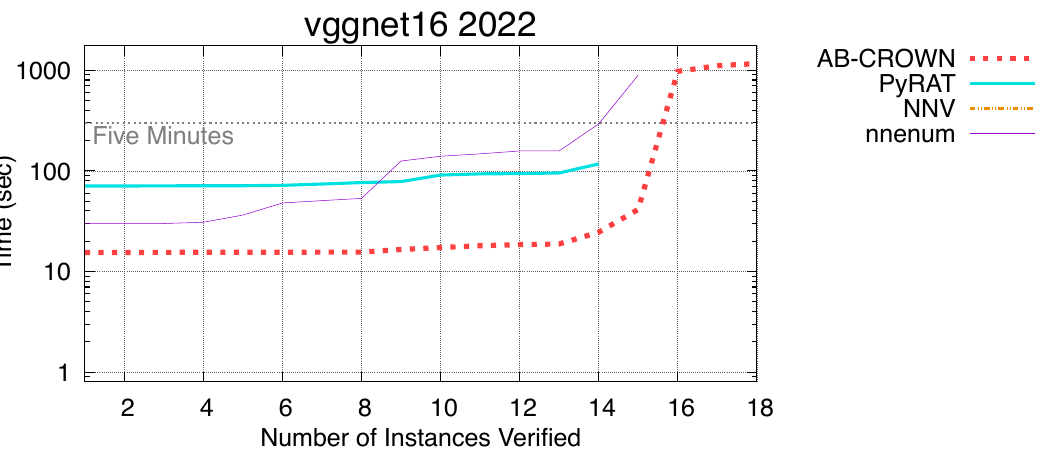}}
\caption{Cactus Plot for vggnet16 2022.}
\label{fig:quantPic_vggnet16}
\end{figure}

\clearpage

\begin{table}[h]
\begin{center}
\caption{Benchmark \texttt{2025-vit-2023}} \label{tab:cat_2025_vit_2023}
{\setlength{\tabcolsep}{2pt}
\begin{tabular}[h]{@{}llllllrrr@{}}
\toprule
\textbf{\# ~} & \textbf{Tool} & \textbf{Verified} & \textbf{Falsified} & \textbf{Fastest} & \textbf{Penalty} & \textbf{Points} & \textbf{Score} & \textbf{Solved}\\
\midrule
1 & $\alpha$-$\beta$-CROWN & 125 & 0 & 0 & 0 & 1250 & 100.0 & 62.5\% \\
2 & PyRAT & 83 & 0 & 0 & 0 & 830 & 66.4 & 41.5\% \\
3 & NeuralSAT & 67 & 0 & 0 & 0 & 670 & 53.6 & 33.5\% \\
\bottomrule
\end{tabular}
}
\end{center}
\end{table}

\begin{figure}[h]
\centerline{\includegraphics[width=\textwidth]{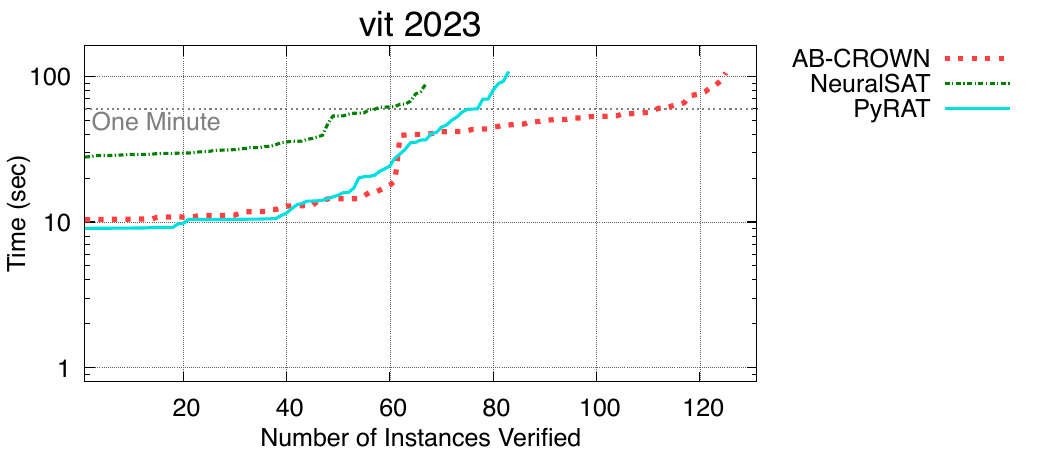}}
\caption{Cactus Plot for vit 2023.}
\label{fig:quantPic_vit}
\end{figure}

\clearpage

\begin{table}[h]
\begin{center}
\caption{Benchmark \texttt{2025-yolo-2023}} \label{tab:cat_2025_yolo_2023}
{\setlength{\tabcolsep}{2pt}
\begin{tabular}[h]{@{}llllllrrr@{}}
\toprule
\textbf{\# ~} & \textbf{Tool} & \textbf{Verified} & \textbf{Falsified} & \textbf{Fastest} & \textbf{Penalty} & \textbf{Points} & \textbf{Score} & \textbf{Solved}\\
\midrule
1 & NNV & 71 & 0 & 0 & 0 & 710 & 100.0 & 98.6\% \\
2 & $\alpha$-$\beta$-CROWN & 61 & 0 & 0 & 0 & 610 & 85.9 & 84.7\% \\
3 & NeuralSAT & 52 & 0 & 0 & 0 & 520 & 73.2 & 72.2\% \\
4 & PyRAT & 40 & 0 & 0 & 0 & 400 & 56.3 & 55.6\% \\
\bottomrule
\end{tabular}
}
\end{center}
\end{table}

\begin{figure}[h]
\centerline{\includegraphics[width=\textwidth]{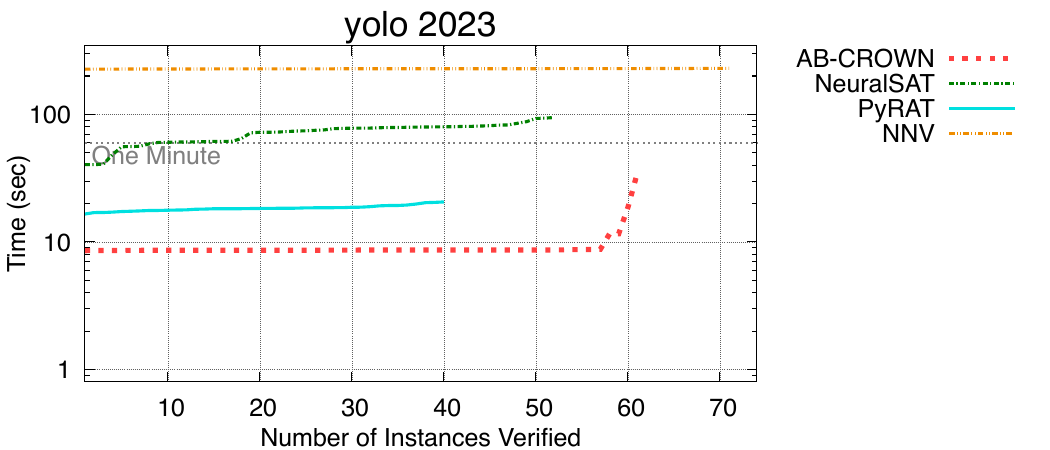}}
\caption{Cactus Plot for yolo 2023.}
\label{fig:quantPic_yolo}
\end{figure}


\subsection{Small Tolerance}
\subsubsection{Regular Track}
\label{sec:benchmark_results_regular_tol}


\begin{table}[h]
\begin{center}
\caption{Benchmark \texttt{2025-acasxu-2023}} \label{tab:cat_2025_acasxu_2023_tol}
{\setlength{\tabcolsep}{2pt}
\begin{tabular}[h]{@{}llllllrrr@{}}
\toprule
\textbf{\# ~} & \textbf{Tool} & \textbf{Verified} & \textbf{Falsified} & \textbf{Fastest} & \textbf{Penalty} & \textbf{Points} & \textbf{Score} & \textbf{Solved}\\
\midrule
1 & nnenum & 139 & 47 & 0 & 0 & 1860 & 100.0 & 100.0\% \\
2 & NeuralSAT & 139 & 47 & 0 & 0 & 1860 & 100.0 & 100.0\% \\
3 & $\alpha$-$\beta$-CROWN & 139 & 47 & 0 & 0 & 1860 & 100.0 & 100.0\% \\
4 & PyRAT & 139 & 46 & 0 & 0 & 1850 & 99.5 & 99.5\% \\
5 & CORA & 137 & 46 & 0 & 0 & 1830 & 98.4 & 98.4\% \\
6 & SobolBox & 118 & 43 & 0 & 1 & 1460 & 78.5 & 86.6\% \\
7 & NNV & 71 & 39 & 0 & 0 & 1100 & 59.1 & 59.1\% \\
\bottomrule
\end{tabular}
}
\end{center}
\end{table}

\begin{figure}[h]
\centerline{\includegraphics[width=\textwidth]{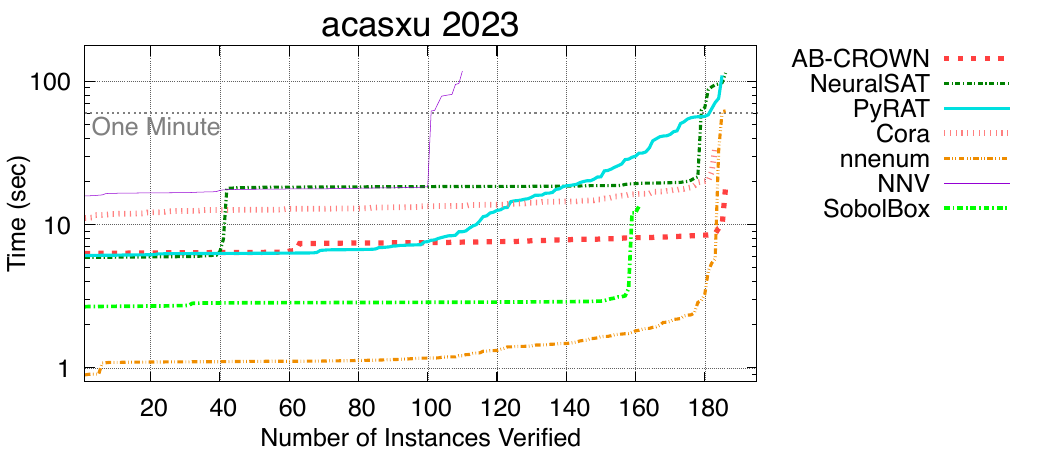}}
\caption{Cactus Plot for acasxu 2023.}
\label{fig:quantPic_acasxu_tol} 
\end{figure}

\clearpage

\begin{table}[h]
\begin{center}
\caption{Benchmark \texttt{2025-cersyve}} \label{tab:cat_2025_cersyve_tol}
{\setlength{\tabcolsep}{2pt}
\begin{tabular}[h]{@{}llllllrrr@{}}
\toprule
\textbf{\# ~} & \textbf{Tool} & \textbf{Verified} & \textbf{Falsified} & \textbf{Fastest} & \textbf{Penalty} & \textbf{Points} & \textbf{Score} & \textbf{Solved}\\
\midrule
1 & $\alpha$-$\beta$-CROWN & 6 & 6 & 0 & 0 & 120 & 100.0 & 100.0\% \\
2 & NeuralSAT & 4 & 6 & 0 & 0 & 100 & 83.3 & 83.3\% \\
3 & PyRAT & 2 & 6 & 0 & 0 & 80 & 66.7 & 66.7\% \\
4 & SobolBox & 0 & 6 & 0 & 0 & 60 & 50.0 & 50.0\% \\
5 & NNV & 0 & 3 & 0 & 0 & 30 & 25.0 & 25.0\% \\
6 & CORA & 6 & 5 & 0 & 1 & -40 & 0 & 91.7\% \\
\bottomrule
\end{tabular}
}
\end{center}
\end{table}

\begin{figure}[h]
\centerline{\includegraphics[width=\textwidth]{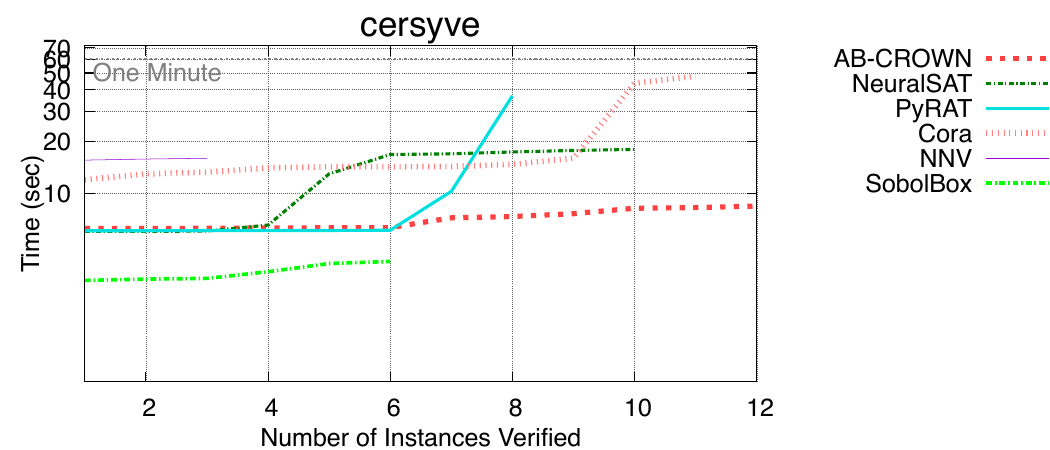}}
\caption{Cactus Plot for cersyve.}
\label{fig:quantPic_cersyve_tol}
\end{figure}

\clearpage

\begin{table}[h]
\begin{center}
\caption{Benchmark \texttt{2025-cgan-2023}} \label{tab:cat_2025_cgan_2023_tol}
{\setlength{\tabcolsep}{2pt}
\begin{tabular}[h]{@{}llllllrrr@{}}
\toprule
\textbf{\# ~} & \textbf{Tool} & \textbf{Verified} & \textbf{Falsified} & \textbf{Fastest} & \textbf{Penalty} & \textbf{Points} & \textbf{Score} & \textbf{Solved}\\
\midrule
1 & PyRAT & 9 & 12 & 0 & 0 & 210 & 100.0 & 100.0\% \\
2 & NeuralSAT & 9 & 12 & 0 & 0 & 210 & 100.0 & 100.0\% \\
3 & $\alpha$-$\beta$-CROWN & 9 & 12 & 0 & 0 & 210 & 100.0 & 100.0\% \\
4 & SobolBox & 9 & 10 & 0 & 0 & 190 & 90.5 & 90.5\% \\
5 & nnenum & 7 & 10 & 0 & 0 & 170 & 81.0 & 81.0\% \\
6 & NNV & 5 & 11 & 0 & 0 & 160 & 76.2 & 76.2\% \\
\bottomrule
\end{tabular}
}
\end{center}
\end{table}

\begin{figure}[h]
\centerline{\includegraphics[width=\textwidth]{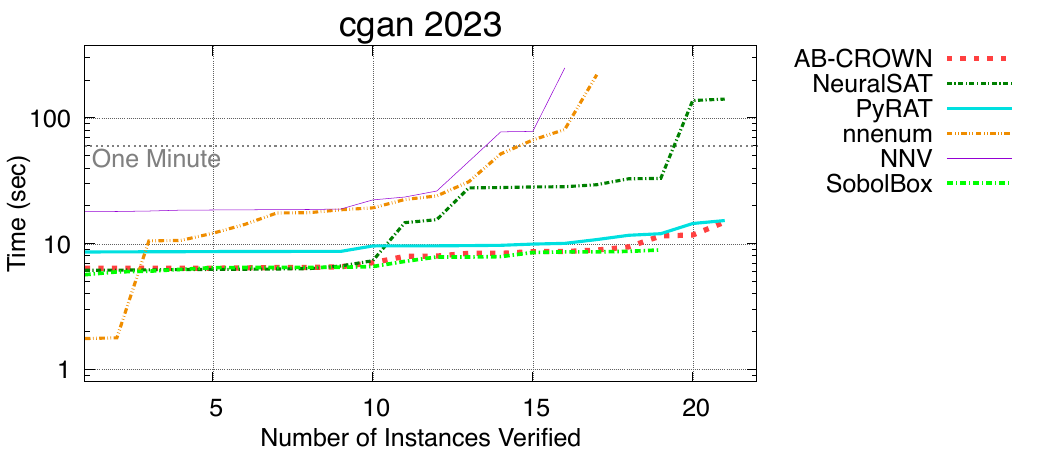}}
\caption{Cactus Plot for cgan 2023.}
\label{fig:quantPic_cgan_tol}
\end{figure}

\clearpage

\begin{table}[h]
\begin{center}
\caption{Benchmark \texttt{2025-cifar100-2024}} \label{tab:cat_2025_cifar100_2024_tol}
{\setlength{\tabcolsep}{2pt}
\begin{tabular}[h]{@{}llllllrrr@{}}
\toprule
\textbf{\# ~} & \textbf{Tool} & \textbf{Verified} & \textbf{Falsified} & \textbf{Fastest} & \textbf{Penalty} & \textbf{Points} & \textbf{Score} & \textbf{Solved}\\
\midrule
1 & $\alpha$-$\beta$-CROWN & 100 & 29 & 0 & 0 & 1290 & 100.0 & 64.5\% \\
2 & PyRAT & 61 & 25 & 0 & 0 & 860 & 66.7 & 43.0\% \\
3 & NeuralSAT & 87 & 25 & 0 & 4 & 520 & 40.3 & 56.0\% \\
4 & CORA & 0 & 10 & 0 & 0 & 100 & 7.8 & 5.0\% \\
5 & NNV & 171 & 0 & 0 & 19 & -1140 & 0 & 85.5\% \\
\bottomrule
\end{tabular}
}
\end{center}
\end{table}

\begin{figure}[h]
\centerline{\includegraphics[width=\textwidth]{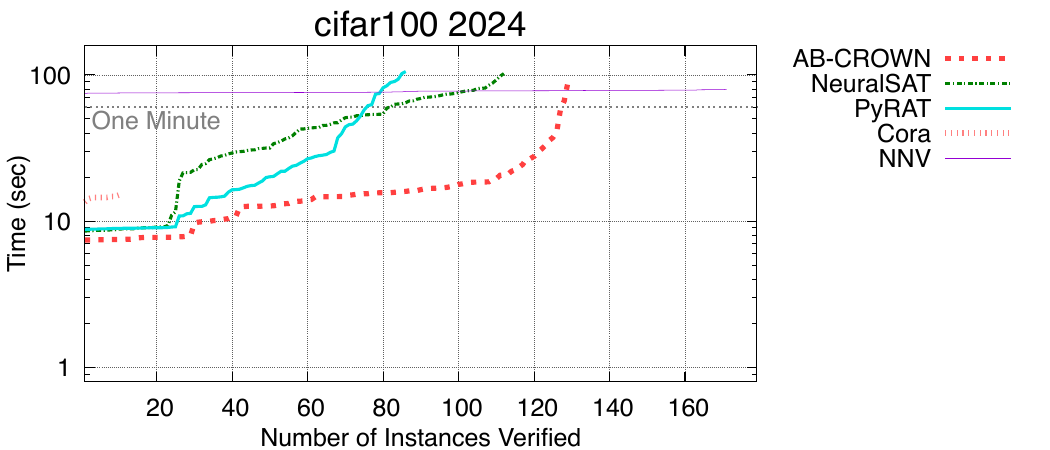}}
\caption{Cactus Plot for cifar100 2024.}
\label{fig:quantPic_cifar100_tol}
\end{figure}

\clearpage

\begin{table}[h]
\begin{center}
\caption{Benchmark \texttt{2025-collins-rul-cnn-2022}} \label{tab:cat_2025_collins_rul_cnn_2022_tol}
{\setlength{\tabcolsep}{2pt}
\begin{tabular}[h]{@{}llllllrrr@{}}
\toprule
\textbf{\# ~} & \textbf{Tool} & \textbf{Verified} & \textbf{Falsified} & \textbf{Fastest} & \textbf{Penalty} & \textbf{Points} & \textbf{Score} & \textbf{Solved}\\
\midrule
1 & nnenum & 39 & 23 & 0 & 0 & 620 & 100.0 & 100.0\% \\
2 & PyRAT & 39 & 23 & 0 & 0 & 620 & 100.0 & 100.0\% \\
3 & NeuralSAT & 39 & 23 & 0 & 0 & 620 & 100.0 & 100.0\% \\
4 & NNV & 39 & 23 & 0 & 0 & 620 & 100.0 & 100.0\% \\
5 & CORA & 39 & 23 & 0 & 0 & 620 & 100.0 & 100.0\% \\
6 & $\alpha$-$\beta$-CROWN & 39 & 23 & 0 & 0 & 620 & 100.0 & 100.0\% \\
7 & SobolBox & 19 & 15 & 0 & 0 & 340 & 54.8 & 54.8\% \\
\bottomrule
\end{tabular}
}
\end{center}
\end{table}

\begin{figure}[h]
\centerline{\includegraphics[width=\textwidth]{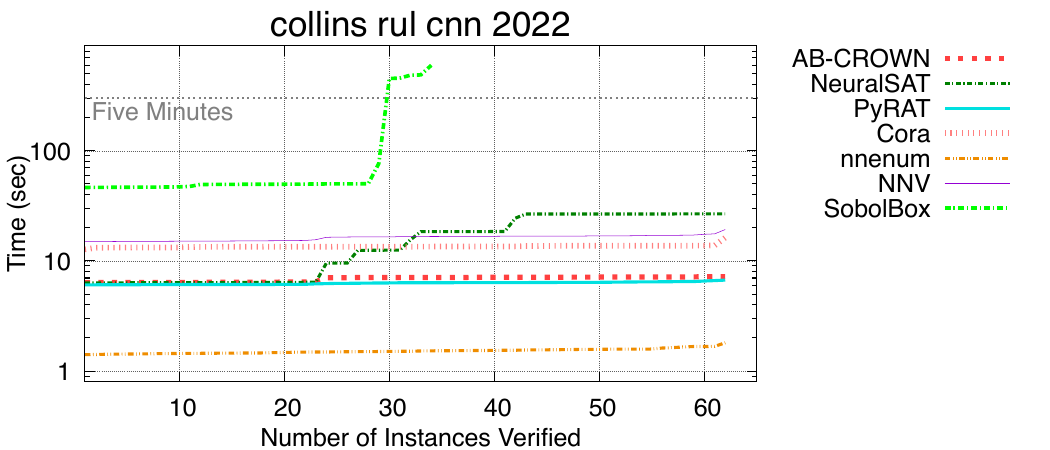}}
\caption{Cactus Plot for collins rul cnn 2022.}
\label{fig:quantPic_rul_tol}
\end{figure}

\clearpage

\begin{table}[h]
\begin{center}
\caption{Benchmark \texttt{2025-cora-2024}} \label{tab:cat_2025_cora_2024_tol}
{\setlength{\tabcolsep}{2pt}
\begin{tabular}[h]{@{}llllllrrr@{}}
\toprule
\textbf{\# ~} & \textbf{Tool} & \textbf{Verified} & \textbf{Falsified} & \textbf{Fastest} & \textbf{Penalty} & \textbf{Points} & \textbf{Score} & \textbf{Solved}\\
\midrule
1 & $\alpha$-$\beta$-CROWN & 22 & 131 & 0 & 0 & 1530 & 100.0 & 85.0\% \\
2 & NeuralSAT & 21 & 131 & 0 & 0 & 1520 & 99.3 & 84.4\% \\
3 & PyRAT & 20 & 128 & 0 & 0 & 1480 & 96.7 & 82.2\% \\
4 & CORA & 19 & 124 & 0 & 0 & 1430 & 93.5 & 79.4\% \\
5 & NNV & 19 & 57 & 0 & 0 & 760 & 49.7 & 42.2\% \\
6 & nnenum & 19 & 4 & 0 & 0 & 230 & 15.0 & 12.8\% \\
\bottomrule
\end{tabular}
}
\end{center}
\end{table}

\begin{figure}[h]
\centerline{\includegraphics[width=\textwidth]{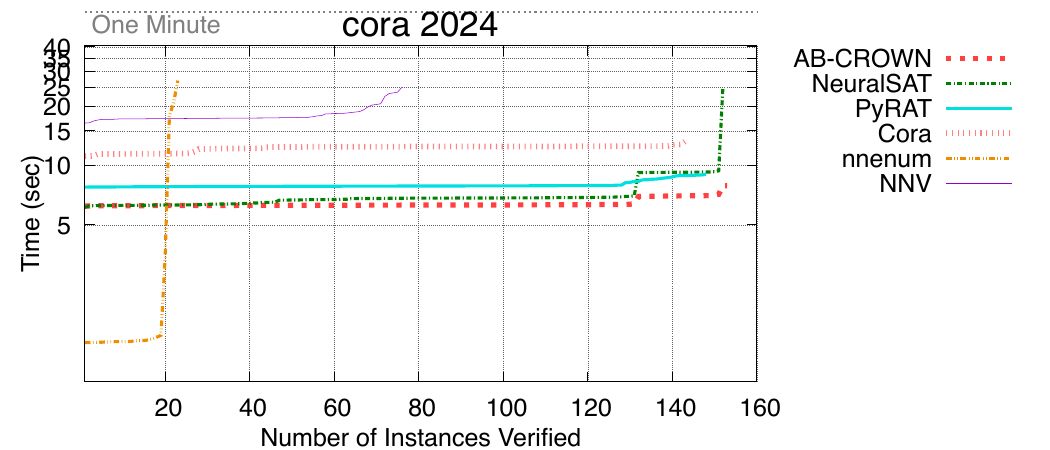}}
\caption{Cactus Plot for cora 2024.}
\label{fig:quantPic_cora_tol}
\end{figure}

\clearpage

\begin{table}[h]
\begin{center}
\caption{Benchmark \texttt{2025-dist-shift-2023}} \label{tab:cat_2025_dist_shift_2023_tol}
{\setlength{\tabcolsep}{2pt}
\begin{tabular}[h]{@{}llllllrrr@{}}
\toprule
\textbf{\# ~} & \textbf{Tool} & \textbf{Verified} & \textbf{Falsified} & \textbf{Fastest} & \textbf{Penalty} & \textbf{Points} & \textbf{Score} & \textbf{Solved}\\
\midrule
1 & NeuralSAT & 65 & 7 & 0 & 0 & 720 & 100.0 & 100.0\% \\
2 & CORA & 65 & 7 & 0 & 0 & 720 & 100.0 & 100.0\% \\
3 & $\alpha$-$\beta$-CROWN & 65 & 7 & 0 & 0 & 720 & 100.0 & 100.0\% \\
4 & PyRAT & 64 & 7 & 0 & 0 & 710 & 98.6 & 98.6\% \\
5 & NNV & 50 & 4 & 0 & 0 & 540 & 75.0 & 75.0\% \\
\bottomrule
\end{tabular}
}
\end{center}
\end{table}

\begin{figure}[h]
\centerline{\includegraphics[width=\textwidth]{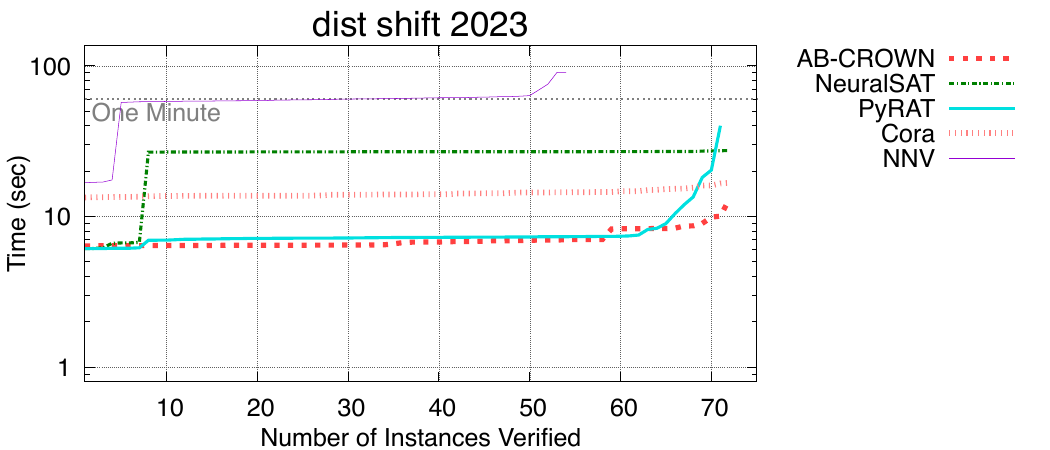}}
\caption{Cactus Plot for dist shift 2023.}
\label{fig:quantPic_dist_shift_tol}
\end{figure}

\clearpage

\begin{table}[h]
\begin{center}
\caption{Benchmark \texttt{2025-linearizenn-2024}} \label{tab:cat_2025_linearizenn_2024_tol}
{\setlength{\tabcolsep}{2pt}
\begin{tabular}[h]{@{}llllllrrr@{}}
\toprule
\textbf{\# ~} & \textbf{Tool} & \textbf{Verified} & \textbf{Falsified} & \textbf{Fastest} & \textbf{Penalty} & \textbf{Points} & \textbf{Score} & \textbf{Solved}\\
\midrule
1 & nnenum & 59 & 1 & 0 & 0 & 600 & 100.0 & 100.0\% \\
2 & SobolBox & 59 & 1 & 0 & 0 & 600 & 100.0 & 100.0\% \\
3 & PyRAT & 59 & 1 & 0 & 0 & 600 & 100.0 & 100.0\% \\
4 & NeuralSAT & 59 & 1 & 0 & 0 & 600 & 100.0 & 100.0\% \\
5 & CORA & 59 & 1 & 0 & 0 & 600 & 100.0 & 100.0\% \\
6 & $\alpha$-$\beta$-CROWN & 59 & 1 & 0 & 0 & 600 & 100.0 & 100.0\% \\
7 & NNV & 40 & 1 & 0 & 0 & 410 & 68.3 & 68.3\% \\
\bottomrule
\end{tabular}
}
\end{center}
\end{table}

\begin{figure}[h]
\centerline{\includegraphics[width=\textwidth]{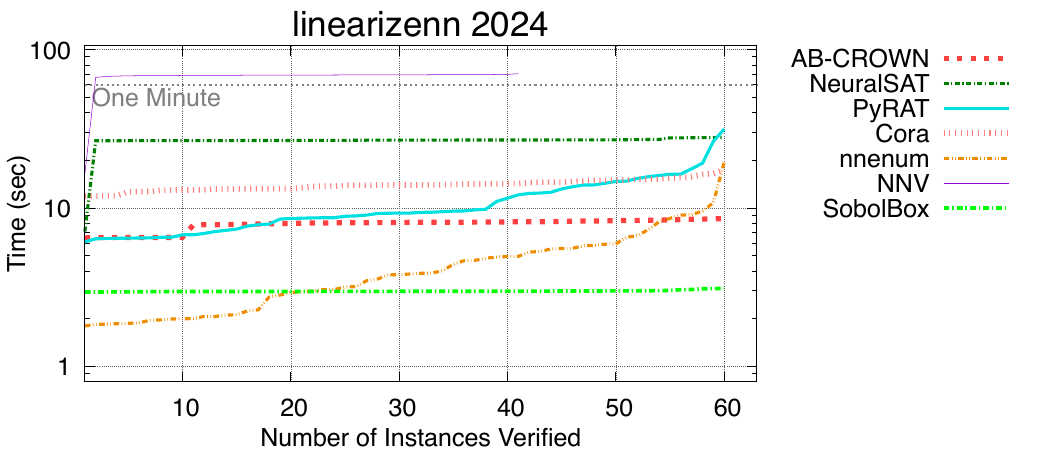}}
\caption{Cactus Plot for linearizenn 2024.}
\label{fig:quantPic_linearizenn_tol}
\end{figure}

\clearpage

\begin{table}[h]
\begin{center}
\caption{Benchmark \texttt{2025-malbeware}} \label{tab:cat_2025_malbeware_tol}
{\setlength{\tabcolsep}{2pt}
\begin{tabular}[h]{@{}llllllrrr@{}}
\toprule
\textbf{\# ~} & \textbf{Tool} & \textbf{Verified} & \textbf{Falsified} & \textbf{Fastest} & \textbf{Penalty} & \textbf{Points} & \textbf{Score} & \textbf{Solved}\\
\midrule
1 & $\alpha$-$\beta$-CROWN & 131 & 19 & 0 & 0 & 1500 & 100.0 & 100.0\% \\
2 & NeuralSAT & 127 & 19 & 0 & 0 & 1460 & 97.3 & 97.3\% \\
3 & PyRAT & 121 & 18 & 0 & 0 & 1390 & 92.7 & 92.7\% \\
4 & nnenum & 125 & 12 & 0 & 0 & 1370 & 91.3 & 91.3\% \\
5 & CORA & 88 & 9 & 0 & 0 & 970 & 64.7 & 64.7\% \\
6 & NNV & 49 & 4 & 0 & 0 & 530 & 35.3 & 35.3\% \\
\bottomrule
\end{tabular}
}
\end{center}
\end{table}

\begin{figure}[h]
\centerline{\includegraphics[width=\textwidth]{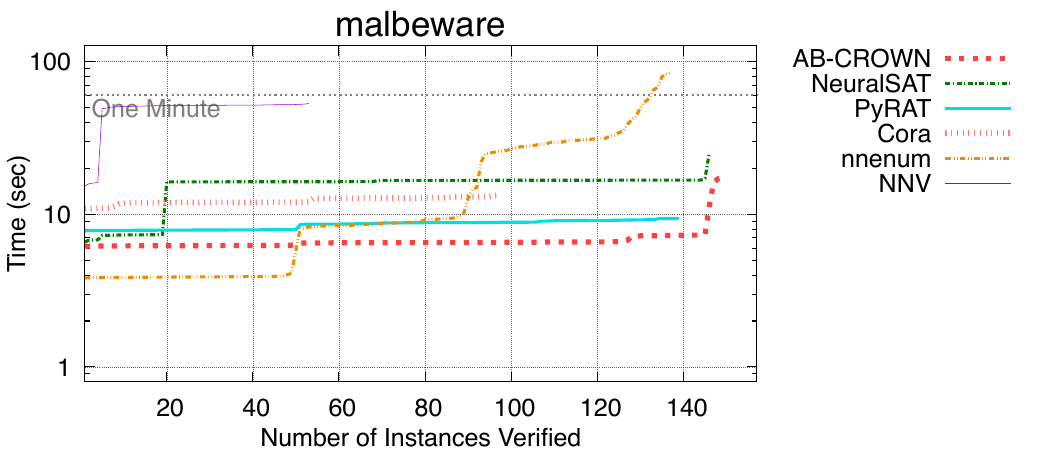}}
\caption{Cactus Plot for malbeware.}
\label{fig:quantPic_malbeware_tol}
\end{figure}

\clearpage

\begin{table}[h]
\begin{center}
\caption{Benchmark \texttt{2025-metaroom-2023}} \label{tab:cat_2025_metaroom_2023_tol}
{\setlength{\tabcolsep}{2pt}
\begin{tabular}[h]{@{}llllllrrr@{}}
\toprule
\textbf{\# ~} & \textbf{Tool} & \textbf{Verified} & \textbf{Falsified} & \textbf{Fastest} & \textbf{Penalty} & \textbf{Points} & \textbf{Score} & \textbf{Solved}\\
\midrule
1 & NeuralSAT & 94 & 5 & 0 & 0 & 990 & 100.0 & 99.0\% \\
2 & $\alpha$-$\beta$-CROWN & 94 & 5 & 0 & 0 & 990 & 100.0 & 99.0\% \\
3 & CORA & 92 & 5 & 0 & 0 & 970 & 98.0 & 97.0\% \\
4 & NNV & 93 & 2 & 0 & 0 & 950 & 96.0 & 95.0\% \\
5 & PyRAT & 95 & 3 & 0 & 2 & 680 & 68.7 & 98.0\% \\
6 & nnenum & 50 & 1 & 0 & 0 & 510 & 51.5 & 51.0\% \\
\bottomrule
\end{tabular}
}
\end{center}
\end{table}

\begin{figure}[h]
\centerline{\includegraphics[width=\textwidth]{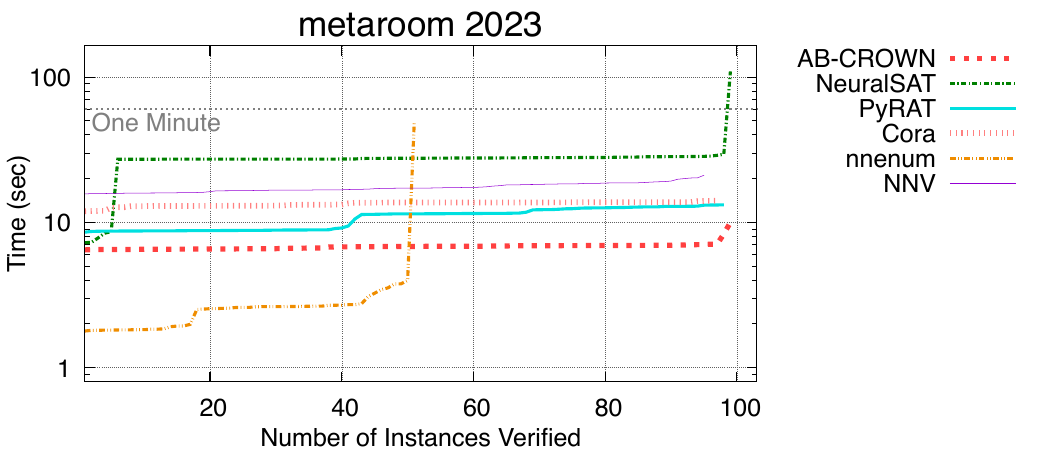}}
\caption{Cactus Plot for metaroom 2023.}
\label{fig:quantPic_metaroom_tol}
\end{figure}

\clearpage

\begin{table}[h]
\begin{center}
\caption{Benchmark \texttt{2025-nn4sys}} \label{tab:cat_2025_nn4sys_tol}
{\setlength{\tabcolsep}{2pt}
\begin{tabular}[h]{@{}llllllrrr@{}}
\toprule
\textbf{\# ~} & \textbf{Tool} & \textbf{Verified} & \textbf{Falsified} & \textbf{Fastest} & \textbf{Penalty} & \textbf{Points} & \textbf{Score} & \textbf{Solved}\\
\midrule
1 & $\alpha$-$\beta$-CROWN & 194 & 0 & 0 & 0 & 1940 & 100.0 & 100.0\% \\
2 & NeuralSAT & 120 & 0 & 0 & 0 & 1200 & 61.9 & 61.9\% \\
3 & SobolBox & 107 & 0 & 0 & 0 & 1070 & 55.2 & 55.2\% \\
4 & PyRAT & 40 & 0 & 0 & 0 & 400 & 20.6 & 20.6\% \\
5 & nnenum & 22 & 0 & 0 & 0 & 220 & 11.3 & 11.3\% \\
6 & NNV & 17 & 0 & 0 & 0 & 170 & 8.8 & 8.8\% \\
\bottomrule
\end{tabular}
}
\end{center}
\end{table}

\begin{figure}[h]
\centerline{\includegraphics[width=\textwidth]{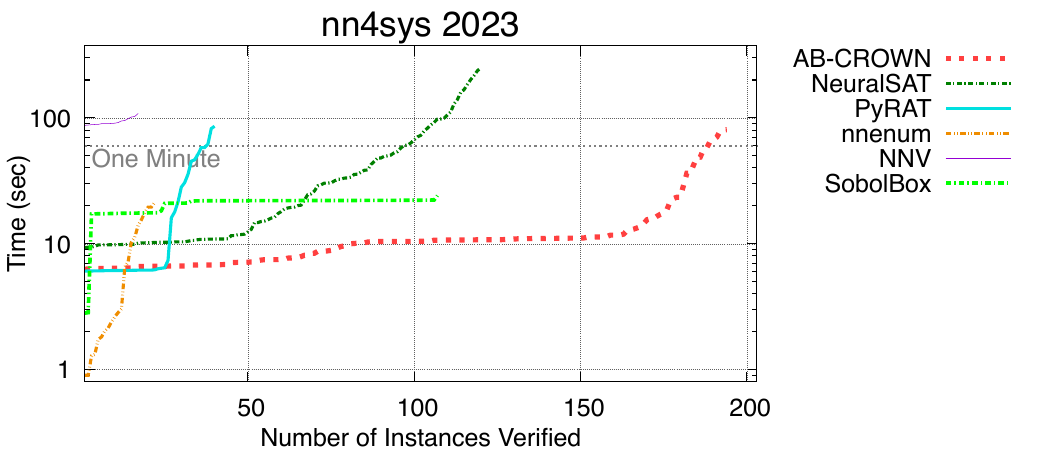}}
\caption{Cactus Plot for nn4sys 2023.}
\label{fig:quantPic_nn4sys_tol}
\end{figure}

\clearpage

\begin{table}[h]
\begin{center}
\caption{Benchmark \texttt{2025-safenlp-2024}} \label{tab:cat_2025_safenlp_2024_tol}
{\setlength{\tabcolsep}{2pt}
\begin{tabular}[h]{@{}llllllrrr@{}}
\toprule
\textbf{\# ~} & \textbf{Tool} & \textbf{Verified} & \textbf{Falsified} & \textbf{Fastest} & \textbf{Penalty} & \textbf{Points} & \textbf{Score} & \textbf{Solved}\\
\midrule
1 & $\alpha$-$\beta$-CROWN & 433 & 647 & 0 & 0 & 10800 & 100.0 & 100.0\% \\
2 & NeuralSAT & 425 & 645 & 0 & 0 & 10700 & 99.1 & 99.1\% \\
3 & PyRAT & 331 & 647 & 0 & 0 & 9780 & 90.6 & 90.6\% \\
4 & nnenum & 340 & 636 & 0 & 0 & 9760 & 90.4 & 90.4\% \\
5 & CORA & 338 & 644 & 0 & 1 & 9670 & 89.5 & 90.9\% \\
6 & NNV & 172 & 176 & 0 & 0 & 3480 & 32.2 & 32.2\% \\
7 & SobolBox & 413 & 215 & 0 & 403 & -54170 & 0 & 58.1\% \\
\bottomrule
\end{tabular}
}
\end{center}
\end{table}

\begin{figure}[h]
\centerline{\includegraphics[width=\textwidth]{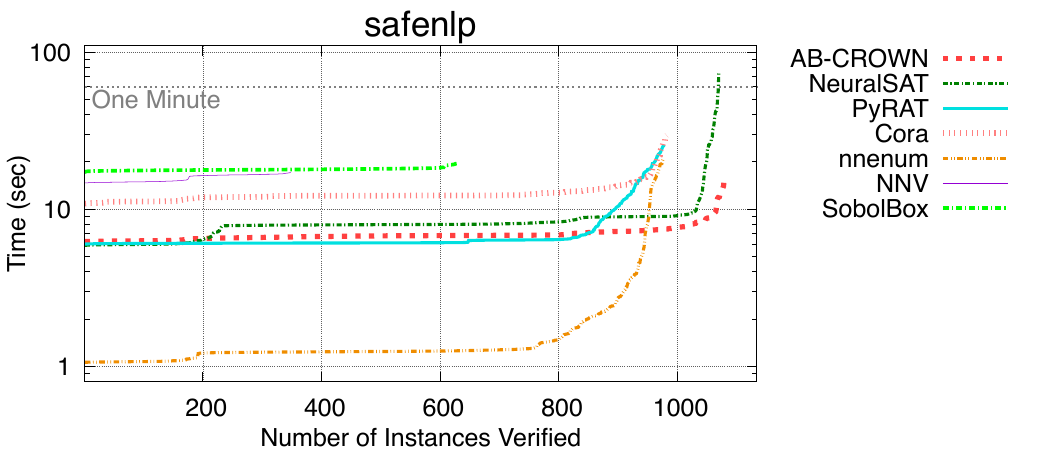}}
\caption{Cactus Plot for safenlp.}
\label{fig:quantPic_safenlp_tol}
\end{figure}

\clearpage

\begin{table}[h]
\begin{center}
\caption{Benchmark \texttt{2025-sat-relu}} \label{tab:cat_2025_sat_relu_tol}
{\setlength{\tabcolsep}{2pt}
\begin{tabular}[h]{@{}llllllrrr@{}}
\toprule
\textbf{\# ~} & \textbf{Tool} & \textbf{Verified} & \textbf{Falsified} & \textbf{Fastest} & \textbf{Penalty} & \textbf{Points} & \textbf{Score} & \textbf{Solved}\\
\midrule
1 & NeuralSAT & 50 & 50 & 0 & 0 & 1000 & 100.0 & 100.0\% \\
2 & CORA & 50 & 50 & 0 & 0 & 1000 & 100.0 & 100.0\% \\
3 & $\alpha$-$\beta$-CROWN & 50 & 50 & 0 & 0 & 1000 & 100.0 & 100.0\% \\
4 & PyRAT & 9 & 50 & 0 & 0 & 590 & 59.0 & 59.0\% \\
5 & nnenum & 9 & 35 & 0 & 0 & 440 & 44.0 & 44.0\% \\
6 & NNV & 2 & 16 & 0 & 0 & 180 & 18.0 & 18.0\% \\
7 & SobolBox & 0 & 10 & 0 & 33 & -4850 & 0 & 10.0\% \\
\bottomrule
\end{tabular}
}
\end{center}
\end{table}

\begin{figure}[h]
\centerline{\includegraphics[width=\textwidth]{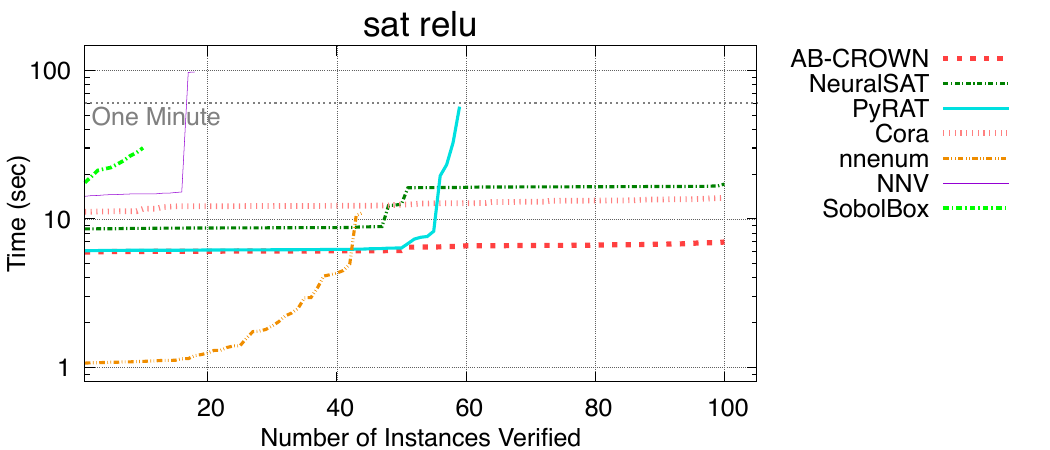}}
\caption{Cactus Plot for sat relu.}
\label{fig:quantPic_sat_relu_tol}
\end{figure}

\clearpage

\begin{table}[h]
\begin{center}
\caption{Benchmark \texttt{2025-soundnessbench}} \label{tab:cat_2025_soundnessbench_tol}
{\setlength{\tabcolsep}{2pt}
\begin{tabular}[h]{@{}llllllrrr@{}}
\toprule
\textbf{\# ~} & \textbf{Tool} & \textbf{Verified} & \textbf{Falsified} & \textbf{Fastest} & \textbf{Penalty} & \textbf{Points} & \textbf{Score} & \textbf{Solved}\\
\midrule
1 & $\alpha$-$\beta$-CROWN & 0 & 50 & 0 & 0 & 500 & 100.0 & 100.0\% \\
2 & NeuralSAT & 0 & 30 & 0 & 12 & -1500 & 0 & 60.0\% \\
3 & CORA & 0 & 0 & 0 & 18 & -2700 & 0 & 0.0\% \\
\bottomrule
\end{tabular}
}
\end{center}
\end{table}

\begin{figure}[h]
\centerline{\includegraphics[width=\textwidth]{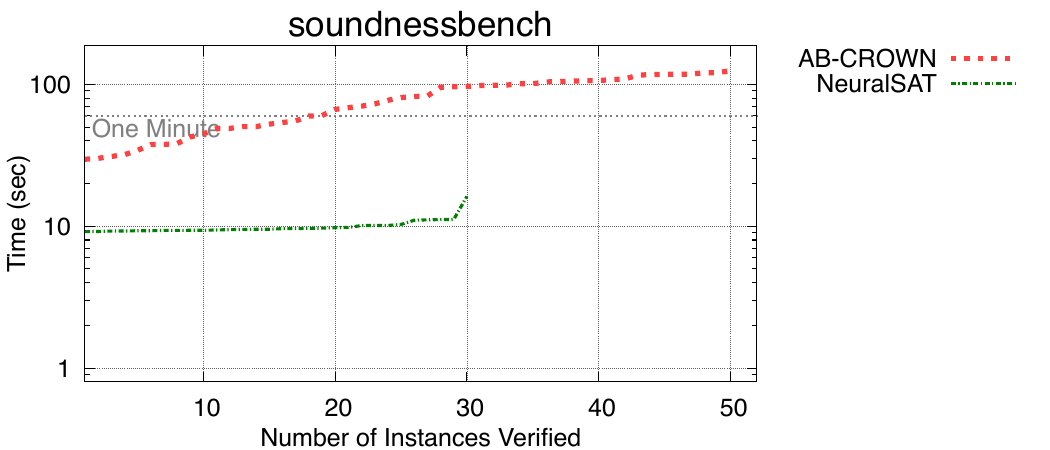}}
\caption{Cactus Plot for soundnessbench.}
\label{fig:quantPic_soundnessbench_tol}
\end{figure}

\clearpage

\begin{table}[h]
\begin{center}
\caption{Benchmark \texttt{2025-tinyimagenet-2024}} \label{tab:cat_2025_tinyimagenet_2024_tol}
{\setlength{\tabcolsep}{2pt}
\begin{tabular}[h]{@{}llllllrrr@{}}
\toprule
\textbf{\# ~} & \textbf{Tool} & \textbf{Verified} & \textbf{Falsified} & \textbf{Fastest} & \textbf{Penalty} & \textbf{Points} & \textbf{Score} & \textbf{Solved}\\
\midrule
1 & $\alpha$-$\beta$-CROWN & 137 & 38 & 0 & 0 & 1750 & 100.0 & 87.5\% \\
2 & NeuralSAT & 116 & 37 & 0 & 1 & 1380 & 78.9 & 76.5\% \\
3 & PyRAT & 68 & 35 & 0 & 0 & 1030 & 58.9 & 51.5\% \\
4 & CORA & 0 & 5 & 0 & 0 & 50 & 2.9 & 2.5\% \\
5 & NNV & 0 & 1 & 0 & 0 & 10 & 0.6 & 0.5\% \\
\bottomrule
\end{tabular}
}
\end{center}
\end{table}

\begin{figure}[h]
\centerline{\includegraphics[width=\textwidth]{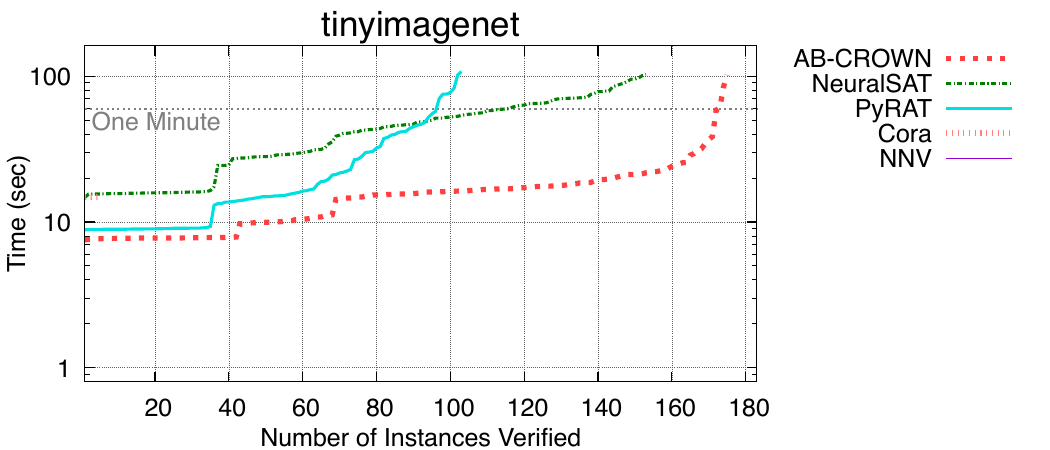}}
\caption{Cactus Plot for tinyimagenet.}
\label{fig:quantPic_tinyimagenet_tol}
\end{figure}

\clearpage

\begin{table}[h]
\begin{center}
\caption{Benchmark \texttt{2025-tllverifybench-2023}} \label{tab:cat_2025_tllverifybench_2023_tol}
{\setlength{\tabcolsep}{2pt}
\begin{tabular}[h]{@{}llllllrrr@{}}
\toprule
\textbf{\# ~} & \textbf{Tool} & \textbf{Verified} & \textbf{Falsified} & \textbf{Fastest} & \textbf{Penalty} & \textbf{Points} & \textbf{Score} & \textbf{Solved}\\
\midrule
1 & SobolBox & 15 & 17 & 0 & 0 & 320 & 100.0 & 100.0\% \\
2 & PyRAT & 15 & 17 & 0 & 0 & 320 & 100.0 & 100.0\% \\
3 & NeuralSAT & 15 & 17 & 0 & 0 & 320 & 100.0 & 100.0\% \\
4 & CORA & 15 & 17 & 0 & 0 & 320 & 100.0 & 100.0\% \\
5 & $\alpha$-$\beta$-CROWN & 15 & 17 & 0 & 0 & 320 & 100.0 & 100.0\% \\
6 & nnenum & 1 & 17 & 0 & 0 & 180 & 56.2 & 56.2\% \\
7 & NNV & 0 & 17 & 0 & 0 & 170 & 53.1 & 53.1\% \\
\bottomrule
\end{tabular}
}
\end{center}
\end{table}

\begin{figure}[h]
\centerline{\includegraphics[width=\textwidth]{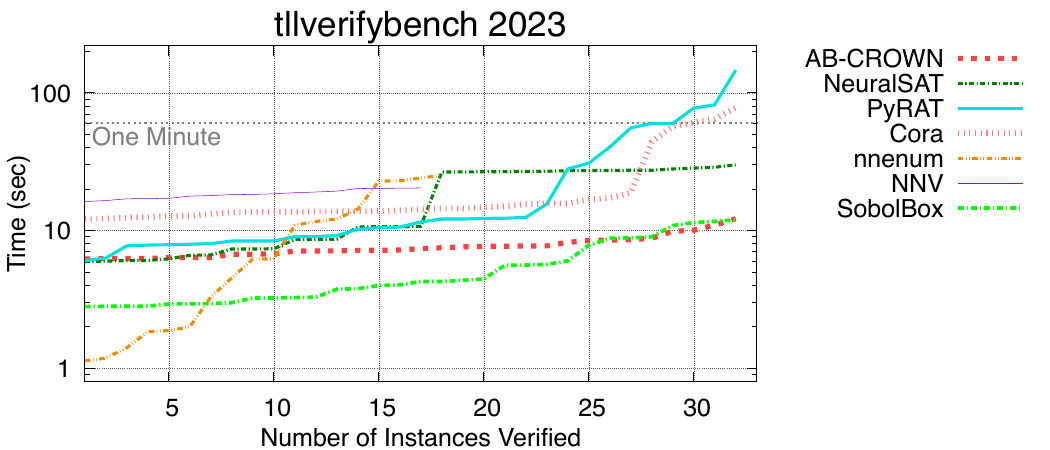}}
\caption{Cactus Plot for tllverifybench 2023.}
\label{fig:quantPic_tll_tol}
\end{figure}

\clearpage
\subsubsection{Extended Track}
\label{sec:benchmark_results_extended_tol}

\clearpage

\begin{table}[h]
\begin{center}
\caption{Benchmark \texttt{2025-cctsdb-yolo-2023}} \label{tab:cat_2025_cctsdb_yolo_2023_tol}
{\setlength{\tabcolsep}{2pt}
\begin{tabular}[h]{@{}llllllrrr@{}}
\toprule
\textbf{\# ~} & \textbf{Tool} & \textbf{Verified} & \textbf{Falsified} & \textbf{Fastest} & \textbf{Penalty} & \textbf{Points} & \textbf{Score} & \textbf{Solved}\\
\midrule
1 & PyRAT & 11 & 28 & 0 & 0 & 390 & 100.0 & 100.0\% \\
2 & $\alpha$-$\beta$-CROWN & 11 & 28 & 0 & 0 & 390 & 100.0 & 100.0\% \\
\bottomrule
\end{tabular}
}
\end{center}
\end{table}

\begin{figure}[h]
\centerline{\includegraphics[width=\textwidth]{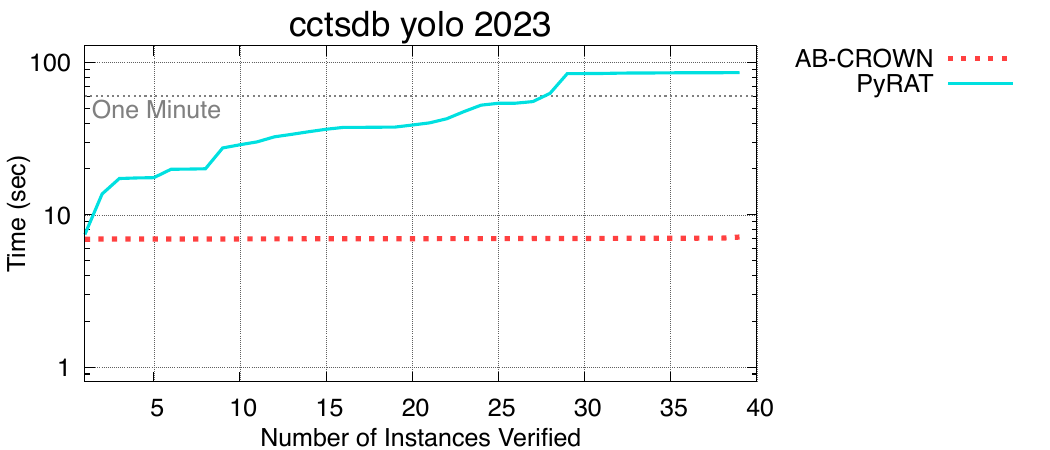}}
\caption{Cactus Plot for cctsdb yolo 2023.}
\label{fig:quantPic_cctsdb_tol}
\end{figure}

\clearpage

\begin{table}[h]
\begin{center}
\caption{Benchmark \texttt{2025-collins-aerospace-benchmark}} \label{tab:cat_2025_collins_aerospace_benchmark_tol}
{\setlength{\tabcolsep}{2pt}
\begin{tabular}[h]{@{}llllllrrr@{}}
\toprule
\textbf{\# ~} & \textbf{Tool} & \textbf{Verified} & \textbf{Falsified} & \textbf{Fastest} & \textbf{Penalty} & \textbf{Points} & \textbf{Score} & \textbf{Solved}\\
\midrule
1 & PyRAT & 0 & 6 & 0 & 0 & 60 & 100.0 & 100.0\% \\
2 & $\alpha$-$\beta$-CROWN & 0 & 6 & 0 & 0 & 60 & 100.0 & 100.0\% \\
\bottomrule
\end{tabular}
}
\end{center}
\end{table}

\begin{figure}[h]
\centerline{\includegraphics[width=\textwidth]{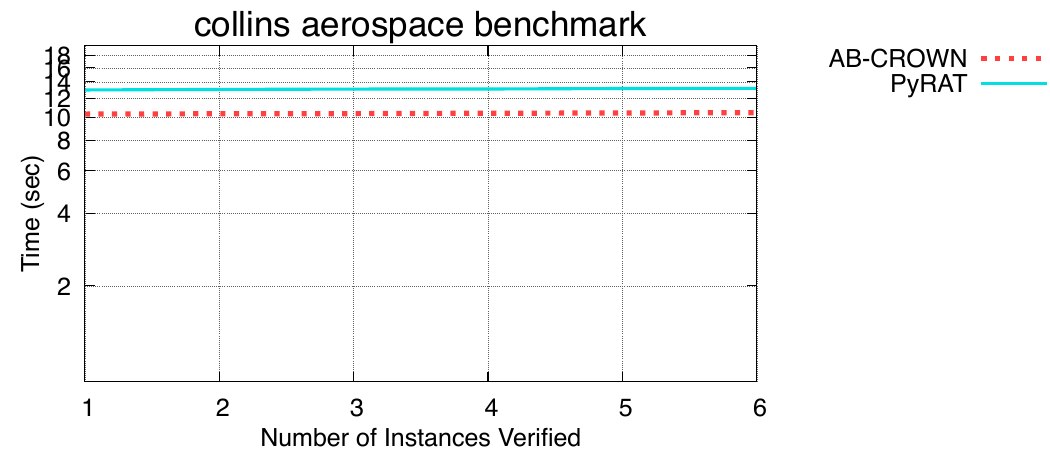}}
\caption{Cactus Plot for collins aerospace benchmark.}
\label{fig:quantPic_collins_tol}
\end{figure}

\clearpage

\begin{table}[h]
\begin{center}
\caption{Benchmark \texttt{2025-lsnc-relu}} \label{tab:cat_2025_lsnc_relu_tol}
{\setlength{\tabcolsep}{2pt}
\begin{tabular}[h]{@{}llllllrrr@{}}
\toprule
\textbf{\# ~} & \textbf{Tool} & \textbf{Verified} & \textbf{Falsified} & \textbf{Fastest} & \textbf{Penalty} & \textbf{Points} & \textbf{Score} & \textbf{Solved}\\
\midrule
1 & $\alpha$-$\beta$-CROWN & 68 & 12 & 0 & 0 & 800 & 100.0 & 100.0\% \\
\bottomrule
\end{tabular}
}
\end{center}
\end{table}

\begin{figure}[h]
\centerline{\includegraphics[width=\textwidth]{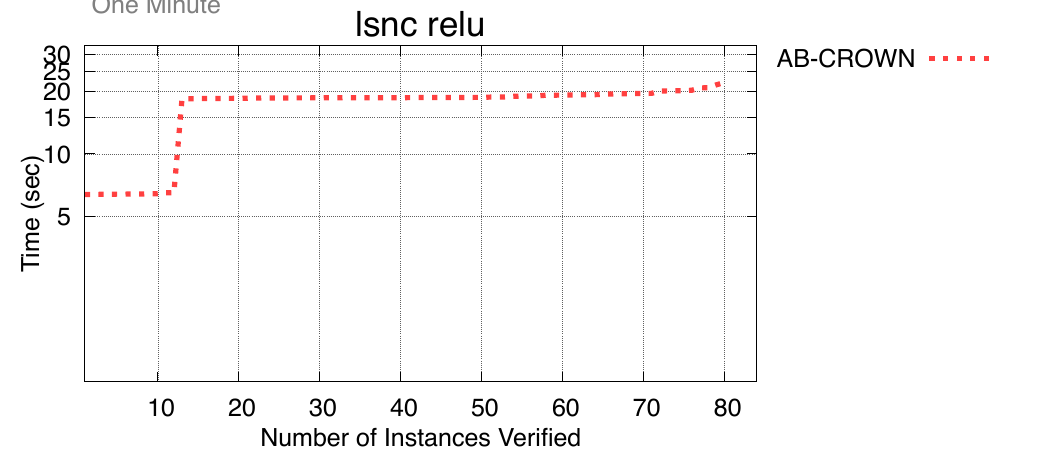}}
\caption{Cactus Plot for lsnc relu.}
\label{fig:quantPic_lsnc_tol}
\end{figure}

\clearpage

\begin{table}[h]
\begin{center}
\caption{Benchmark \texttt{2025-ml4acopf-2024}} \label{tab:cat_2025_ml4acopf_2024_tol}
{\setlength{\tabcolsep}{2pt}
\begin{tabular}[h]{@{}llllllrrr@{}}
\toprule
\textbf{\# ~} & \textbf{Tool} & \textbf{Verified} & \textbf{Falsified} & \textbf{Fastest} & \textbf{Penalty} & \textbf{Points} & \textbf{Score} & \textbf{Solved}\\
\midrule
1 & $\alpha$-$\beta$-CROWN & 58 & 5 & 0 & 0 & 630 & 100.0 & 91.3\% \\
2 & SobolBox & 54 & 3 & 0 & 0 & 570 & 90.5 & 82.6\% \\
3 & PyRAT & 36 & 3 & 0 & 0 & 390 & 61.9 & 56.5\% \\
4 & NNV & 17 & 0 & 0 & 0 & 170 & 27.0 & 24.6\% \\
\bottomrule
\end{tabular}
}
\end{center}
\end{table}

\begin{figure}[h]
\centerline{\includegraphics[width=\textwidth]{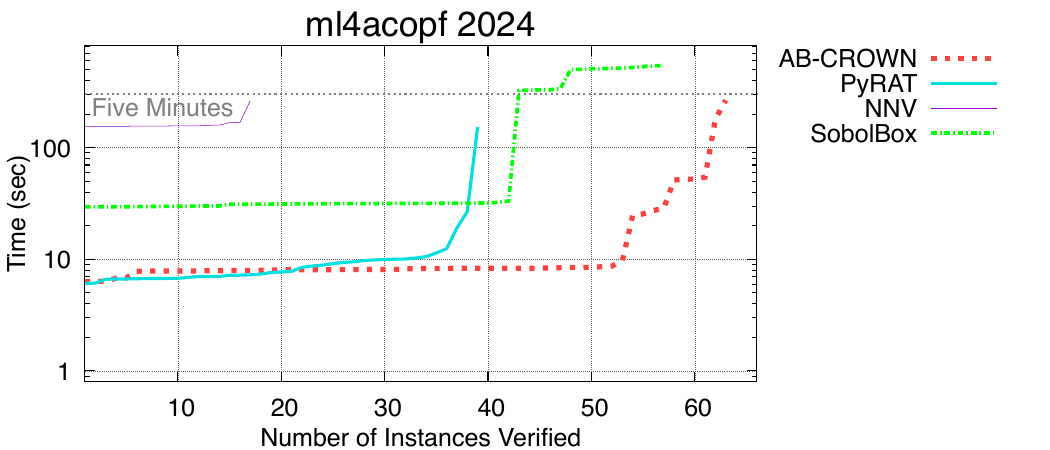}}
\caption{Cactus Plot for ml4acopf 2024.}
\label{fig:quantPic_ml4acopf_tol}
\end{figure}

\clearpage

\begin{table}[h]
\begin{center}
\caption{Benchmark \texttt{2025-relusplitter}} \label{tab:cat_2025_relusplitter_tol}
{\setlength{\tabcolsep}{2pt}
\begin{tabular}[h]{@{}llllllrrr@{}}
\toprule
\textbf{\# ~} & \textbf{Tool} & \textbf{Verified} & \textbf{Falsified} & \textbf{Fastest} & \textbf{Penalty} & \textbf{Points} & \textbf{Score} & \textbf{Solved}\\
\midrule
1 & $\alpha$-$\beta$-CROWN & 151 & 20 & 0 & 0 & 1710 & 100.0 & 77.7\% \\
2 & PyRAT & 61 & 16 & 0 & 0 & 770 & 45.0 & 35.0\% \\
3 & nnenum & 22 & 2 & 0 & 0 & 240 & 14.0 & 10.9\% \\
4 & CORA & 6 & 0 & 0 & 0 & 60 & 3.5 & 2.7\% \\
5 & NNV & 0 & 4 & 0 & 0 & 40 & 2.3 & 1.8\% \\
\bottomrule
\end{tabular}
}
\end{center}
\end{table}

\begin{figure}[h]
\centerline{\includegraphics[width=\textwidth]{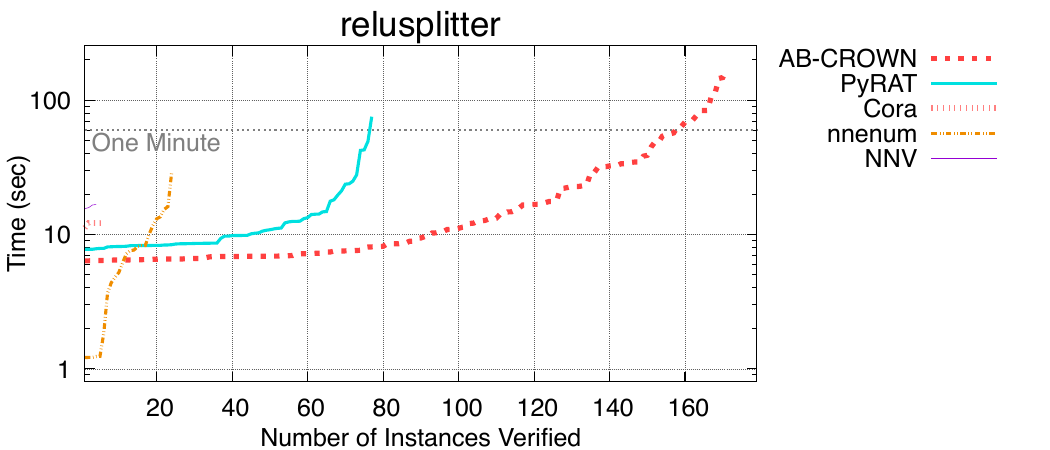}}
\caption{Cactus Plot for relusplitter.}
\label{fig:quantPic_relusplitter_tol}
\end{figure}

\clearpage

\begin{table}[h]
\begin{center}
\caption{Benchmark \texttt{2025-traffic-signs-recognition-2023}} \label{tab:cat_2025_traffic_signs_recognition_2023_tol}
{\setlength{\tabcolsep}{2pt}
\begin{tabular}[h]{@{}llllllrrr@{}}
\toprule
\textbf{\# ~} & \textbf{Tool} & \textbf{Verified} & \textbf{Falsified} & \textbf{Fastest} & \textbf{Penalty} & \textbf{Points} & \textbf{Score} & \textbf{Solved}\\
\midrule
1 & $\alpha$-$\beta$-CROWN & 0 & 43 & 0 & 0 & 430 & 100.0 & 95.6\% \\
\bottomrule
\end{tabular}
}
\end{center}
\end{table}

\begin{figure}[h]
\centerline{\includegraphics[width=\textwidth]{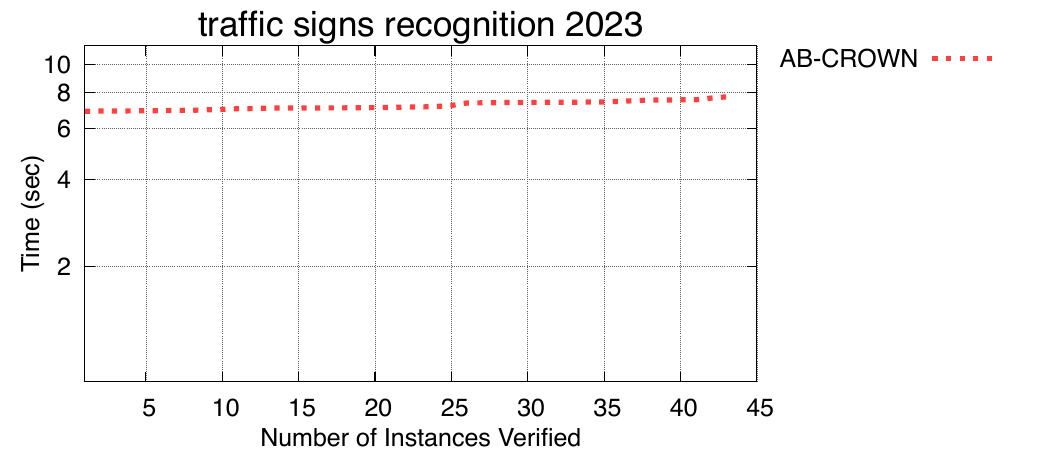}}
\caption{Cactus Plot for traffic signs recognition 2023.}
\label{fig:quantPic_traffic_tol}
\end{figure}

\clearpage

\begin{table}[h]
\begin{center}
\caption{Benchmark \texttt{2025-vggnet16-2022}} \label{tab:cat_2025_vggnet16_2022_tol}
{\setlength{\tabcolsep}{2pt}
\begin{tabular}[h]{@{}llllllrrr@{}}
\toprule
\textbf{\# ~} & \textbf{Tool} & \textbf{Verified} & \textbf{Falsified} & \textbf{Fastest} & \textbf{Penalty} & \textbf{Points} & \textbf{Score} & \textbf{Solved}\\
\midrule
1 & $\alpha$-$\beta$-CROWN & 17 & 1 & 0 & 0 & 180 & 100.0 & 100.0\% \\
2 & nnenum & 14 & 1 & 0 & 0 & 150 & 83.3 & 83.3\% \\
3 & PyRAT & 13 & 1 & 0 & 0 & 140 & 77.8 & 77.8\% \\
4 & NNV & 0 & 1 & 0 & 0 & 10 & 5.6 & 5.6\% \\
\bottomrule
\end{tabular}
}
\end{center}
\end{table}

\begin{figure}[h]
\centerline{\includegraphics[width=\textwidth]{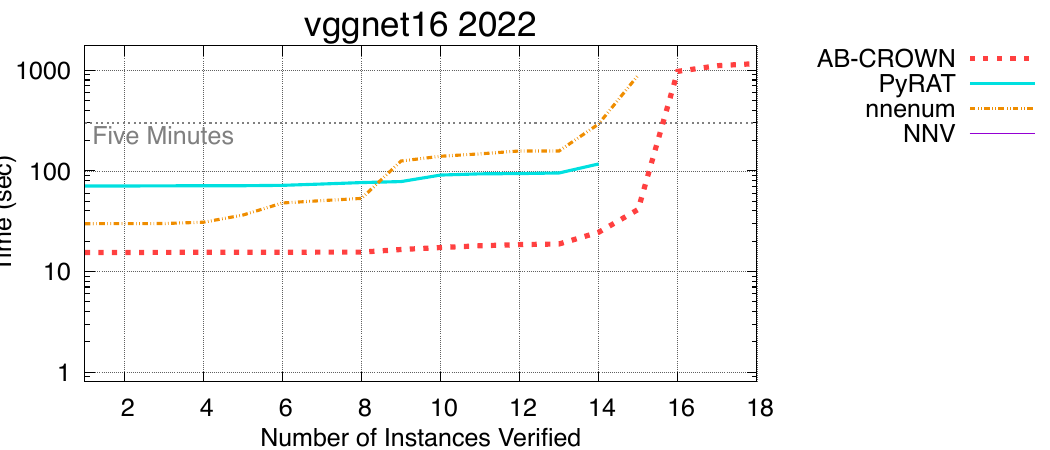}}
\caption{Cactus Plot for vggnet16 2022.}
\label{fig:quantPic_vggnet16_tol}
\end{figure}

\clearpage

\begin{table}[h]
\begin{center}
\caption{Benchmark \texttt{2025-vit-2023}} \label{tab:cat_2025_vit_2023_tol}
{\setlength{\tabcolsep}{2pt}
\begin{tabular}[h]{@{}llllllrrr@{}}
\toprule
\textbf{\# ~} & \textbf{Tool} & \textbf{Verified} & \textbf{Falsified} & \textbf{Fastest} & \textbf{Penalty} & \textbf{Points} & \textbf{Score} & \textbf{Solved}\\
\midrule
1 & $\alpha$-$\beta$-CROWN & 125 & 0 & 0 & 0 & 1250 & 100.0 & 62.5\% \\
2 & PyRAT & 83 & 0 & 0 & 0 & 830 & 66.4 & 41.5\% \\
3 & NeuralSAT & 67 & 0 & 0 & 0 & 670 & 53.6 & 33.5\% \\
\bottomrule
\end{tabular}
}
\end{center}
\end{table}

\begin{figure}[h]
\centerline{\includegraphics[width=\textwidth]{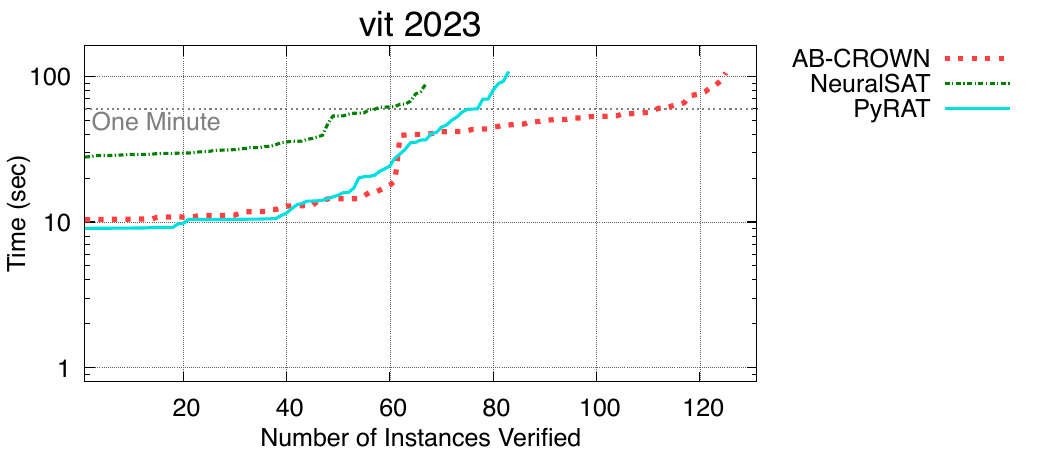}}
\caption{Cactus Plot for vit 2023.}
\label{fig:quantPic_vit_tol}
\end{figure}

\clearpage

\begin{table}[h]
\begin{center}
\caption{Benchmark \texttt{2025-yolo-2023}} \label{tab:cat_2025_yolo_2023_tol}
{\setlength{\tabcolsep}{2pt}
\begin{tabular}[h]{@{}llllllrrr@{}}
\toprule
\textbf{\# ~} & \textbf{Tool} & \textbf{Verified} & \textbf{Falsified} & \textbf{Fastest} & \textbf{Penalty} & \textbf{Points} & \textbf{Score} & \textbf{Solved}\\
\midrule
1 & NNV & 71 & 0 & 0 & 0 & 710 & 100.0 & 98.6\% \\
2 & $\alpha$-$\beta$-CROWN & 61 & 0 & 0 & 0 & 610 & 85.9 & 84.7\% \\
3 & NeuralSAT & 52 & 0 & 0 & 0 & 520 & 73.2 & 72.2\% \\
4 & PyRAT & 40 & 0 & 0 & 0 & 400 & 56.3 & 55.6\% \\
\bottomrule
\end{tabular}
}
\end{center}
\end{table}

\begin{figure}[h]
\centerline{\includegraphics[width=\textwidth]{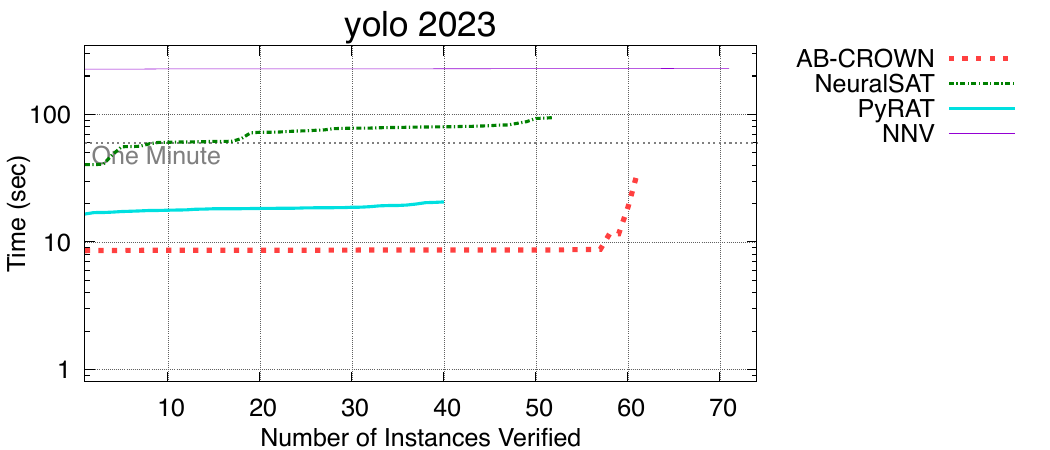}}
\caption{Cactus Plot for yolo 2023.}
\label{fig:quantPic_yolo_tol}
\end{figure}

\clearpage
\section{Instance Results}

\subsection{Zero Tolerance}
\label{sec:results_detailed}

\begin{center}
{\setlength{\tabcolsep}{1pt}
\scriptsize
\begin{longtable}{@{}llllllllll@{}}
\caption{\footnotesize Instance Runtimes. Fastest times are \textcolor{blue}{blue}. Second fastest are \textcolor{second}{green}. Penalties are red crosses (\textbf{\textcolor{red}{\ding{55}}}).} \label{tab:all_results} \\
\toprule
\textbf{Category} & \textbf{Id} & \textbf{Result} & \textbf{$\alpha$-$\beta$-C} & \textbf{NSAT} & \textbf{PyRAT} & \textbf{CORA} & \textbf{NNV} & \textbf{NNen} & \textbf{SB} \\
\midrule
\endhead
2025 Acasxu 2023 & 0 & ~\textsc{unsat} & \textcolor{darkgray}{8.45} & \textcolor{darkgray}{18.1} & \textcolor{darkgray}{13.2} & \textcolor{darkgray}{15.9} & - & \textcolor{darkgray}{1.42} & \textcolor{darkgray}{3.04} \\
2025 Acasxu 2023 & 1 & ~\textsc{unsat} & \textcolor{darkgray}{7.76} & \textcolor{darkgray}{18.1} & \textcolor{darkgray}{16.1} & \textcolor{darkgray}{14.0} & - & \textcolor{darkgray}{1.45} & \textcolor{darkgray}{2.84} \\
2025 Acasxu 2023 & 2 & ~\textsc{unsat} & \textcolor{darkgray}{7.90} & \textcolor{darkgray}{18.3} & \textcolor{darkgray}{18.9} & \textcolor{darkgray}{14.0} & - & \textcolor{darkgray}{1.52} & \textcolor{darkgray}{2.84} \\
2025 Acasxu 2023 & 3 & ~\textsc{unsat} & \textcolor{darkgray}{7.92} & \textcolor{darkgray}{18.3} & \textcolor{darkgray}{16.0} & \textcolor{darkgray}{12.7} & - & \textcolor{darkgray}{1.41} & \textcolor{darkgray}{2.83} \\
2025 Acasxu 2023 & 4 & ~\textsc{unsat} & \textcolor{darkgray}{7.77} & \textcolor{darkgray}{18.2} & \textcolor{darkgray}{12.7} & \textcolor{darkgray}{13.8} & - & \textcolor{darkgray}{1.45} & \textcolor{darkgray}{2.88} \\
2025 Acasxu 2023 & 5 & ~\textsc{unsat} & \textcolor{darkgray}{7.83} & \textcolor{darkgray}{18.1} & \textcolor{darkgray}{16.6} & \textcolor{darkgray}{13.5} & - & \textcolor{darkgray}{1.65} & \textcolor{darkgray}{2.82} \\
2025 Acasxu 2023 & 6 & ~\textsc{unsat} & \textcolor{darkgray}{7.75} & \textcolor{darkgray}{18.0} & \textcolor{darkgray}{13.1} & \textcolor{darkgray}{13.5} & - & \textcolor{darkgray}{1.32} & \textcolor{darkgray}{2.87} \\
2025 Acasxu 2023 & 7 & ~\textsc{unsat} & \textcolor{darkgray}{7.76} & \textcolor{darkgray}{18.0} & \textcolor{darkgray}{10.6} & \textcolor{darkgray}{12.3} & - & \textcolor{darkgray}{1.48} & \textcolor{darkgray}{2.85} \\
\bottomrule
\end{longtable}
}
\end{center}

\subsection{Small Tolerance}
\label{sec:results_detailed_tol}

\begin{center}
{\setlength{\tabcolsep}{1pt}
\scriptsize
\begin{longtable}{@{}llllllllll@{}}
\caption{\footnotesize Instance Runtimes. Fastest times are \textcolor{blue}{blue}. Second fastest are \textcolor{second}{green}. Penalties are red crosses (\textbf{\textcolor{red}{\ding{55}}}).} \label{tab:all_results_tol} \\
\toprule
\textbf{Category} & \textbf{Id} & \textbf{Result} & \textbf{$\alpha$-$\beta$-C} & \textbf{NSAT} & \textbf{PyRAT} & \textbf{CORA} & \textbf{NNen} & \textbf{NNV} & \textbf{SB} \\
\midrule
\endhead
2025 Acasxu 2023 & 0 & ~\textsc{unsat} & \textcolor{darkgray}{8.45} & \textcolor{darkgray}{18.1} & \textcolor{darkgray}{13.2} & \textcolor{darkgray}{15.9} & \textcolor{darkgray}{1.42} & - & \textcolor{darkgray}{3.04} \\
2025 Acasxu 2023 & 1 & ~\textsc{unsat} & \textcolor{darkgray}{7.76} & \textcolor{darkgray}{18.1} & \textcolor{darkgray}{16.1} & \textcolor{darkgray}{14.0} & \textcolor{darkgray}{1.45} & - & \textcolor{darkgray}{2.84} \\
2025 Acasxu 2023 & 2 & ~\textsc{unsat} & \textcolor{darkgray}{7.90} & \textcolor{darkgray}{18.3} & \textcolor{darkgray}{18.9} & \textcolor{darkgray}{14.0} & \textcolor{darkgray}{1.52} & - & \textcolor{darkgray}{2.84} \\
2025 Acasxu 2023 & 3 & ~\textsc{unsat} & \textcolor{darkgray}{7.92} & \textcolor{darkgray}{18.3} & \textcolor{darkgray}{16.0} & \textcolor{darkgray}{12.7} & \textcolor{darkgray}{1.41} & - & \textcolor{darkgray}{2.83} \\
2025 Acasxu 2023 & 4 & ~\textsc{unsat} & \textcolor{darkgray}{7.77} & \textcolor{darkgray}{18.2} & \textcolor{darkgray}{12.7} & \textcolor{darkgray}{13.8} & \textcolor{darkgray}{1.45} & - & \textcolor{darkgray}{2.88} \\
2025 Acasxu 2023 & 5 & ~\textsc{unsat} & \textcolor{darkgray}{7.83} & \textcolor{darkgray}{18.1} & \textcolor{darkgray}{16.6} & \textcolor{darkgray}{13.5} & \textcolor{darkgray}{1.65} & - & \textcolor{darkgray}{2.82} \\
2025 Acasxu 2023 & 6 & ~\textsc{unsat} & \textcolor{darkgray}{7.75} & \textcolor{darkgray}{18.0} & \textcolor{darkgray}{13.1} & \textcolor{darkgray}{13.5} & \textcolor{darkgray}{1.32} & - & \textcolor{darkgray}{2.87} \\
2025 Acasxu 2023 & 7 & ~\textsc{unsat} & \textcolor{darkgray}{7.76} & \textcolor{darkgray}{18.0} & \textcolor{darkgray}{10.6} & \textcolor{darkgray}{12.3} & \textcolor{darkgray}{1.48} & - & \textcolor{darkgray}{2.85} \\
\bottomrule
\end{longtable}
}
\end{center}


\end{document}